\documentclass{article}

\PassOptionsToPackage{numbers}{natbib}
\usepackage[preprint]{neurips_2025}
\setcitestyle{numbers,square,citesep={,}}

\usepackage[utf8]{inputenc}
\usepackage[T1]{fontenc}

\usepackage[sc]{mathpazo}

\usepackage{amsmath, amssymb, amsfonts, amsthm}
\usepackage{mathrsfs}
\usepackage{booktabs}
\usepackage{nicefrac}

\usepackage{microtype}
\usepackage{xcolor}
\usepackage{adjustbox}
\usepackage{enumitem}
\usepackage{xspace}
\usepackage{algorithm}
\usepackage{algorithmic}
\usepackage{fontawesome5}
\usepackage{url}
\usepackage{xurl}
\usepackage{subcaption}
\usepackage{multirow}
\usepackage{wrapfig}
\usepackage{graphicx}
\usepackage{placeins}
\usepackage{hyperref}
\newcommand{\ours}[0]{\textbf{EnFlow}\xspace}
\newcommand{\oursregular}[0]{EnFlow\xspace}

\theoremstyle{plain}
\newtheorem{theorem}{Theorem}
\newtheorem{proposition}[theorem]{Proposition}

\theoremstyle{definition}

\theoremstyle{remark}

\renewcommand{\tableautorefname}{Tab.}

\title{Energy-Guided Generative Modeling for\\ Low-Energy Molecular Structure Discovery  
}

\author{ Guikun Xu$^{1}$, Xiaohan Yi$^{3}$, Ziqiao Meng$^{2}$, Peilin Zhao$^{1,}$\textsuperscript{\faEnvelope},
Yatao Bian$^{2,}$\textsuperscript{\faEnvelope} \\[5pt]
$^{1}$School of Artificial Intelligence, Shanghai Jiao Tong University, Shanghai,
China\\
$^{2}$Department of Computer Science, National University of Singapore,\\
Singapore, Singapore \\
$^{3}$Shenzhen International Graduate School, Tsinghua University, Shenzhen,
China\\
[6pt]\faTools~:~\texttt{richxu945@sjtu.edu.cn} \\
[6pt]\faEnvelope~:~\texttt{peilinzhao@sjtu.edu.cn}; \texttt{ybian@nus.edu.sg} \\
[6pt]\faGithub~:~\texttt{\url{https://github.com/Rich-XGK/EnFlow}}
}

\begin{document}
   \maketitle
   \begin{abstract}
      \vspace{-0.2cm}

Exploring molecular energy landscapes and identifying ground-state conformations
are central challenges in computational chemistry. However, generating diverse
low-energy conformers from molecular graphs remains expensive with traditional
physics-based pipelines. Existing learning-based approaches remain fragmented:
generative models capture conformational diversity but often lack reliable energy
calibration, whereas deterministic predictors focus on a single structure and
fail to represent ensemble variability.
Here we introduce \ours, to our knowledge, the first energy-guided generative
framework that couples flow-based conformer generation with explicit energy
landscape modeling for joint conformational ensemble generation and ground-state
identification. By integrating generative dynamics with a learned energy model,
\ours guides sampling toward low-energy regions of the conformational landscape,
improving structural fidelity under extremely few sampling steps while enabling
energy-based ranking of generated conformations.
Experiments on GEOM-QM9 and GEOM-Drugs show that \ours achieves strong
performance in conformer generation and ground-state identification while
requiring only 1--2 ordinary differential equation (ODE) sampling steps.
Single-point GFN2-xTB evaluations further
show that the learned energy scores preserve physically meaningful energetic
rankings of generated conformations. These results support explicit energy
landscape modeling as an effective strategy for low-energy molecular structure
discovery through joint modeling of conformational ensembles and their associated
energies.

   \end{abstract}
   \begin{figure*}[t]
    \refstepcounter{figure}
    \centering
    \vspace{-0.5cm}
    \includegraphics[width=\textwidth]{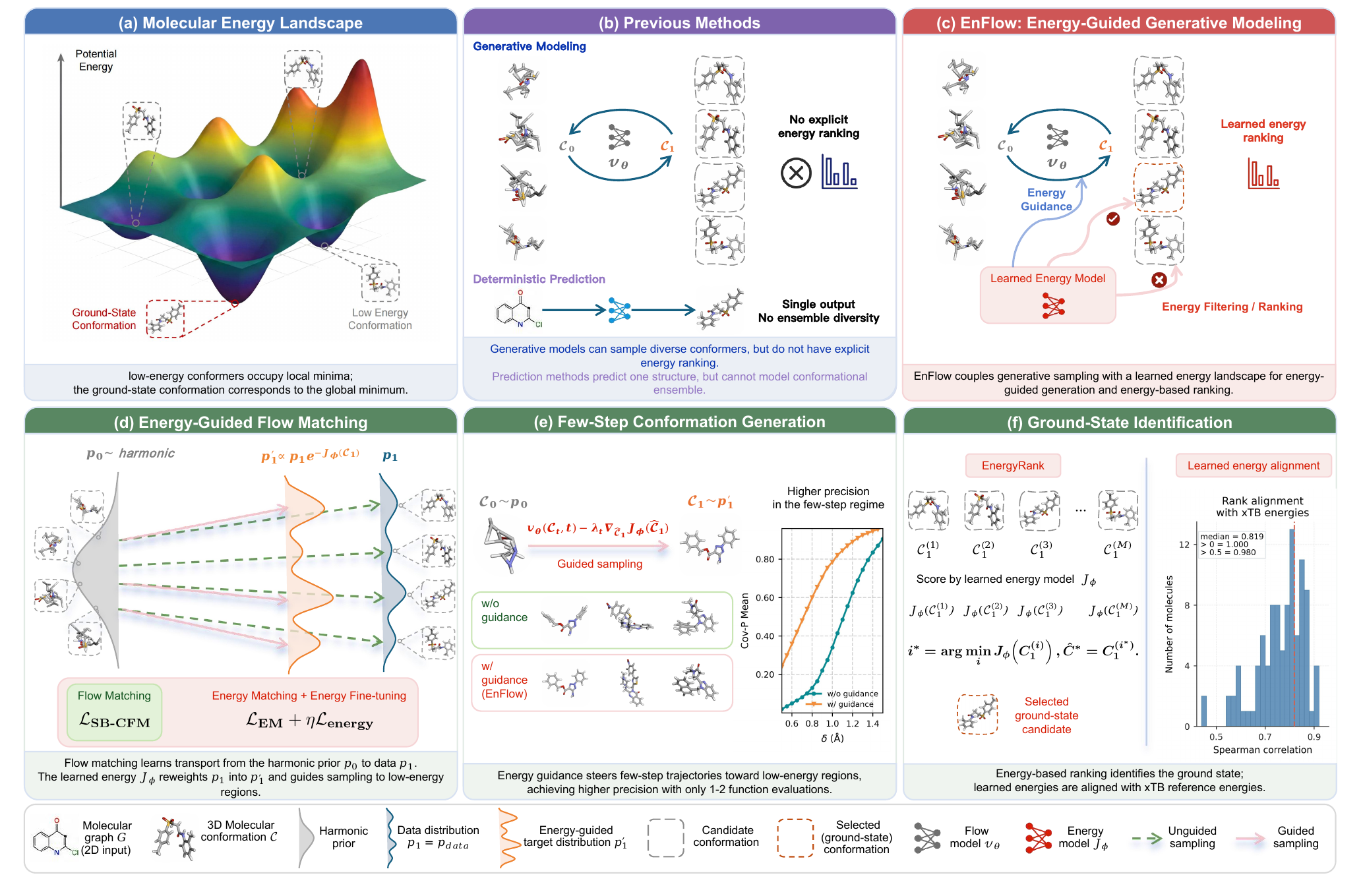}
    \par\smallskip
    \textbf{Figure~\thefigure}
    \vspace{-0.35cm}
    \label{fig:task_teaser}
\end{figure*}
\vspace{-0.5cm}
\section{Introduction}

Understanding molecular conformations and their underlying energy landscapes is central
to computational chemistry and molecular design. Given a molecular graph
specifying atom and bond types, the three-dimensional coordinates of its atoms define
a set of possible conformations that determine chemical properties and biological
functions~\cite{guimaraes2012use,schwab2010conformations,bhati2026integrative}. Identifying
energetically favorable conformations is therefore a fundamental problem in
computational chemistry and drug discovery~\cite{hawkins2017conformation}. Conventional
approaches, including molecular dynamics simulations~\cite{ballard2015exploiting,de2016role,pracht2020automated}
and density functional theory~\cite{parr1979local} optimizations, are computationally
intensive and time-consuming, limiting their applicability to large-scale
studies. Among these conformations, the ground-state structure corresponds to
the global minimum of the molecular potential energy surface~\cite{muller1994glossary}
and is therefore the most thermodynamically stable. Accurately identifying this state
is essential for understanding binding affinities, reaction mechanisms, and spectroscopic
properties, and for reliable molecular prediction and rational design.

Recent advances in machine learning~\cite{zhou2021machine,alpaydin2021machine,janiesch2021machine}, particularly generative modeling, have substantially
improved the efficiency of molecular conformer generation.
Modern generative approaches, including diffusion and flow-based models, can
sample diverse conformations while maintaining geometric consistency~\cite{sohl2015deep,song2019generative,ho2020denoising,lipmanflow,liu2022flow}.
In particular, GeoDiff~\cite{xu2022geodiff} introduced an equivariant diffusion
formulation for molecular structures, improving conformer sampling under geometric
constraints. More recently, flow matching has emerged as an alternative generative
paradigm with improved sample efficiency and faster generation, making it
attractive for large-scale conformer generation~\cite{lipmanflow,liu2022flow,hassan2024flow,cao2025efficient}.
Despite these advances, accurately identifying the ground-state conformation
from the sampled ensemble remains challenging.

To address this challenge, recent studies have explored deterministic prediction
methods that directly model the ground-state conformation, reducing reliance on
stochastic sampling and post hoc ranking. In these approaches~\citep{xu2021molecule3d,xu2023gtmgc,luo2024bridging,wangwgformer,kim2025rebind},
neural networks---particularly graph transformers~\cite{ying2021transformers} with
task-specific adaptations---are trained to regress the ground-state structure from
datasets that provide a single ground-state label for each molecule. While such
approaches can produce structures close to the ground state, they inherently focus
on predicting a single conformation and therefore do not capture the broader conformational
ensemble associated with the molecular energy landscape, which governs
structural variability and thermodynamic stability.

As illustrated in~\autoref{fig:task_teaser}(b), a practical computational
framework for low-energy molecular structure discovery remains lacking. Generative
approaches~\cite{xu2022geodiff,hassan2024flow} can sample diverse conformations but
typically lack an explicit and well-calibrated representation of the underlying energy
landscape, making reliable energetic evaluation and ground-state identification
difficult. Deterministic prediction methods~\cite{xu2023gtmgc,kim2025rebind}, by
contrast, aim to directly approximate the ground-state structure but inherently produce
only a single conformation and therefore fail to capture ensemble variability.
As a result, existing approaches remain fragmented not only between ensemble generation
and ground-state prediction, but more fundamentally between conformational
generation and energetic evaluation. This fragmentation highlights the need for a
unified computational framework that can simultaneously capture conformational diversity,
model the associated energy landscape, and identify energetically stable structures.

To address these challenges, we introduce \ours, to our knowledge, the first
energy-guided generative framework that couples flow-based conformational generation
with explicit energy landscape modeling for low-energy molecular structure
discovery. The framework integrates generative dynamics with a learned energy
model that guides sampling toward low-energy regions of molecular conformational
space, thereby supporting both low-energy conformer discovery and ground-state
identification within a single formulation. A key reason why direct energy-guided conformer generation has remained difficult
is that conventional energy-based model training, such as contrastive divergence,
requires expensive and often unstable negative sampling. This difficulty is
amplified in molecular conformational modeling, where a single model must capture
energy landscapes across many chemically distinct small molecules. \ours{}
addresses this challenge by training the energy function with the Energy Matching
objective~\citep{balcerak2025energy}, followed by supervised energy fine-tuning
to capture molecule-specific energetic variation. By making explicit energy landscape learning feasible within a flow-based
generative model, \ours{} provides a unified computational strategy for generating
conformational ensembles, evaluating their energetic plausibility, and
prioritizing energetically stable molecular structures.

Experiments on GEOM-QM9 and GEOM-Drugs show that \oursregular improves few-step
conformer generation and ground-state identification, particularly under highly
limited ODE sampling budgets. Beyond these task-level evaluations, we further
assess the physical relevance of the learned energy scores using single-point
GFN2-xTB calculations~\citep{bannwarth2019gfn2}, a semi-empirical quantum-chemical
method commonly used in prior conformer-generation studies to evaluate energetic
properties of generated ensembles~\citep{jing2022torsional,wang2023swallowing,hassan2024flow,cao2025efficient}.
The results show that the learned energy scores preserve physically meaningful
energetic rankings of generated conformations. Together, these findings support
explicit energy landscape modeling as an effective strategy for low-energy
molecular structure discovery through the joint modeling of conformational
ensembles and their associated energies.

   \section{Results}

We evaluate whether an energy-guided generative framework can jointly support three
capabilities required for low-energy molecular structure discovery:
conformational ensemble generation, ground-state identification, and physically
meaningful energetic ranking. Experiments are conducted on GEOM-QM9 and GEOM-Drugs~\cite{axelrod2022geom},
which provide diverse molecular conformations with associated energy annotations.
Across these benchmarks, \ours improves few-step conformer generation, enables
learned-energy-based ground-state identification from generated ensembles, and
learns energy scores that align with single-point GFN2-xTB energetic rankings.

\begin{wrapfigure}
   {r}{0.48\textwidth}
   \refstepcounter{figure}
   \centering
   \vspace{-0.4cm}
   \includegraphics[width=0.46\textwidth]{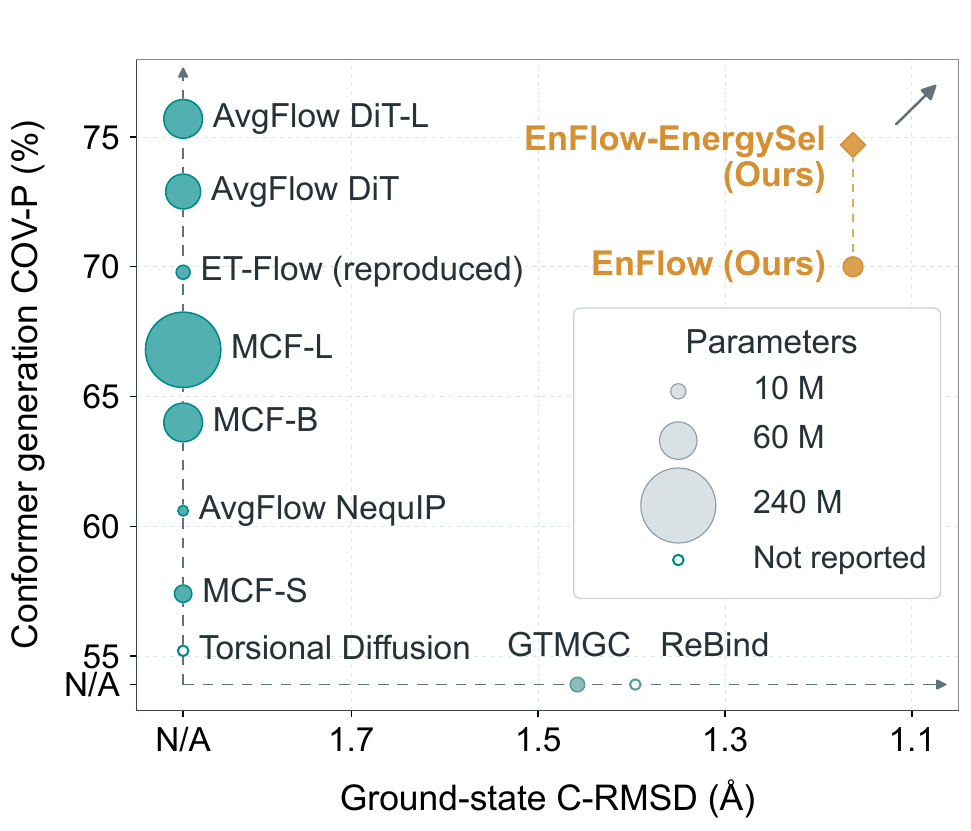}
   \par\smallskip
   \textbf{Figure~\thefigure}
   \label{fig: joint_performance}
   \vspace{-0.35cm}
\end{wrapfigure}

As illustrated in~\autoref{fig: joint_performance}, \oursregular achieves a favorable
balance between ensemble generation quality and ground-state prediction accuracy
while using a relatively small number of neural parameters. The following
sections define the evaluation setup and then examine three aspects of the
framework: few-step conformer generation (the ``Energy-guided flow matching improves
few-step conformer generation'' subsection), ground-state identification (the
``Learned energy enables ground-state identification from generated ensembles''
subsection), and the physical relevance of the learned energy scores (the ``Learned
energy scores align with quantum-chemical energetic rankings'' subsection).
Additional mechanistic analyses are provided in the ``More Experimental Results''
subsection of the Supplementary Information.

\subsection{Problem formulation and evaluation setup}

We evaluate whether \ours can jointly support two complementary tasks in molecular
conformational modeling: conformer ensemble generation and ground-state
conformation identification. A molecule is represented as $\mathcal{M}:=\{\mathcal{G}
,\mathcal{C}\}$, where $\mathcal{G}$ is the molecular graph and
$\mathcal{C}\in\mathbb{R}^{n\times 3}$ is the 3D conformation. Given $\mathcal{G}$,
conformer generation aims to sample plausible low-energy conformations from
$p(\mathcal{C}\mid\mathcal{G})$, whereas ground-state identification aims to recover
the most stable conformation $\mathcal{C}^{*}$.

Experiments are conducted on GEOM-QM9 and GEOM-Drugs~\citep{axelrod2022geom}, following
the splits of Refs.~\citep{ganea2021geomol,jing2022torsional}. Each molecule
contains multiple low-energy conformations with Boltzmann energies and weights. Following
Ref.~\citep{kim2025rebind}, we define the ground-state conformation as the
conformer with the highest Boltzmann weight. Both test sets contain 1,000 molecules;
additional dataset details are provided in the ``Dataset Details'' subsection
of the Supplementary Information.

For conformer ensemble generation, we report root mean square deviation
(RMSD)-based coverage (COV) and average minimum RMSD (AMR) under
Recall and Precision protocols, yielding recall coverage (COV-R), recall
average minimum RMSD (AMR-R), precision coverage (COV-P), and precision average
minimum RMSD (AMR-P). Recall-oriented metrics measure coverage of reference
conformational diversity, whereas Precision-oriented metrics measure the quality
of generated conformations. Unless otherwise specified, the RMSD threshold is
$0. 5~\text{\AA}$ for GEOM-QM9 and $0.75~\text{\AA}$ for GEOM-Drugs. For ground-state
identification, we report coordinate root mean square deviation
($\mathcal{C}\text{-RMSD}$), distance mean absolute error
($\mathbf{D}\text{-MAE}$), and distance root mean squared error
($\mathbf{D}\text{-RMSE}$), which measure coordinate- and
pairwise-distance-level agreement with the ground-state conformation. Full metric
definitions are provided in the ``Metric Definitions'' subsection of the
Supplementary Information.

For conformer generation, we compare against GeoMol~\citep{ganea2021geomol},
GeoDiff~\citep{xu2022geodiff}, Torsional Diffusion~\citep{jing2022torsional}, ET-Flow~\citep{hassan2024flow},
MCF~\citep{wang2023swallowing}, and AvgFlow~\citep{cao2025efficient}, including
their reported few-step variants where applicable. For ground-state identification,
we compare with RDKit~\citep{landrum2013rdkit}, GINE~\citep{hu2019strategies},
GATv2~\citep{brody2021attentive}, GraphGPS~\citep{rampavsek2022recipe}, GTMGC~\citep{xu2023gtmgc},
Torsional Diffusion~\citep{jing2022torsional}, and ReBind~\citep{kim2025rebind},
following Ref.~\citep{kim2025rebind}. Within the ET-Flow family, ET-Flow denotes
the model with post hoc correction, whereas ET-Flow-$SO(3)$ denotes the
special orthogonal group in three dimensions ($SO(3)$)-equivariant
variant. We follow the same convention and denote our variants by \ours and
\ours-$SO(3)$.

\subsection{Unified performance across ensemble generation and ground-state
identification}

We first examine whether \ours provides a unified solution for the two central objectives
considered in this work: the generation of conformational ensembles and the
 identification of ground-state conformations. These objectives are typically addressed by different
model classes. Generative methods focus on sampling diverse conformers, whereas deterministic
predictors aim to output a single low-energy structure. In contrast, \ours couples
conformer generation with a learned energy function, allowing the same framework
to support both sampling and ranking.

As shown in~\autoref{fig: joint_performance}, \ours achieves strong performance across
both conformer generation and ground-state identification on GEOM-Drugs, while using
fewer neural parameters than many competing methods. The standard \ours model performs
well on both axes, indicating that energy-guided generative dynamics can support
ensemble-level sampling and learned-energy-based ground-state identification within a single
framework. When the learned energy model is further used for post-generation selection,
\ours-\textsc{EnergySel} improves conformer-generation precision without changing
the learned-energy-based ground-state identification procedure. This suggests that the learned energy
function provides a shared mechanism for generating low-energy conformations and
selecting energetically favorable structures from generated ensembles.

The following sections analyze these capabilities in detail. We first evaluate few-step
conformer generation, then assess learned-energy-based ground-state identification from generated
ensembles, and finally examine whether the learned energy scores align with quantum-chemical
energies.

\begin{table*}
   [t]
   \caption{Molecular conformer generation results on GEOM-QM9 ($\delta = 0.5\text{\AA}$).}
   \label{tab:qm9}
   \centering
   \begin{adjustbox}
      {max width=0.95\textwidth}
      \renewcommand{\arraystretch}{1.1}
      \begin{tabular}{lccccccccc}
         \toprule \multirow{2}{*}{Method}                                          & \multirow{2}{*}{ODE steps} & \multicolumn{2}{c}{COV-R $\uparrow$} & \multicolumn{2}{c}{AMR-R $\downarrow$} & \multicolumn{2}{c}{COV-P $\uparrow$} & \multicolumn{2}{c}{AMR-P $\downarrow$} \\
         \cmidrule(lr){3-4}\cmidrule(lr){5-6}\cmidrule(lr){7-8}\cmidrule(lr){9-10} &                            & mean                                 & median                                 & mean                                 & median                                & mean                & median               & mean                & median              \\
         \midrule CGCF                                                             & 1000                       & 69.47                                & 96.15                                  & 0.425                                & 0.374                                 & 38.20               & 33.33                & 0.711               & 0.695               \\
         GeoDiff                                                                   & 1000                       & 76.50                                & \textbf{100.00}                        & 0.297                                & 0.229                                 & 50.00               & 33.50                & 1.524               & 0.510               \\
         GeoMol                                                                    & --                         & 91.50                                & \textbf{100.00}                        & 0.225                                & 0.193                                 & 87.60               & \textbf{100.00}      & 0.270               & 0.241               \\
         Torsional Diff.                                                           & 20                         & 92.80                                & \textbf{100.00}                        & 0.178                                & 0.147                                 & 92.70               & \textbf{100.00}      & 0.221               & 0.195               \\
         MCF                                                                       & 1000                       & 95.00                                & \textbf{100.00}                        & 0.103                                & 0.044                                 & 93.70               & \textbf{100.00}      & 0.119               & 0.055               \\
         ET-Flow                                                                   & 50                         & \textbf{96.47}                       & \textbf{100.00}                        & \textbf{0.073}                       & 0.047                                 & 94.05               & \textbf{100.00}      & 0.098               & 0.039               \\
         ET-Flow-$SO(3)$                                                           & 50                         & 95.98                                & \textbf{100.00}                        & \underline{0.076}                    & \underline{0.030}                     & 92.10               & \textbf{100.00}      & 0.110               & 0.047               \\
         $\text{ET-Flow}_{\text{reproduced}}$                                      & 50                         & 95.81                                & \textbf{100.00}                        & 0.076                                & 0.030                                 & 92.30               & \textbf{100.00}      & 0.105               & 0.035               \\
         $\text{ET-Flow-}SO(3)_{\text{reproduced}}$                                & 50                         & 95.69                                & \textbf{100.00}                        & 0.079                                & 0.028                                 & 94.77               & \textbf{100.00}      & 0.088               & 0.033               \\
         $\text{AvgFlow}_{\text{NequIP}}$                                          & 50                         & \underline{96.40}                    & \textbf{100.00}                        & 0.089                                & 0.042                                 & 92.80               & \textbf{100.00}      & 0.132               & 0.084               \\
         \ours                                                                     & 5                          & 95.83                                & \textbf{100.00}                        & 0.082                                & 0.032                                 & 92.67               & \textbf{100.00}      & 0.108               & 0.044               \\
         \ours                                                                     & 50                         & 95.74                                & \textbf{100.00}                        & 0.078                                & \underline{0.030}                     & 92.59               & \textbf{100.00}      & 0.100               & \underline{0.036}   \\
         \ours-$SO(3)$                                                             & 5                          & 96.11                                & \textbf{100.00}                        & 0.078                                & 0.033                                 & \underline{95.11}   & \textbf{100.00}      & \underline{0.090}   & 0.041               \\
         \ours-$SO(3)$                                                             & 50                         & 96.26                                & \textbf{100.00}                        & \underline{0.076}                    & \textbf{0.028}                        & \textbf{95.48}      & \textbf{100.00}      & \textbf{0.083}      & \textbf{0.034}      \\
         \midrule \multicolumn{10}{l}{\emph{Energy-based selection}}                \\
         \ours-$SO(3)$-\textsc{EnergySel}                                          & 5                          & 91.47                                & \textbf{100.00}                        & 0.116                                & 0.053                                 & 96.39               & \textbf{100.00}      & 0.075               & 0.026               \\
         \ours-$SO(3)$-\textsc{EnergySel}                                          & 50                         & 90.91                                & \textbf{100.00}                        & 0.119                                & 0.021                                 & 96.15               & \textbf{100.00}      & 0.069               & 0.021               \\
         \midrule \multicolumn{10}{l}{\textbf{\emph{2-step Generation}}}            \\
         $\text{ET-Flow}_{\text{reproduced}}$                                      & 2                          & \underline{96.27}                    & \textbf{100.00}                        & 0.119                                & 0.071                                 & 90.41               & \textbf{100.00}      & 0.182               & 0.140               \\
         $\text{ET-Flow-}SO(3)_{\text{reproduced}}$                                & 2                          & \textbf{96.70}                       & \textbf{100.00}                        & \underline{0.112}                    & 0.067                                 & \underline{93.54}   & \textbf{100.00}      & 0.152               & 0.110               \\
         $\text{AvgFlow}_{\text{NequIP-R}}$                                        & 2                          & 95.90                                & \textbf{100.00}                        & 0.151                                & 0.104                                 & 87.70               & \textbf{100.00}      & 0.236               & 0.207               \\
         \ours                                                                     & 2                          & 94.81                                & \textbf{100.00}                        & \underline{0.112}                    & \textbf{0.055}                        & 92.58               & \textbf{100.00}      & \underline{0.138}   & \underline{0.075}   \\
         \ours-$SO(3)$                                                             & 2                          & 95.49                                & \textbf{100.00}                        & \textbf{0.108}                       & \underline{0.058}                     & \textbf{95.64}      & \textbf{100.00}      & \textbf{0.115}      & \textbf{0.066}      \\
         \midrule \multicolumn{10}{l}{\emph{Energy-based selection}}                \\
         \ours-$SO(3)$-\textsc{EnergySel}                                          & 2                          & 93.61                                & \textbf{100.00}                        & 0.113                                & 0.055                                 & 96.20               & \textbf{100.00}      & 0.096               & 0.051               \\
         \midrule \multicolumn{10}{l}{\textbf{\emph{1-step Generation}}}            \\
         $\text{ET-Flow}_{\text{reproduced}}$                                      & 1                          & 50.47                                & 50.00                                  & 0.483                                & 0.505                                 & 36.38               & 25.00                & 0.550               & 0.580               \\
         $\text{ET-Flow-}SO(3)_{\text{reproduced}}$                                & 1                          & 76.12                                & \textbf{100.00}                        & 0.330                                & 0.323                                 & 67.93               & 87.50                & 0.382               & 0.387               \\
         $\text{AvgFlow}_{\text{NequIP-D}}$                                        & 1                          & \textbf{95.10}                       & \textbf{100.00}                        & 0.220                                & 0.195                                 & 84.80               & \textbf{100.00}      & 0.304               & 0.283               \\
         \ours                                                                     & 1                          & 89.43                                & \textbf{100.00}                        & \underline{0.215}                    & \underline{0.178}                     & \underline{88.09}   & \textbf{100.00}      & \underline{0.256}   & \underline{0.227}   \\
         \ours-$SO(3)$                                                             & 1                          & \underline{90.45}                    & \textbf{100.00}                        & \textbf{0.195}                       & \textbf{0.150}                        & \textbf{91.23}      & \textbf{100.00}      & \textbf{0.213}      & \textbf{0.177}      \\
         \ours-$SO(3)_{\text{Reflow}}$                                             & 1                          & 96.70                                & 100.00                                  & 0.122                                & 0.087                                 & 93.48               & 100.00              & 0.170               & 0.132               \\
         \midrule \multicolumn{10}{l}{\emph{Energy-based selection}}                \\
         \ours-$SO(3)$-\textsc{EnergySel}                                          & 1                          & 89.64                                & \textbf{100.00}                        & 0.205                                & 0.162                                 & 93.80               & \textbf{100.00}      & 0.202               & 0.162               \\
         \bottomrule
      \end{tabular}
   \end{adjustbox}
   \begin{minipage}{0.95\textwidth}
      \scriptsize \textbf{Table notes.} Results of most baseline methods are
      taken from the corresponding literature. $\text{ET-Flow}_{\text{reproduced}}$
      and $\text{ET-Flow-}SO(3)_{\text{reproduced}}$ denote results reproduced using
      the official ET-Flow~\cite{hassan2024flow} implementation; reproduction
      details are provided in the ``Reproduction Issues of ET-Flow'' subsection
      of the Supplementary Information. For the
      standard fixed-budget setting, bold and underlined values indicate the
      best and second-best results, respectively. \ours-$SO(3)_{\text{Reflow}}$
      denotes the optional Reflow~\cite{liu2022flow} variant applied to \oursregular{}
      for one-step generation. \textsc{EnergySel} denotes an additional energy-based
      selection setting in which $3K$ candidate conformations are generated and
      the $2K$ conformations with the lowest learned energy are retained. These results
      evaluate the post-generation selection utility of the learned energy scores
      and are reported separately from the standard fixed-budget comparison.
   \end{minipage}
   \vspace{-0.4cm}
\end{table*}

\begin{table*}
   [t]
   \caption{Molecular conformer generation results on GEOM-Drugs ($\delta = 0.75\text{\AA}$).}
   \label{tab:drugs}
   \centering
   \begin{adjustbox}
      {max width=0.95\textwidth}
      \renewcommand{\arraystretch}{1.1}
      \begin{tabular}{lccccccccc}
         \toprule \multirow{2}{*}{Method}                                          & \multirow{2}{*}{ODE steps} & \multicolumn{2}{c}{COV-R $\uparrow$} & \multicolumn{2}{c}{AMR-R $\downarrow$} & \multicolumn{2}{c}{COV-P $\uparrow$} & \multicolumn{2}{c}{AMR-P $\downarrow$} \\
         \cmidrule(lr){3-4}\cmidrule(lr){5-6}\cmidrule(lr){7-8}\cmidrule(lr){9-10} &                            & mean                                 & median                                 & mean                                 & median                                & mean               & median             & mean                & median              \\
         \midrule GeoDiff                                                          & 1000                       & 42.10                                & 37.80                                  & 0.835                                & 0.809                                 & 24.90              & 14.50              & 1.136               & 1.090               \\
         GeoMol                                                                    & --                         & 44.60                                & 41.40                                  & 0.875                                & 0.834                                 & 43.00              & 36.40              & 0.928               & 0.841               \\
         Torsional Diff.                                                           & 20                         & 72.70                                & 80.00                                  & 0.582                                & 0.565                                 & 55.20              & 56.90              & 0.778               & 0.729               \\
         MCF-S (13M)                                                               & 1000                       & 79.4                                 & 87.5                                   & 0.512                                & 0.492                                 & 57.4               & 57.6               & 0.761               & 0.715               \\
         MCF-B (64M)                                                               & 1000                       & \underline{84.0}                     & \underline{91.5}                       & 0.427                                & 0.402                                 & 64.0               & 66.2               & 0.667               & 0.605               \\
         MCF-L (242M)                                                              & 1000                       & \textbf{84.7}                        & \textbf{92.2}                          & \textbf{0.390}                       & \textbf{0.247}                        & 66.8               & 71.3               & 0.618               & 0.530               \\
         ET-Flow (8.3M)                                                            & 50                         & 79.53                                & 84.57                                  & 0.452                                & 0.419                                 & 74.38              & 81.04              & 0.541               & 0.470               \\
         $\text{ET-Flow}_{\text{reproduced}}$ (8.3M)                               & 50                         & 79.54                                & 85.00                                  & 0.470                                & 0.444                                 & 69.79              & 75.53              & 0.604               & 0.538               \\
         ET-Flow-SS (8.3M)                                                         & 50                         & 79.62                                & 84.63                                  & 0.439                                & 0.406                                 & \underline{75.19}  & \underline{81.66}  & \underline{0.517}   & \textbf{0.442}      \\
         ET-Flow-$SO(3)$ (9.1M)                                                    & 50                         & 78.18                                & 83.33                                  & 0.480                                & 0.459                                 & 67.27              & 71.15              & 0.637               & 0.567               \\
         $\text{AvgFlow}_{\text{NequIP}}$ (4.7M)                                   & 102                        & 76.8                                 & 83.6                                   & 0.523                                & 0.511                                 & 60.6               & 63.5               & 0.706               & 0.670               \\
         $\text{AvgFlow}_{\text{DiT}}$ (52M)                                       & 100                        & 82.0                                 & 86.7                                   & 0.428                                & 0.401                                 & 72.9               & 78.4               & 0.566               & 0.506               \\
         $\text{AvgFlow}_{\text{DiT-L}}$ (64M)                                     & 100                        & 82.0                                 & 87.3                                   & \underline{0.409}                    & \underline{0.381}                     & \textbf{75.7}      & \textbf{81.9}      & \textbf{0.516}      & \underline{0.456}   \\
         \ours (16.6M)                                                             & 5                          & 77.2                                 & 82.3                                   & 0.499                                & 0.479                                 & 70.0               & 76.5               & 0.607               & 0.541               \\
         \ours (16.6M)                                                             & 50                         & 78.8                                 & 84.6                                   & 0.475                                & 0.455                                 & 70.7               & 76.9               & 0.590               & 0.521               \\
         \midrule \multicolumn{10}{l}{\emph{Energy-based selection}}                \\
         \ours-\textsc{EnergySel} (16.6M)                                          & 5                          & 74.2                                 & 77.5                                   & 0.527                                & 0.504                                 & 74.7               & 82.8               & 0.655               & 0.483               \\
         \midrule \multicolumn{10}{l}{\textbf{\emph{2-step Generation}}}            \\
         MCF-B (64M)                                                               & 2                          & 46.7                                 & 42.4                                   & 0.790                                & 0.791                                 & 21.5               & 13.2               & 1.155               & 0.715               \\
         MCF-L (242M)                                                              & 2                          & 54.2                                 & 54.4                                   & 0.752                                & 0.746                                 & 25.7               & 18.8               & 1.119               & 1.115               \\
         ET-Flow (8.3M)                                                            & 2                          & \underline{73.2}                     & \underline{76.6}                       & \underline{0.577}                    & \underline{0.563}                     & \underline{63.8}   & \underline{67.9}   & \underline{0.681}   & \underline{0.643}   \\
         $\text{ET-Flow}_{\text{reproduced}}$ (8.3M)                               & 2                          & 72.3                                 & 76.9                                   & 0.592                                & 0.583                                 & 58.3               & 60.4               & 0.733               & 0.699               \\
         $\text{AvgFlow}_{\text{NequIP-Rreflow}}$ (4.7M)                           & 2                          & 64.2                                 & 67.7                                   & 0.663                                & 0.661                                 & 43.1               & 38.9               & 0.871               & 0.853               \\
         $\text{AvgFlow}_{\text{DiT-Reflow}}$ (52M)                                & 2                          & \textbf{75.7}                        & \textbf{81.8}                          & \textbf{0.545}                       & \textbf{0.533}                        & 57.2               & 59.0               & 0.748               & 0.705               \\
         \ours (16.6M)                                                             & 2                          & 70.7                                 & 74.6                                   & 0.596                                & 0.578                                 & \textbf{69.1}      & \textbf{75.7}      & \textbf{0.623}      & \textbf{0.575}      \\
         \midrule \multicolumn{10}{l}{\emph{Energy-based selection}}                \\
         \ours-\textsc{EnergySel} (16.6M)                                          & 2                          & 70.1                                 & 74.2                                   & 0.602                                & 0.579                                 & 72.2               & 78.8               & 0.589               & 0.529               \\
         \midrule \multicolumn{10}{l}{\textbf{\emph{1-step Generation}}}            \\
         MCF-B (64M)                                                               & 1                          & 22.1                                 & 6.9                                    & 0.962                                & 0.967                                 & 7.6                & 1.5                & 1.535               & 1.541               \\
         MCF-L (242M)                                                              & 1                          & 27.2                                 & 13.6                                   & 0.932                                & 0.928                                 & 8.9                & 2.9                & 1.511               & 1.514               \\
         ET-Flow (8.3M)                                                            & 1                          & 27.6                                 & 8.8                                    & 0.996                                & 1.006                                 & 25.7               & 5.8                & 0.939               & 0.929               \\
         $\text{ET-Flow}_{\text{reproduced}}$ (8.3M)                               & 1                          & 14.0                                 & 0.0                                    & 1.116                                & 1.142                                 & 10.7               & 0.0                & 1.122               & 1.125               \\
         $\text{AvgFlow}_{\text{NequIP-(Reflow+Distill)}}$ (4.7M)                  & 1                          & \underline{55.6}                     & \underline{56.8}                       & \underline{0.739}                    & \underline{0.734}                     & 36.4               & 30.5               & 0.912               & 0.888               \\
         $\text{AvgFlow}_{\text{DiT-(Reflow+Distill)}}$ (52M)                      & 1                          & \textbf{76.8}                        & \textbf{82.8}                          & \textbf{0.548}                       & \textbf{0.541}                        & \textbf{61.0}      & \textbf{64.0}      & \textbf{0.720}      & \textbf{0.675}      \\
         \ours (16.6M)                                                             & 1                          & 53.1                                 & 50.00                                  & 0.802                                & 0.773                                 & \underline{54.3}   & \underline{56.3}   & \underline{0.773}   & \underline{0.743}   \\
         $\ours_{\text{Reflow}}$ (16.6M)                                           & 1                          & 74.3                                 & 80.0                                    & 0.566                                & 0.548                                 & 60.6                & 63.5                & 0.719               & 0.666               \\
         \midrule \multicolumn{10}{l}{\emph{Energy-based selection}}                \\
         \ours-\textsc{EnergySel} (16.6M)                                          & 1                          & 54.4                                 & 52.4                                   & 0.802                                & 0.781                                 & 61.5               & 66.9               & 0.718               & 0.679               \\
         \bottomrule
      \end{tabular}
   \end{adjustbox}
   \begin{minipage}{0.95\textwidth}
      \scriptsize \textbf{Table notes.} Results of most baseline methods are
      taken from the corresponding literature. The one-step and two-step ET-Flow
      results are taken from Ref.~\cite{cao2025efficient}. $\text{ET-Flow}_{\text{reproduced}}$
      denotes results reproduced using the official ET-Flow~\cite{hassan2024flow}
      implementation; reproduction details are provided in the ``Reproduction
      Issues of ET-Flow'' subsection of the Supplementary Information.
      For the standard fixed-budget setting, bold and underlined values indicate
      the best and second-best results, respectively. $\ours_{\text{Reflow}}$
      denotes the optional Reflow~\cite{liu2022flow} variant applied to \oursregular{}
      for one-step generation. \textsc{EnergySel} denotes an additional energy-based
      selection setting in which $3K$ candidate conformations are generated and
      the $2K$ conformations with the lowest learned energy are retained. These results
      evaluate the post-generation selection utility of the learned energy scores
      and are reported separately from the standard fixed-budget comparison.
   \end{minipage}
   \vspace{-0.3cm}
\end{table*}

\begin{table*}
   [t]
   \caption{Ground-state conformation prediction on GEOM-Drugs. EnFlow improves single-conformation
   prediction through energy-guided sampling and further reduces prediction error
   through ensemble-based learned-energy selection. All metrics are reported in \AA.}
   \centering
   \label{tab:geom_drugs_grpund_state}
   \begin{adjustbox}
      {max width=0.95\textwidth}
      \begin{tabular}{lccccc}
         \toprule Method                                  & Inference mode              & Steps & $\mathbf{D}\text{-MAE}\downarrow$ & $\mathbf{D}\text{-RMSE}\downarrow$ & $\mathcal{C}\text{-RMSD}\downarrow$ \\
         \midrule RDKit-DG                                & --                          & 1     & 1.181                             & 2.132                              & 2.097                               \\
         RDKit-ETKDG                                      & --                          & 1     & 1.120                             & 2.055                              & 1.934                               \\
         GINE                                             & --                          & 1     & 1.125                             & 1.777                              & 2.033                               \\
         GATv2                                            & --                          & 1     & 1.042                             & 1.662                              & 1.901                               \\
         GraphGPS (RW)                                    & --                          & 1     & 0.879                             & 1.399                              & 1.768                               \\
         GraphGPS (LP)                                    & --                          & 1     & 0.815                             & 1.300                              & 1.698                               \\
         GTMGC                                            & --                          & 1     & 0.823                             & 1.319                              & 1.458                               \\
         Torsional Diffusion                              & --                          & 20    & 0.959                             & 1.648                              & 1.751                               \\
         ReBind                                           & --                          & 1     & 0.776                             & \underline{1.283}                  & 1.396                               \\
         ET-Flow                                          & --                          & 5     & 0.844$\pm$0.009                   & 1.491$\pm$0.020                    & 1.633$\pm$0.011                     \\
         \midrule \ours                                   & \texttt{JustFM}             & 5     & 0.793$\pm$0.011                   & 1.412$\pm$0.024                    & 1.550$\pm$0.029                     \\
         \midrule \ours                                   & \texttt{EnergyRank} ($M=5$)  & 5     & 0.745                             & 1.361                              & 1.422                               \\
         \ours                                            & \texttt{EnergyRank} ($M=5$)  & 50    & 0.714                             & 1.296                              & 1.338                               \\
         \ours                                            & \texttt{EnergyRank} ($M=50$) & 5     & \underline{0.703}                 & 1.331                              & \underline{1.312}                   \\
         \ours                                            & \texttt{EnergyRank} ($M=50$) & 50    & \textbf{0.644}                    & \textbf{1.263}                     & \textbf{1.163}                      \\
         \midrule Relative improvement over previous best & --                          & --    & 17.01\%                           & 1.56\%                             & 16.69\%                             \\
         \bottomrule
      \end{tabular}
   \end{adjustbox}
   \vspace{2pt}
   \begin{minipage}{0.95\textwidth}
      \scriptsize \textbf{Table notes.} Results of Torsional Diffusion are taken
      from Ref.~\cite{kim2025rebind}. Results of ET-Flow and \oursregular{} (\texttt{JustFM})
      are reported as mean $\pm$ standard deviation over 10 independent runs, each
      generating one conformation per molecule. Because single-sample generative
      outputs are not explicitly selected by energy, their error with respect to
      the ground-state label is reported here as a reference comparison.

      For \ours, two inference modes are considered. \texttt{JustFM} generates
      one conformation per molecule using energy-guided sampling. \texttt{EnergyRank}
      generates $M$ conformations per molecule and selects the candidate with the
      lowest predicted energy. For \texttt{JustFM}, the guidance strength is set
      to $\lambda_{t}= 0.5(1-t)^{2}$ to favor precise generation. For \texttt{EnergyRank},
      a smaller guidance strength $\lambda_{t}= 0.2(1-t)^{2}$ is used to
      preserve ensemble diversity for learned-energy selection.
   \end{minipage}
   \vspace{-0.5cm}
\end{table*}

\subsection{Energy-guided flow matching improves few-step conformer generation}

We next evaluate whether energy-guided flow matching improves conformer generation,
especially when only a small number of ODE sampling steps is allowed. In \ours,
samples from the \emph{Harmonic Prior} $p_{0}(\mathcal{C}_{0})$ are transported toward
an energy-guided target distribution,
$p^{\prime}_{1}(\mathcal{C}) \propto p_{1}(\mathcal{C}_{1})e^{-J_{\phi}(\mathcal{C}_{1})}$,
rather than the original data distribution $p_{1}(\mathcal{C}_{1})$. This
encourages sampling trajectories to avoid high-energy regions of conformational
space and to concentrate on structurally plausible, low-energy conformations.

We report conformer generation results under the standard fixed-budget protocol,
where each method outputs $2K$ conformations for evaluation. To further assess whether
the learned energy function can support post-generation selection, we also
include an energy-based selection setting, denoted as \textsc{EnergySel}. In this
setting, \ours first generates $3K$ candidate conformations and then retains the
$2K$ candidates with the lowest learned energy. \textsc{EnergySel} is reported
separately from the standard fixed-budget comparison and is used to evaluate the
selection utility of the learned energy scores.

\paragraph{Results on GEOM-QM9.}

The GEOM-QM9 conformer generation results are summarized in~\autoref{tab:qm9}. Under
the standard fixed-budget protocol, \oursregular-$SO(3)$ achieves competitive
performance with substantially fewer sampling steps than many existing baselines.
With 5 ODE steps, it approaches the performance of multi-step baselines; with 50
ODE steps, it maintains strong Recall-oriented performance while improving
Precision-oriented metrics.

The advantages of energy guidance become more evident as the sampling budget
decreases. With 2 ODE steps, \oursregular-$SO(3)$ maintains competitive Recall-oriented
performance and improves Precision-oriented metrics, including COV-P and AMR-P.
In the 1-step setting, \oursregular-$SO(3)$ continues to generate high-quality conformations,
whereas unguided flow-matching baselines degrade markedly. These results indicate
that the learned energy model provides an effective inductive bias for low-step conformer
generation, where sampling trajectories must reach plausible conformational
regions with limited numerical integration.

The \textsc{EnergySel} setting further shows that the learned energy function can
be used for post-generation selection. On GEOM-QM9, selecting low-energy candidates
improves Precision-oriented metrics across 1-step, 2-step, and multi-step settings.
The accompanying reduction in Recall-oriented metrics reflects the expected
diversity--quality trade-off: energy-based filtering preferentially retains
lower-energy conformations while narrowing ensemble diversity.

Applying Reflow~\cite{liu2022flow} further improves one-step generation, showing
that energy-guided flow matching remains compatible with post-training acceleration
techniques. Coverage-curve ablations in Supplementary Fig.~S9
also show that energy guidance improves coverage, especially at smaller RMSD thresholds
and under 1--2 ODE steps.

\paragraph{Results on GEOM-Drugs.}

We further evaluate conformer generation on the more challenging GEOM-Drugs
dataset, with results shown in~\autoref{tab:drugs}. Under the standard fixed-budget
protocol, \oursregular achieves competitive performance relative to existing diffusion-
and flow-based methods. With 50 ODE steps, it maintains a favorable Recall--Precision
balance and improves over the reproduced ET-Flow baseline. With only 5 ODE steps,
\oursregular remains competitive while requiring fewer sampling steps than several
classical baselines, including GeoDiff, GeoMol, and Torsional Diffusion.

The few-step results again highlight the role of energy guidance. With 2 ODE steps,
\oursregular improves Precision-oriented metrics relative to ET-Flow and AvgFlow
variants while maintaining competitive Recall-oriented performance. In the 1-step
setting, \oursregular improves generation quality relative to unguided ET-Flow, indicating
that energy-guided flow matching remains effective under highly constrained sampling
budgets.

Consistent with the GEOM-QM9 results, \textsc{EnergySel} improves Precision-oriented
metrics on GEOM-Drugs, particularly in the 1-step and 2-step settings. These gains
show that the learned energy model can serve as an effective filter for retaining
higher-quality conformations from a generated candidate pool. The corresponding
decrease in Recall-oriented metrics reflects the same diversity--quality trade-off,
as the selected ensemble is concentrated toward lower-energy regions.

Applying Reflow to \oursregular yields further improvements and produces results
comparable to strong AvgFlow variants, while using fewer model parameters (16.6M
vs.\ 52M) and requiring no additional distillation stage. Ablation results in Supplementary Fig.~S10
show that energy guidance consistently improves coverage across sampling budgets,
with the largest gains in the 1--2 step regime.

Overall, results on GEOM-QM9 and GEOM-Drugs show that energy-guided flow matching
improves conformer generation under limited sampling budgets, with the most
consistent gains appearing in Precision-oriented metrics. The \textsc{EnergySel}
results further indicate that the learned energy function supports not only
trajectory guidance during sampling, but also energy-based selection of generated
conformers.

\subsection{Learned energy enables ground-state identification from generated
ensembles}

We next evaluate whether the learned energy model can identify low-energy
ground-state candidates from generated conformational ensembles. Unlike
deterministic structure predictors that directly regress a single target
conformation, \ours approaches ground-state identification through energy-guided
generation followed by learned-energy ranking. This formulation allows the same
framework to support both ensemble generation and ground-state candidate
selection.

The ground-state prediction results on GEOM-Drugs are summarized in
\autoref{tab:geom_drugs_grpund_state}. Among existing methods, ReBind serves as
the strongest baseline for this task.

We first evaluate single-conformation generation under identical sampling budgets.
In this setting, one conformation is generated per molecule without ensemble-based
selection. With 5 sampling steps, ET-Flow achieves $\mathbf{D}\text{-MAE}=0.844\pm
0.009~\text{\AA}$, $\mathbf{D}\text{-RMSE}=1.491\pm 0.020~\text{\AA}$, and
$\mathcal{C}\text{-RMSD}=1.633\pm0.011~\text{\AA}$. Under the same setting,
\oursregular{} (\texttt{JustFM}) improves all three metrics to
$\mathbf{D}\text{-MAE}=0.793\pm0.011~\text{\AA}$,
$\mathbf{D}\text{-RMSE}=1.412\pm0.024~\text{\AA}$, and
$\mathcal{C}\text{-RMSD}=1.550\pm0.029~\text{\AA}$. These improvements indicate
that energy-guided sampling alone shifts generation toward lower-energy and more
ground-state-like conformations, even without explicit ensemble ranking.

We then evaluate the proposed \texttt{EnergyRank} inference scheme, which
generates multiple candidate conformations and selects the structure with the
lowest learned energy. This strategy further reduces prediction errors and
consistently improves performance as the ensemble size increases. With $M=50$
generated conformations and 50 sampling steps, \ours achieves the strongest
overall performance across all compared methods, reaching
$\mathbf{D}\text{-MAE}=0.644~\text{\AA}$,
$\mathbf{D}\text{-RMSE}=1.263~\text{\AA}$, and
$\mathcal{C}\text{-RMSD}=1.163~\text{\AA}$.

Relative to the strongest previous baseline, these results correspond to
improvements of $17.01\%$ in $\mathbf{D}\text{-MAE}$, $1.56\%$ in
$\mathbf{D}\text{-RMSE}$, and $16.69\%$ in $\mathcal{C}\text{-RMSD}$.
Importantly, these gains are achieved without introducing a separate ground-state
prediction model. Instead, the same learned energy function used during conformer
generation is reused to rank and select generated candidates.

Supplementary Fig.~S11 further analyzes the effect of the
ensemble size $M$. Increasing $M$ consistently improves $\mathbf{D}\text{-MAE}$,
$\mathbf{D}\text{-RMSE}$, and $\mathcal{C}\text{-RMSD}$, with particularly
pronounced gains when increasing $M$ from 1 to 20 under 1--5 ODE sampling steps.
Notably, sufficiently large ensembles generated with very few ODE steps can
approach the accuracy obtained with substantially longer sampling trajectories.
This result suggests that the learned energy function is highly discriminative
and can effectively prioritize near-optimal conformations within generated
ensembles.

Overall, these results show that the learned energy model serves two
complementary roles within the same framework: it guides sampling toward
low-energy conformational regions during generation and subsequently enables
learned-energy-based selection of candidate structures. This unified formulation
connects conformer generation and ground-state identification within a single
energy-guided generative framework.

\begin{table}[t]
   \centering
   \caption{\textbf{Alignment between learned energy scores and single-point
   GFN2-xTB energies.} Statistics are computed over 100 GEOM-Drugs molecules.
   Energy values are relative xTB energies within each generated ensemble in
   kcal/mol, with lower values indicating lower-energy conformations. Gains are
   computed per molecule before taking the median.}
   \label{tab: energy_xtb_alignment}
   \scriptsize
   \setlength{\tabcolsep}{3pt}
   \renewcommand{\arraystretch}{0.88}
   \begin{tabular}{lc}
      \toprule
      Metric & Value \\
      \midrule
      \multicolumn{2}{l}{\textbf{Rank correlation}} \\
      Median Spearman $\rho$        & 0.819 \\
      Mean Spearman $\rho$          & 0.794 \\
      Std. Spearman $\rho$          & 0.113 \\
      Min / Max Spearman $\rho$     & 0.359 / 0.972 \\
      Molecules with $\rho > 0.5$   & 98.0\% \\
      Molecules with $\rho > 0.8$   & 56.0\% \\
      \midrule
      \multicolumn{2}{l}{\textbf{Top-$k$ selection energy}} \\
      Learned-energy top-$k$        & 2.311 \\
      Random top-$k$                & 16.576 \\
      Median per-molecule gain      & 14.819 \\
      \midrule
      \multicolumn{2}{l}{\textbf{Best conformer in top-$k$}} \\
      Learned-energy top-$k$        & 0.000 \\
      Random top-$k$                & 2.988 \\
      Median per-molecule gain      & 2.812 \\
      \bottomrule
   \end{tabular}
\end{table}

\begin{figure}[t]
   \refstepcounter{figure}
   \centering
   \includegraphics[width=\textwidth]{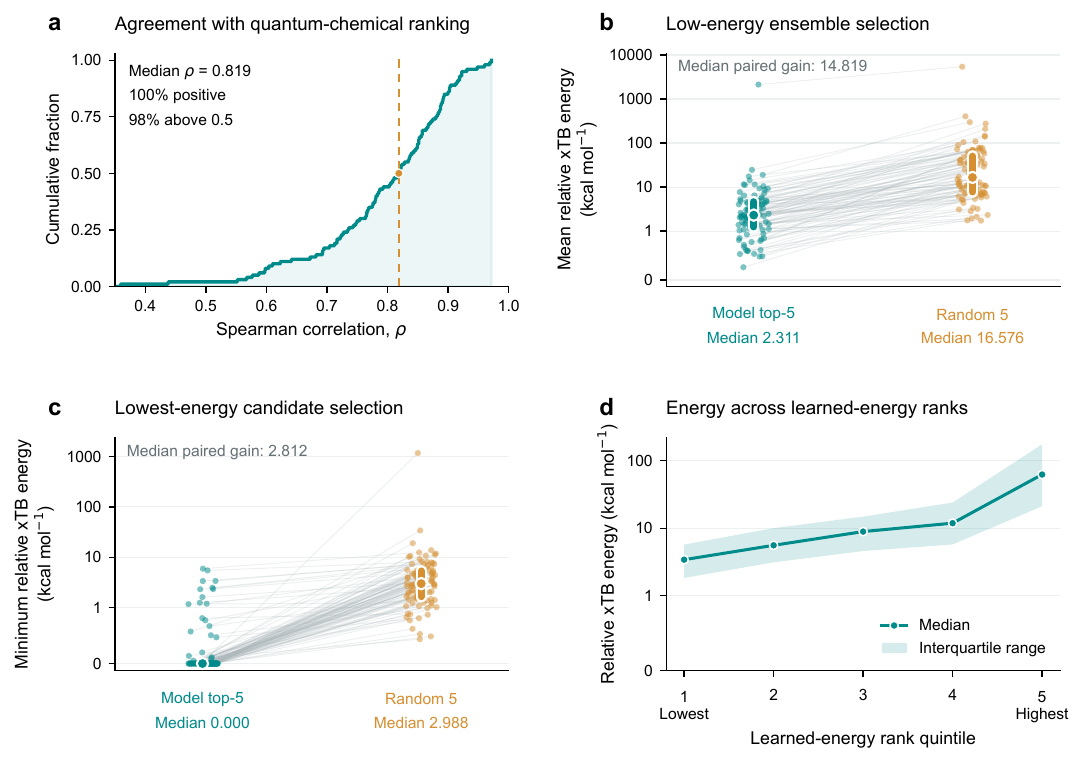}
   \par\smallskip
   \textbf{Figure~\thefigure}
   \label{fig: energy_xtb_alignment}
\end{figure}

\begin{figure}[t]
   \refstepcounter{figure}
   \centering
   \includegraphics[width=\textwidth]{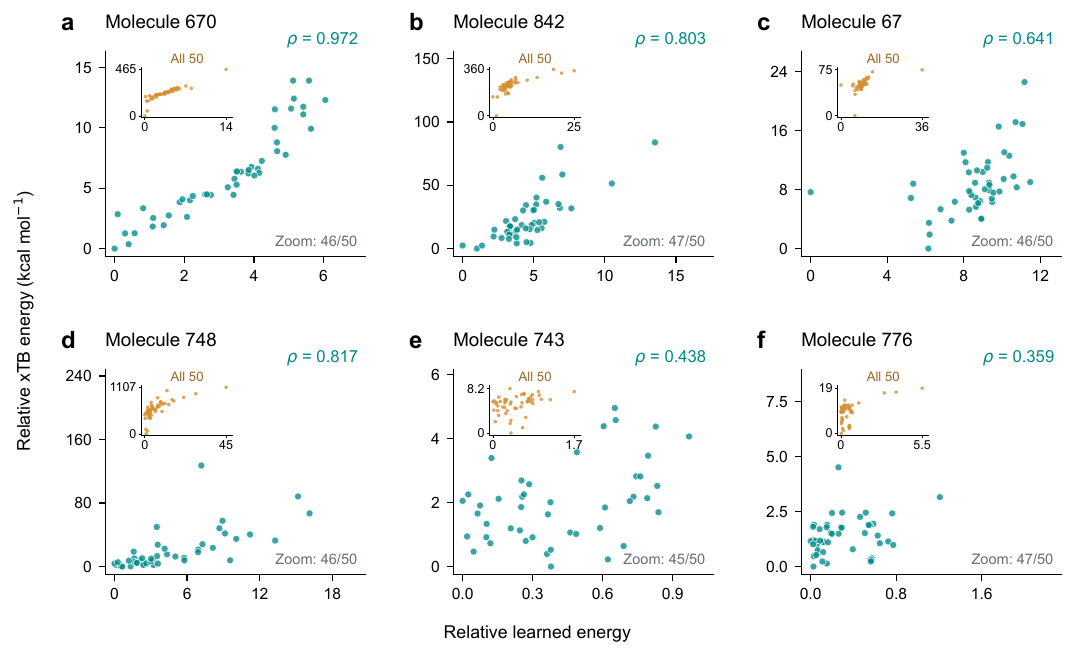}
   \par\smallskip
   \textbf{Figure~\thefigure}
   \label{fig: energy_xtb_scatter}
\end{figure}

\subsection{Learned energy scores align with quantum-chemical energetic rankings}

A central question is whether the learned energy model provides physically
meaningful energetic information, rather than serving only as an internal scoring
function for the generative model. To evaluate this, we compare learned energy scores
with single-point GFN2-xTB energies computed on generated conformations. We
randomly sample 100 molecules from the GEOM-Drugs test set and generate 50
conformations for each molecule using \ours. For every generated conformation,
we record the learned energy score $J_{\phi}(\mathcal{C})$ and compute the
corresponding xTB energy without geometry relaxation. Because absolute energy scales
differ across molecules, all analyses use relative energies obtained by
subtracting the minimum energy within each generated ensemble.

As shown in~\autoref{fig: energy_xtb_alignment}(a), the learned energy scores
exhibit strong rank-level agreement with xTB energies across molecules. The median
Spearman correlation is $0.819$, all molecules show positive correlation, and $98
.0\%$ of molecules achieve $\rho > 0.5$ (\autoref{tab: energy_xtb_alignment}).
These results indicate that the learned energy model consistently preserves the
relative energetic ordering of generated conformations, which is essential for energy-guided
ranking and selection.

As shown in~\autoref{fig: energy_xtb_alignment}(b), the median per-molecule mean
relative xTB energy among the model-selected top-five conformations is
$2.311~\mathrm{kcal/mol}$, compared with $16.576~\mathrm{kcal/mol}$ for random selection,
with a median per-molecule gain of $14.819~\mathrm{kcal/mol}$. Thus, learned-energy
ranking favors lower-energy conformations.

As shown in~\autoref{fig: energy_xtb_alignment}(c), the minimum relative xTB energy
within the model-selected top-five conformations has a median of
$0.000~\mathrm{kcal/mol}$, compared with $2.988~\mathrm{kcal/mol}$ for the
trial-averaged random-selection minima, supporting the retention of low-energy
candidates.

Finally, we analyze the global relationship between learned-energy rank and quantum-chemical
energy. As shown in~\autoref{fig: energy_xtb_alignment}(d), the median of the
per-molecule mean relative xTB energies increases monotonically across the five
learned-energy rank quintiles. The molecule-level examples
in~\autoref{fig: energy_xtb_scatter} further illustrate this relationship and the
variation in the strength of rank agreement across individual molecules.

Together, these results show that the learned energy model captures physically
meaningful energetic ordering of generated conformations. This provides direct
support for its dual role in \ours: guiding sampling toward low-energy regions
and enabling learned-energy-based selection and ground-state identification of
generated molecular structures.

\subsection{Additional Results and Analysis}

Additional experimental results and analyses are provided in the ``More Experimental
Results'' subsection of the Supplementary Information.
These results further examine the design choices and physical behavior of \ours.
In particular, we analyze the guidance schedule $\lambda_{t}$ in the ``Observations
on the Guidance Schedule'' subsection of the Supplementary Information,
study the contribution of \emph{Energy Matching} training in
the ``Why Energy-Matching Training Is Necessary for the Energy Model'' subsection
of the Supplementary Information, and evaluate the robustness of the energy-guided
vector field relative to alternative formulations in the ``Ablation Studies on
the Choice of the Vector Field for Sampling'' subsection of the Supplementary
Information. We also assess ensemble-level property consistency in the
``Energy-guided sampling improves ensemble-level property consistency'' subsection
of the Supplementary Information,
showing that energy-guided sampling improves agreement between generated and
reference ensembles in Boltzmann-weighted molecular properties, including energy,
dipole moment, highest occupied molecular orbital--lowest unoccupied
molecular orbital (HOMO--LUMO) gap, and minimum energy. We additionally provide
mechanistic visualizations of the learned energy landscape and guided sampling
behavior in the ``Energy Landscape Analysis and Guided Conformational Sampling''
subsection of the Supplementary Information. We further examine molecular
factors associated with learned-energy/xTB rank correlation in the ``Molecular
Factors Affecting Learned-Energy/xTB Rank Correlation'' subsection of the
Supplementary Information, evaluate zero-shot out-of-distribution generalization
in the ``Zero-Shot Out-of-Distribution Evaluation'' subsection of the Supplementary
Information, and report sampling efficiency and energy-guidance overhead in the
``Sampling Efficiency and Energy-Guidance Overhead'' subsection of the Supplementary
Information. Together, these analyses characterize the
reliability, generalization, and practical efficiency trade-offs of
learned-energy guidance.

   \section{Discussion}

Although the proposed framework achieves strong performance in \emph{molecular
conformation generation}, \emph{ground-state conformation prediction}, and
energy-based conformer ranking, several limitations remain. First, \oursregular enables
high-quality few-step sampling by steering \emph{harmonic-prior} samples toward
the energy-guided distribution
$p_{1}(\mathcal{C}_{1})e^{-J_{\phi}(\mathcal{C}_1)}$ rather than the original
distribution $p_{1}(\mathcal{C}_{1})$. However, inference is slower than in
unguided flow-matching frameworks such as ET-Flow~\cite{hassan2024flow}, because
each ODE step requires evaluating energy gradients. Second, although the learned
energy landscape improves ground-state conformation identification, inference
remains slower than in deterministic predictors, since flow-matching and diffusion-based
methods require iterative sampling rather than a single forward evaluation.
Third, the use of a non-Gaussian prior, namely the \emph{Harmonic Prior}, makes
classical Gaussian-path guidance techniques inapplicable. Our energy-guided sampling
therefore relies on an approximate non-Gaussian guidance strategy~\cite{feng2025guidance},
whose theoretical optimality and controllability remain limited. Finally,
although the framework jointly models conformational ensembles, learned energy landscapes,
and ground-state candidates, task-specific performance and computational efficiency
can be further improved. Future work will focus on improving the efficiency, robustness,
and theoretical understanding of energy-guided generative dynamics.

   \section{Conclusions}

This work addresses the challenge of low-energy molecular structure discovery by
coupling conformational generation with explicit energetic evaluation. Existing
approaches typically treat these components separately, either producing diverse
conformations without reliable energy assessment or predicting a single lowest-energy
structure without modeling ensemble variability. To address this limitation, we introduce
a unified framework that couples flow-based generative modeling with explicit energy
landscape learning. The proposed energy-guided flow matching scheme supports low-energy
conformer generation from a harmonic prior, while the learned energy model provides
a principled mechanism for evaluating, selecting, and ranking generated structures.
Experiments on GEOM-QM9 and GEOM-Drugs show that the framework improves few-step
conformer generation and ground-state identification. Single-point GFN2-xTB evaluations
further show that the learned energy scores preserve physically meaningful energetic
rankings of generated conformations. Together, these results support explicit energy
landscape modeling as a promising strategy for low-energy molecular structure discovery
through the joint modeling of conformational ensembles and their associated energies.

   \section{Methods}

\subsection{Overview of \ours}

\ours{} couples flow-based conformer generation with explicit molecular energy landscape
learning. The framework consists of three components: (i) a flow-matching generator
that transports samples from a molecular prior to conformational data, (ii) a
learned energy model that represents conformational energy variations, and (iii)
an energy-guided inference procedure that uses the learned energy both to steer sampling
trajectories and to rank generated structures. This formulation enables
conformer generation, energy-based selection, and learned-energy-based
ground-state identification within a single computational pipeline.

Throughout this section, we use $\mathcal{C}\in\mathbb{R}^{n\times 3}$ to denote
a molecular conformation and $\mathcal{G}$ to denote the corresponding molecular
graph. The vector field is parameterized by $v_{\theta}$, and the learned energy
model is denoted by $J_{\phi}(\mathcal{C})$.

\subsection{Flow-matching backbone for conformer generation}

We use ET-Flow~\citep{hassan2024flow} as the unguided flow-matching backbone for
molecular conformer generation. ET-Flow employs a \emph{Schrödinger Bridge
Conditional Flow Matching} (SB-CFM) path~\citep{tong2024improving} to transport
samples from a non-Gaussian molecular prior, namely the \emph{Harmonic Prior}~\citep{stark2023harmonic,jing2023eigenfold},
to molecular conformations. General background on conditional flow matching is provided
in the ``Conditional Flow Matching'' subsection of the Supplementary Information.

\paragraph{Harmonic Prior.}
The Harmonic Prior encodes spatial proximity between atom positions connected by
covalent bonds. Given the molecular graph $\mathcal{G}$ with adjacency matrix $A\in
[0,1]^{n\times n}$, let $L=D-A$ be the graph Laplacian, where $D$ is the degree matrix.
The prior is defined as
\begin{equation}
    p_{0}(\mathcal{C}_{0}) \propto
    \exp\left[-\frac{1}{2}\operatorname{tr}(\mathcal{C}_{0}^{T}L\mathcal{C}_{0})\right].
\end{equation}

\paragraph{SB-CFM path.}
The SB-CFM path is defined through an entropy-regularized optimal transport
coupling $p(z)=\pi_{2\sigma^2}(\mathcal{C}_{0},\mathcal{C}_{1})$, where $\pi_{2\sigma^2}$
denotes the coupling between prior samples $\mathcal{C}_{0}$ and data conformations
$\mathcal{C}_{1}$ with entropy regularization parameter $\sigma^{2}$~\citep{cuturi2013sinkhorn,tong2024improving}.
The corresponding conditional probability path is
\begin{equation}
    p_{t}(\mathcal{C}_{t}\mid \mathcal{C}_{0}, \mathcal{C}_{1}) = \mathcal{N}\left
    ( \mathcal{C}_{t}; (1-t)\mathcal{C}_{0}+t\mathcal{C}_{1}, t(1-t)\sigma^{2}I \right
    ), 
\end{equation}
and the associated conditional vector field is
\begin{equation}
    v_{t}(\mathcal{C}_{t}\mid \mathcal{C}_{0}, \mathcal{C}_{1}) = \frac{1-2t}{2t(1-t)}
    \left\{ \mathcal{C}_{t}-[(1-t)\mathcal{C}_{0}+t\mathcal{C}_{1}] \right\} +(\mathcal{C}
    _{1}-\mathcal{C}_{0}). 
\end{equation}
The parameterized vector field $v_{\theta}(\mathcal{C}_{t},t)$ is trained with the
conditional flow matching loss
\begin{equation}
    \mathcal{L}_{\mathrm{SB-CFM}}= \mathbb{E}_{\mathcal{C}_{t}\sim p_{t}(\mathcal{C}_{t}\mid
    \mathcal{C}_{0},\mathcal{C}_{1}),\, t\sim\mathcal{U}(0,1)}\left[ \left\| v_{\theta}
    (\mathcal{C}_{t},t)- v_{t}(\mathcal{C}_{t}\mid\mathcal{C}_{0},\mathcal{C}_{1}
    ) \right\|^{2}\right]. 
\end{equation}
After training, conformations are generated by solving the ODE
$d\mathcal{C}_{t}=v_{\theta}(\mathcal{C}_{t},t)\,dt$ with
$\mathcal{C}_{0}\sim p_{0}(\mathcal{C}_{0})$.

\subsection{Learning an explicit molecular energy landscape}

To couple conformer generation with energetic evaluation, \ours{} learns an energy
function $J_{\phi}(\mathcal{C})$ over molecular conformations. The energy model
defines a Boltzmann distribution
\begin{equation}
    p_{\phi}(\mathcal{C}) \propto \exp[-J_{\phi}(\mathcal{C})], 
\end{equation}
where lower-energy conformations are assigned higher probability.

\paragraph{Energy Matching.}
We train the energy model using the Energy Matching objective~\citep{balcerak2025energy},
which connects energy-based model training with optimal transport through the
first-order optimality conditions of the Jordan--Kinderlehrer--Otto scheme. For
paired samples $\mathcal{C}_{0}\sim p_{0}(\mathcal{C}_{0})$ and
$\mathcal{C}_{1}\sim p_{1}(\mathcal{C}_{1})$, define the deterministic
interpolation
\begin{equation}
    \mathcal{C}_{t}^{\prime}
    =
    (1-t)\mathcal{C}_{0}+t\mathcal{C}_{1}.
\end{equation}
The Energy Matching loss encourages the negative energy gradient to align with
the transport direction from the prior sample to the data conformation:
\begin{equation}
    \mathcal{L}_{\mathrm{EM}}
    =
    \mathbb{E}_{\mathcal{C}_{0}\sim p_{0},\,
    \mathcal{C}_{1}\sim p_{1},\,t\sim\mathcal{U}(0,1)}
    \left[
    \left\|
    -\nabla_{\mathcal{C}_{t}^{\prime}}
    J_{\phi}(\mathcal{C}_{t}^{\prime})
    -
    (\mathcal{C}_{1}-\mathcal{C}_{0})
    \right\|^{2}
    \right].
\end{equation}
This objective shapes the energy landscape so that its negative gradient points
from prior samples toward data conformations along transport paths, providing
off-manifold guidance for sampling. However, Energy Matching alone does not
explicitly calibrate relative energies among conformations of the same molecule,
which are needed for energy-based ranking and ground-state identification.

\paragraph{Energy fine-tuning.}
To resolve molecule-specific energy differences, we further fine-tune
$J_{\phi}$ using conformation-level energy annotations from the GEOM
dataset~\citep{axelrod2022geom}. Instead of relying on contrastive-divergence
training with Markov chain Monte Carlo (MCMC)-based negative sampling, we directly optimize the energy model
with supervised energy regression. Specifically, we normalize energies on a
per-molecule basis and minimize the mean absolute error between predicted and
reference energies:
\begin{equation}
    \mathcal{L}_{\mathrm{energy}}
    =
    \mathbb{E}_{\mathcal{C}\sim p_{\mathrm{data}}(\mathcal{C})}
    \left[
    \left|
    J_{\phi}(\mathcal{C})-E_{\mathrm{true}}(\mathcal{C})
    \right|
    \right].
\end{equation}
The fine-tuning objective combines Energy Matching with supervised energy
regression:
\begin{equation}
    \mathcal{L}_{\mathrm{fine}}
    =
    \mathcal{L}_{\mathrm{EM}}
    +
    \gamma_{\mathrm{energy}}\mathcal{L}_{\mathrm{energy}},
    \label{eq:fine_tune_loss}
\end{equation}
where $\gamma_{\mathrm{energy}}$ controls the weight of the energy regression
term. This fine-tuning step improves the ability of the energy model to capture
subtle energy variations across conformations of the same molecule.
\subsection{Joint training with the flow-matching generator}

The flow-matching generator and the energy model are optimized jointly. At each training
step, we sample $\mathcal{C}_{0}\sim p_{0}(\mathcal{C}_{0})$ and
$\mathcal{C}_{1}\sim p_{1}(\mathcal{C}_{1})$, construct the noisy SB-CFM sample
\begin{equation}
    \mathcal{C}_{t}= \mathcal{C}_{t}^{\prime}+ \sigma\sqrt{t(1-t)}\,\epsilon, \qquad
    \epsilon\sim\mathcal{N}(0,I), 
\end{equation}
and train $v_{\theta}$ using $\mathcal{L}_{\mathrm{SB-CFM}}$ while training $J_{\phi}$
using $\mathcal{L}_{\mathrm{EM}}$. In the energy fine-tuning phase, we retain $\mathcal{L}
_{\mathrm{EM}}$ and add the supervised energy regression term in~\autoref{eq:fine_tune_loss}.
The overall training procedure is summarized in Supplementary Algorithm~S1. This
joint optimization allows the vector field and energy model to learn compatible generative
and energetic structures.

\subsection{Energy-guided sampling and energy-based selection}

After training, the learned energy model is used to guide sampling. The goal is
to transport prior samples toward an energy-guided target distribution
\begin{equation}
    p_{1}^{\prime}(\mathcal{C}) \propto p_{1}(\mathcal{C})\exp[-J_{\phi}(\mathcal{C}
    )], 
\end{equation}
which shifts generated conformations toward lower-energy regions of the learned
energy landscape.

Because the Harmonic Prior is non-Gaussian, classical Gaussian-path guidance does
not directly apply. Following approximate guidance for non-Gaussian flow-matching
paths~\citep{feng2025guidance}, summarized in
the ``Approximate Guidance for Non-Gaussian Flow-Matching Paths'' subsection of
the Supplementary Information, we use the guided vector field
\begin{equation}
    v_{t}^{\prime}(\mathcal{C}_{t}) \approx v_{\theta}(\mathcal{C}_{t},t) - \lambda
    _{t}\nabla_{\hat{\mathcal{C}}_{1}}J_{\phi}(\hat{\mathcal{C}}_{1}), \qquad \hat
    {\mathcal{C}}_{1}\approx \mathcal{C}_{t}+(1-t)v_{\theta}(\mathcal{C}_{t},t),
    \label{eq:guided_vector_field}
\end{equation}
where $\lambda_{t}$ is a time-dependent guidance schedule that decays toward zero
as $t\rightarrow 1$. Sampling is performed by numerical integration:
\begin{equation}
    \mathcal{C}_{t+\Delta t}= \mathcal{C}_{t}+ v_{t}^{\prime}(\mathcal{C}_{t})\Delta
    t, \qquad \mathcal{C}_{0}\sim p_{0}(\mathcal{C}_{0}). 
\end{equation}
This procedure follows the learned generative dynamics while biasing generated conformations
toward lower-energy regions according to $J_{\phi}$. The complete energy-guided
sampling procedure is summarized in Supplementary Algorithm~S2.

\paragraph{Energy-based selection.}
The learned energy model can also be used after sampling to select conformations
from a generated candidate pool. In the \textsc{EnergySel} setting, we first
generate $3K$ candidate conformations and retain the $2K$ conformations with the
lowest learned energy. This setting is reported separately from the standard fixed-budget
protocol and evaluates the post-generation selection utility of the learned energy
scores. The procedure is summarized in Supplementary Algorithm~S3.

\paragraph{Reflow.}
For one-step generation, we additionally consider Reflow~\citep{liu2022flow}, a
post-training technique that improves single-step sampling quality. Following AvgFlow~\citep{cao2025efficient},
we apply Reflow along the SB-CFM path. Details are provided in the Supplementary
Information.

\subsection{Ground-state conformation identification by learned-energy ranking}

The learned energy model enables ground-state conformation identification by ranking
generated structures. We consider two inference modes, \texttt{JustFM} and
\texttt{EnergyRank}, as summarized in Supplementary Algorithm~S4.

\paragraph{\texttt{JustFM}.}
This mode generates a single conformation per molecule using the energy-guided sampling
procedure in Supplementary Algorithm~S2. The learned energy model
affects the trajectory during sampling, but no ensemble ranking is performed.

\paragraph{\texttt{EnergyRank}.}
This mode first generates an ensemble of $M$ conformations for a given molecular
graph using Supplementary Algorithm~S2, and then selects the conformation
with the lowest predicted energy:
\begin{equation}
    \hat{\mathcal{C}}^{*}= \arg\min_{\hat{\mathcal{C}}^{m}\in \{\hat{\mathcal{C}}^{1},\ldots,\hat{\mathcal{C}}^{M}\}}
    J_{\phi}(\hat{\mathcal{C}}^{m}). 
\end{equation}
This inference mode uses the same learned energy model for both generation and
learned-energy-based ground-state identification.

\subsection{Model architecture and implementation}

For a fair comparison with ET-Flow~\citep{hassan2024flow}, we use $\mathtt{TorchMD}
\text{-}\mathtt{NET}$~\citep{tholke2022torchmd} as the backbone architecture for
both the vector field $v_{\theta}$ and the energy model $J_{\phi}$. Given a molecular
graph $\mathcal{G}$ and conformation $\mathcal{C}$,
$\mathtt{TorchMD}\text{-}\mathtt{NET}$ produces scalar features
$x\in\mathbb{R}^{n\times d}$ and vector features
$\Vec{v}\in\mathbb{R}^{n\times 3}$:
\begin{equation}
    x,\Vec{v}= \mathtt{TorchMD}\text{-}\mathtt{NET}(\mathcal{G},\mathcal{C}). 
\end{equation}
For the energy model, we apply mean pooling to the scalar features followed by a
linear layer:
\begin{equation}
    J_{\phi}(\mathcal{C}) = \operatorname{MeanPooling}(x,\mathrm{dim}=0)\cdot W,
    \qquad W\in\mathbb{R}^{d\times 1}. 
\end{equation}
For the vector field, we directly use the vector features as the output, $v_{\theta}
=\Vec{v}$. Additional architectural details, hyperparameters, guidance schedules,
reproduction details, algorithms, and single-point GFN2-xTB evaluation settings are
provided in the Supplementary Information.
Experiments were conducted using Python 3.10.18, PyTorch 2.6.0+cu118 with
CUDA 11.8, RDKit 2024.3.2, and xTB 6.5.0. Complete software and dependency
versions are provided in the ``Software and Computational Environment'' subsection
of the Supplementary Information.

\subsection{\texorpdfstring{Equivariance properties of the harmonic prior and guided dynamics}{Equivariance properties of the harmonic prior and guided dynamics}}

We close the methodological description by making explicit the symmetry properties
of \ours{} under global rotations. The Harmonic Prior used in the flow-matching
backbone is graph-Laplacian structured, rather than an isotropic Gaussian over all
atom-coordinate degrees of freedom. Nevertheless, because the graph Laplacian acts
on atom indices and the same quadratic form is applied to each Cartesian
coordinate, the prior remains invariant to global rotations of the coordinate
frame. The role of the $SO(3)$-equivariant architecture is therefore to impose the
corresponding covariance on the learned transport field. This covariance is also
preserved by the approximate energy-guided field in
\autoref{eq:guided_vector_field}, provided that the learned energy is
rotation-invariant.

\begin{proposition}[Rotation invariance of the Harmonic Prior]
\label{prop:harmonic_prior_rotation_invariance}
Let $\mathcal{C}\in\mathbb{R}^{n\times 3}$ denote a molecular conformation whose
rows are atomic Cartesian coordinates, and let $L=D-A$ be the graph Laplacian of
an undirected molecular graph. Define the harmonic energy
\begin{equation}
    U_{L}(\mathcal{C})
    =
    \frac{1}{2}\operatorname{tr}(\mathcal{C}^{T}L\mathcal{C})
    =
    \frac{1}{2}\sum_{i<j}A_{ij}\|\mathcal{C}_{i}-\mathcal{C}_{j}\|^{2}.
    \label{eq:harmonic_energy_rotation}
\end{equation}
Then, for any rotation matrix $R\in SO(3)$,
\begin{equation}
    U_{L}(\mathcal{C}R^{T})=U_{L}(\mathcal{C}).
\end{equation}
Consequently, the Harmonic Prior density $p_{0}(\mathcal{C})\propto
\exp[-U_{L}(\mathcal{C})]$ is invariant under global rotations of the molecular
coordinate frame.
\end{proposition}

\begin{proof}
Since $R\in SO(3)$ is orthogonal, $R^{T}R=I$. Using cyclic invariance of the trace,
\begin{equation}
    U_{L}(\mathcal{C}R^{T})
    =
    \frac{1}{2}\operatorname{tr}\!\left((\mathcal{C}R^{T})^{T}L(\mathcal{C}R^{T})\right)
    =
    \frac{1}{2}\operatorname{tr}\!\left(R\mathcal{C}^{T}L\mathcal{C}R^{T}\right)
    =
    \frac{1}{2}\operatorname{tr}\!\left(\mathcal{C}^{T}L\mathcal{C}R^{T}R\right)
    =
    U_{L}(\mathcal{C}).
\end{equation}
Equivalently, the edge-based form in \autoref{eq:harmonic_energy_rotation} depends
only on squared interatomic distances, which are unchanged by a common rotation of
all coordinates. This proves the invariance of both the harmonic energy and the
corresponding prior density.
\end{proof}

\begin{proposition}[Equivariance of the energy-guided vector field]

Assume that the learned transport field is $SO(3)$-equivariant,
\begin{equation}
    v_{\theta}(\mathcal{C}R^{T},t)=v_{\theta}(\mathcal{C},t)R^{T},
    \qquad \forall R\in SO(3),
    \label{eq:vector_field_equivariance}
\end{equation}
and that the learned energy is rotation-invariant,
\begin{equation}
    J_{\phi}(\mathcal{C}R^{T})=J_{\phi}(\mathcal{C}),
    \qquad \forall R\in SO(3).
\end{equation}
Let
\begin{equation}
    g_{\phi}(\mathcal{Y})=\nabla_{\mathcal{Y}}J_{\phi}(\mathcal{Y})
\end{equation}
denote the coordinate gradient of the learned energy at an arbitrary conformation
\(\mathcal{Y}\). For a fixed time \(t\), define the endpoint estimate, consistent
with \autoref{eq:guided_vector_field}, as
$\hat{\mathcal{C}}_{1}(\mathcal{C}_{t})=\mathcal{C}_{t}+(1-t)v_{\theta}(\mathcal{C}_{t},t)$.
The corresponding approximate energy-guided vector field is
\begin{equation}
    v_{t}^{\prime}(\mathcal{C}_{t})
    =
    v_{\theta}(\mathcal{C}_{t},t)
    -
    \lambda_{t}g_{\phi}(\hat{\mathcal{C}}_{1}(\mathcal{C}_{t})).
\end{equation}
Then this guided vector field is also $SO(3)$-equivariant:
\begin{equation}
    v_{t}^{\prime}(\mathcal{C}_{t}R^{T})=v_{t}^{\prime}(\mathcal{C}_{t})R^{T}.
\end{equation}
\end{proposition}

\begin{proof}
The equivariance of $v_{\theta}$ first implies that the endpoint estimate transforms
covariantly under the same rotation:
\begin{equation}
    \hat{\mathcal{C}}_{1}(\mathcal{C}_{t}R^{T})
    =
    \mathcal{C}_{t}R^{T}+(1-t)v_{\theta}(\mathcal{C}_{t}R^{T},t)
    =
    \left[\mathcal{C}_{t}+(1-t)v_{\theta}(\mathcal{C}_{t},t)\right]R^{T}
    =
    \hat{\mathcal{C}}_{1}(\mathcal{C}_{t})R^{T}.
    \label{eq:endpoint_estimate_equivariance}
\end{equation}
The coordinate gradient of a rotation-invariant scalar energy is likewise
equivariant. Indeed, from $J_{\phi}(\mathcal{Y}R^{T})=J_{\phi}(\mathcal{Y})$, the
chain rule gives
\begin{equation}
    g_{\phi}(\mathcal{Y}R^{T})
    =
    g_{\phi}(\mathcal{Y})R^{T}.
    \label{eq:energy_gradient_equivariance}
\end{equation}
Combining \autoref{eq:vector_field_equivariance}, \autoref{eq:endpoint_estimate_equivariance},
and \autoref{eq:energy_gradient_equivariance}, we obtain
\begin{equation}
    \begin{aligned}
    v_{t}^{\prime}(\mathcal{C}_{t}R^{T})
    &=
    v_{\theta}(\mathcal{C}_{t}R^{T},t)
    -
    \lambda_{t}
    g_{\phi}(\hat{\mathcal{C}}_{1}(\mathcal{C}_{t}R^{T})) \\
    &=
    v_{\theta}(\mathcal{C}_{t},t)R^{T}
    -
    \lambda_{t}
    g_{\phi}(\hat{\mathcal{C}}_{1}(\mathcal{C}_{t}))R^{T} \\
    &=
    v_{t}^{\prime}(\mathcal{C}_{t})R^{T}.
    \end{aligned}
\end{equation}
Thus the approximate guidance term preserves the same transformation law as the
learned transport field. Together with the rotation-invariant Harmonic Prior in
Proposition~\ref{prop:harmonic_prior_rotation_invariance}, this establishes that the
energy-guided dynamics used by \ours{} are consistent with the global rotational
symmetry of molecular conformations.
\end{proof}

   \section{Data Availability}

Experiments were conducted using the \textbf{GEOM} dataset~\cite{axelrod2022geom}.
The corresponding dataset record is cited separately~\cite{DVN/JNGTDF_2021}.
The original dataset is publicly available at
\url{https://dataverse.harvard.edu/api/access/datafile/4327252}, and we use the
reorganized version released with Torsional Diffusion~\cite{jing2022torsional},
available at
\url{https://drive.google.com/drive/folders/1BBRpaAvvS2hTrH81mAE4WvyLIKMyhwN7}.

The processed data are available in Zenodo~\cite{xu2026enflowprocessed} at
\url{https://zenodo.org/records/18044125},
and model checkpoints are available in Zenodo~\cite{xu2026enflowcheckpoints} at
\url{https://zenodo.org/records/18045494}.

Source data underlying Fig.~2 are provided as Supplementary Data~1, and
source data underlying Figs.~3 and 4 are provided as Supplementary Data~2.
Any additional data required to reproduce the findings are available from
the corresponding author Yatao Bian (\texttt{ybian@nus.edu.sg}) upon reasonable
request.

\section{Code Availability}

The source code is publicly available at
\url{https://github.com/Rich-XGK/EnFlow}.
Detailed software versions and the model, sampling, and analysis parameters
required to reproduce the study are provided in the Methods and Supplementary
Information.

\section{Acknowledgements}

\paragraph{Funding.}
This work was supported by the National Research Foundation, Singapore (Grant No. NRF-NRFF18-2026-0004);
the Ministry of Education, Singapore (Grant No. T1 251RES2607);
and the NUS Artificial Intelligence Institute (NAII), National University of Singapore (Grant No. NAII-SG-2025-36).

\paragraph{Author contributions.}
G.X., Y.B. and P.Z. conceived and designed the study. G.X. and Y.B. developed the method, G.X. implemented the model, performed the experiments, and analyzed the data. X.Y. and
Z.M. contributed to discussions and manuscript revision. P.Z. and Y.B. supervised
the research. G.X. drafted the manuscript with input from all authors. All authors
reviewed the manuscript and approved the final version.

\section{Competing Interest Declaration}

There are no competing interests.

   \clearpage
   \begingroup
   \small
   \section{Figure legends}

\noindent\textbf{Figure~\ref{fig:task_teaser}. Overview of \ours{} for low-energy
molecular structure discovery.} (a) Molecular conformational energy landscapes,
where low-energy conformers occupy local minima and the ground-state conformation
corresponds to the global minimum. (b) Existing approaches remain fragmented:
generative methods can sample diverse conformations but generally lack explicit
energy-based ranking, whereas deterministic predictors estimate a single low-energy
structure without modeling conformational ensembles. (c) \ours{} couples generative
sampling with a learned explicit energy model, enabling energy-guided conformer
generation and energy-based ranking of generated structures. (d) During training,
flow matching learns transport from the Harmonic Prior to the conformational data
distribution, while the learned energy model reshapes the target distribution toward
lower-energy regions through Energy Matching and supervised energy fine-tuning.
(e) During sampling, energy guidance steers few-step trajectories toward energetically
favorable conformations, improving generation precision under extremely limited ODE
sampling budgets. (f) The learned energy model further enables ensemble-based
ground-state identification by ranking generated conformations according to their
predicted energies and selecting the lowest-energy candidate. The learned energy
scores additionally align with single-point geometry, frequency, noncovalent,
extended tight-binding (GFN2-xTB) energies, supporting their physical relevance for
molecular energy landscape modeling.

\medskip
\noindent\textbf{Figure~\ref{fig: joint_performance}. Joint performance in
conformer generation and ground-state identification on GEOM-Drugs.}
The vertical axis shows mean precision coverage (COV-P; higher is better) at
$\delta=0.75\,\text{\AA}$ from Table~\ref{tab:drugs}, using the standard-budget
setting listed for each baseline and five ODE steps for the \ours variants. The
horizontal axis shows coordinate root mean square deviation
($\mathcal{C}\text{-RMSD}$; lower is better) from
Table~\ref{tab:geom_drugs_grpund_state}: one-step results for GTMGC and ReBind and the
50-step \texttt{EnergyRank} result with $M=50$ for both \ours markers, because
\textsc{EnergySel} changes only post-generation conformer selection.
The upper-right arrow ($\nearrow$) indicates better joint performance: higher COV-P
and lower $\mathcal{C}$-RMSD; the horizontal axis is reversed. Teal and light teal
circles denote methods evaluated only for conformer generation and only for
ground-state identification, respectively; gold markers denote \ours variants
evaluated on both tasks. The gold circle represents \ours, and the gold diamond
represents \ours-\textsc{EnergySel}. Bubble area encodes the reported number of model
parameters. Parameter counts are unavailable for Torsional Diffusion and ReBind;
their uniform minimum-size hollow circles are shown only for visibility and do not represent
parameter-count estimates. Dashed N/A axes indicate that the corresponding metric was
not reported.

\medskip
\noindent\textbf{Figure~\ref{fig: energy_xtb_alignment}. Learned energy scores
recover quantum-chemical energetic rankings and support low-energy conformer
selection.} Analyses use 100 GEOM-Drugs test molecules with 50 generated
conformations each. \textbf{a}, Empirical cumulative distribution of per-molecule
Spearman correlations between learned energy scores and single-point GFN2-xTB
energies; the gold dashed line marks the median. \textbf{b,c}, Mean (\textbf{b})
and minimum (\textbf{c}) relative GFN2-xTB energies within the five conformations
of lowest learned energy (teal), compared with randomly selected sets of five
(gold). For random selection, each statistic is computed within each sampled set
and then averaged over trials. Each strategy has one small point per molecule
($n=100$); grey lines connect the two values for the same molecule. Large circles and
thick bars denote medians and 25th--75th percentiles across molecules. Annotations
show medians and median per-molecule gains (random minus model).
\textbf{d}, Conformations are sorted by learned energy and divided into five
quintiles of ten conformations per molecule. Mean relative GFN2-xTB energy is
computed within each quintile per molecule; points and lines show the median
across molecules, and shading spans the 25th--75th percentiles. GFN2-xTB energies
are relative to the lowest-energy generated conformation within each molecule,
in kcal\,mol$^{-1}$. Panels \textbf{b--d} use symmetric-logarithmic vertical
axes, linear from 0 to 1\,kcal\,mol$^{-1}$; tick labels show original relative
energy values.

\medskip
\noindent\textbf{Figure~\ref{fig: energy_xtb_scatter}. Molecule-level
alignment between learned energy scores and single-point GFN2-xTB energies.}
Scatter plots show six GEOM-Drugs molecules: \textbf{a}, molecule 670,
Spearman $\rho=0.972$; \textbf{b}, molecule 842, $\rho=0.803$; \textbf{c},
molecule 67, $\rho=0.641$; \textbf{d}, molecule 748, $\rho=0.817$; \textbf{e},
molecule 743, $\rho=0.438$; and \textbf{f}, molecule 776, $\rho=0.359$.
Each point represents one generated conformation. Learned and GFN2-xTB energies
are shifted separately so that each has a minimum of zero within each molecule;
GFN2-xTB energies are in kcal\,mol$^{-1}$. Teal main panels use linear axes and
show conformations at or below the 95th percentile of both energies within each
molecule, retaining 46, 47, 46, 46, 45 and 47 points in \textbf{a--f}, respectively.
Gold insets include all 50 conformations, with linear horizontal and
symmetric-logarithmic vertical axes (linear from 0 to 1\,kcal\,mol$^{-1}$);
tick labels show original relative energy values. All Spearman correlations
use the complete set of 50 conformations per molecule.

   \endgroup

   \clearpage
   \appendix
   \section*{Supplementary Information}
   \setcounter{figure}{0}
   \setcounter{table}{0}
   \setcounter{algorithm}{0}
   \setcounter{equation}{0}
   \setcounter{theorem}{0}
   \renewcommand{\thefigure}{S\arabic{figure}}
   \renewcommand{\thetable}{S\arabic{table}}
   \renewcommand{\thealgorithm}{S\arabic{algorithm}}
   \renewcommand{\theequation}{S\arabic{equation}}
   \renewcommand{\theHfigure}{supp.\arabic{figure}}
   \renewcommand{\theHtable}{supp.\arabic{table}}
   \renewcommand{\theHalgorithm}{supp.\arabic{algorithm}}
   \renewcommand{\theHequation}{supp.\arabic{equation}}
   \renewcommand{\theHtheorem}{supp.\arabic{theorem}}
   \renewcommand{\tableautorefname}{Table}
   \section{Related Work \& Dataset Details}

\subsection{Related Work}

\subsubsection{Molecular Conformation Generation}

The task of \textit{molecular conformation generation} entails predicting the 3D
structure of molecules based on their topological graphs. In response to the efficiency
limitations of traditional methods~\citep{rappe1992uff, halgren1996merck, parr1979local},
there has been a notable shift towards data-driven approaches, particularly
those leveraging deep learning techniques. Early works~\citep{simm2019generative, xu2021learning, xu2021end, shi2021learning}
explored the application of VAEs~\citep{kingma2013auto}, Normalizing Flows~\citep{rezende2015variational},
and noise conditional score network (NCSN)~\citep{song2019generative} to generate atomic distance matrices, which
are subsequently converted into 3D coordinates through distance geometry methods~\citep{liberti2014euclidean}.
However, these methods often produce suboptimal performance, yielding low-quality
3D structures unsuitable for practical use.

Although there are other approaches that optimize performance using various
techniques~\citep{ganea2021geomol, guan2021energy}, recent advancements have led to
the emergence of more powerful diffusion models~\citep{sohl2015deep, ho2020denoising, song2020score}.
GeoDiff~\citep{xu2022geodiff} is the first to apply a denoising diffusion
probabilistic model (DDPM)~\citep{ho2020denoising}
for the direct generation of 3D coordinates. TorDiff~\citep{jing2022torsional}
first obtains an RDKit-generated~\citep{rdkit} conformation and then applies DDPM
to its hypertorus surface. MCF~\citep{wang2023swallowing} further improves performance
by scaling a large transformer~\citep{vaswani2017attention} instead of
equivariant GNNs~\citep{thomas2018tensor, satorras2021n, tholke2022torchmd} for model
selection, but it still suffers from sample inefficiency.

To improve sample efficiency, recent ODE-based methods, such as consistency
models~\citep{song2023consistency} and flow matching~\citep{lipmanflow, liu2022flow, lipman2024flow, tong2024improving},
have been applied to generate fast and accurate conformations~\citep{fan2024ec, hassan2024flow, cao2025efficient, xu2025cofm}.
ET-Flow~\citep{hassan2024flow} transports harmonic priors~\citep{stark2023harmonic} to
the conformation space via the Schrodinger Bridge CFM~\citep{tong2024improving},
achieving state-of-the-art performance. AvgFlow~\citep{cao2025efficient} notably
contributes 1(2)-step fast sampling techniques through Reflow and Distillation~\citep{liu2022flow}.

\subsubsection{Molecular Ground-State Conformation Prediction}

Motivated by the observation that the above methods in the ``Molecular Conformation
Generation'' subsection
generate conformations without explicit energy minimization, recent work has shifted
toward deterministic prediction of molecular ground-state conformations. The
work in Ref.~\cite{xu2021molecule3d} first introduced the Molecule3D benchmark,
in which each molecule is paired with a high-level density functional theory (DFT)-optimized ground-state
conformation; however, the proposed task formulation is suboptimal. GTMGC~\citep{xu2023gtmgc}
complements this benchmark with a more suitable task definition and strong
baselines based on Graph Transformers~\cite{ying2021transformers}, and subsequent
works~\cite{luo2024bridging,wangwgformer} further improve performance on Molecule3D,
though their experimental settings lack comparisons to generative methods, limiting
comprehensiveness. ReBind~\cite{kim2025rebind} extends the experiments to the
more comprehensive GEOM-Drugs~\cite{axelrod2022geom} dataset by selecting, for each
molecule, the conformation with the highest Boltzmann weight within its ensemble
as the ground state. It then re-evaluates baselines and extends Torsional Diffusion~\cite{jing2022torsional}
for a fairer and more comprehensive comparison. Our work is orthogonal to these
approaches: we focus on enhancing generative modeling for molecular conformation
generation and subsequently identify the ground state through energy-based selection.
Accordingly, following ReBind~\cite{kim2025rebind}, we evaluate ground-state
conformation prediction on GEOM-Drugs and report baseline results as provided in
ReBind~\cite{kim2025rebind}.

\subsubsection{Guidance in Diffusion and Flow Matching}

Guided generative techniques, such as class-guided~\citep{dhariwal2021diffusion, ho2022classifier}
and energy-guided~\citep{lu2023contrastive} modeling, have garnered significant attention
within the field of generative modeling. These methods aim to generate samples
that align with specific properties or classes, as opposed to merely sampling from
a learned distribution~\citep{rombach2022high, song2023loss}. These techniques have
been well-established in standard diffusion models~\citep{sohl2015deep, song2019generative, ho2020denoising, song2020score},
which have become prominent in recent years~\citep{dhariwal2021diffusion, ho2022classifier, chung2022diffusion, rombach2022high, song2023loss, zheng2024ensemble}.

Given that flow matching~\citep{lipmanflow, liu2022flow, lipman2024flow} has recently
emerged as a more efficient alternative to diffusion models, an increasing
number of studies have begun investigating guidance mechanisms in this setting~\citep{zheng2023guided, kollovieh2024flow, zhang2025energy}.
However, existing works largely consider flow matching with Gaussian priors,
which is conceptually equivalent to diffusion models~\citep{zheng2023guided},
and thus adopt guidance strategies that closely mirror those used in diffusion
models. In contrast, a distinctive advantage of flow matching is its flexibility
in transforming arbitrary prior distributions into target distributions, rather than
being restricted to Gaussian priors as in diffusion models. This flexibility
indicates substantial untapped potential for developing both theoretical and
practical guidance approaches under non-Gaussian priors, beyond the Gaussian-based
paradigm. In this direction, the recent work of Ref.~\cite{feng2025guidance}
provides a theoretical framework that lays the foundation for such exploration.

\subsection{Dataset Details}

In this work, we leverage the \textbf{GEOM}~\cite{axelrod2022geom} dataset for a
comprehensive evaluation. Below, we summarize its key characteristics along four
axes: molecular sources, conformation sources, conformational energy annotations,
and Boltzmann weighting of conformers.

\paragraph{Molecular sources:}
GEOM consolidates drug-like compounds from AICures and multiple MoleculeNet
collections, together with the QM9 small-molecule set. Canonical Simplified
Molecular Input Line Entry System (SMILES) strings are produced with RDKit to merge properties from heterogeneous sources into a single
molecular entry; cluster SMILES are de-salted and, when appropriate, protonated to
standardize ionization states. GEOM also includes the BACE subset from
MoleculeNet with experimental binding affinities, while most other biophysics sets
are omitted for size reasons. Recovery of vacuum conformer–rotamer ensembles
exceeds 98\% for the included MoleculeNet datasets, with water-solvent ensembles
generated for virtually all BACE molecules.

\paragraph{Conformation sources:}
For each molecule, an initial geometry is prepared and optimized before CREST
sampling. CREST (Conformer--Rotamer Ensemble Sampling Tool) is an automated
workflow for exploring low-energy conformational space via metadynamics and
semiempirical quantum chemistry~\cite{pracht2024crest}. Drug-like molecules receive
RDKit embeddings (multiple trial conformers), Merck molecular force field (MMFF) optimization and pruning, followed
by xTB optimization; the lowest-energy xTB conformer seeds CREST. Although QM9 structures
are already DFT-optimized, they are re-optimized with xTB to align with the
level of theory employed by CREST. CREST then performs metadynamics under a
constant-number, constant-volume, constant-temperature (NVT)
thermostat with a history-dependent RMSD-based bias; newly discovered structures
along the trajectory are added as reference conformers. Multiple metadynamics runs
with varied bias parameters further improve the coverage of torsional space while
avoiding bond-breaking events.

\paragraph{Energy of conformations:}
The primary label for large-scale ensembles is the GFN2-xTB energy computed
within CREST. A higher-accuracy route is provided for selected subsets (notably
BACE), where DFT single-point energies and, in some cases, full CENSO
refinements with vibrational and solvation contributions are supplied.

\paragraph{Boltzmann weights:}
Let $\lambda = 1/(k_{\mathrm{B}}T)$. GEOM provides two weighting schemes. In the
\emph{CREST/xTB} protocol, each conformer $i$ is assigned an approximate weight from
its GFN2-xTB energy $E_{i}$ together with an explicit degeneracy factor $d_{i}$ (counting
chemically equivalent rotamers),
\[
    w_{i}^{\mathrm{xTB}}=\frac{d_{i}\,e^{-\lambda E_i}}{\sum_{j}d_{j}\,e^{-\lambda
    E_j}},
\]
which serves as a proxy for free-energy probabilities and omits translational, rotational,
and vibrational contributions. In the \emph{CENSO/DFT} protocol, statistical weights
are computed from conformer-specific free energies,
\[
    G_{i}= E_{\text{gas}}^{(i)}+ \Delta G_{\text{solv}}^{(i)}(T) + G_{\text{trv}}
    ^{(i)}(T), \qquad w_{i}^{\mathrm{DFT}}=\frac{e^{-\lambda G_i}}{\sum_{j}e^{-\lambda
    G_j}},
\]
with no explicit degeneracy factor. For numerical stability, $E_{i}$ or $G_{i}$
can be replaced by relative values (e.g., subtracting the minimum) before
evaluation.

\paragraph{Splitting strategies:}
Following recent work~\cite{hassan2024flow,cao2025efficient} and the splitting protocol
of Refs.~\citep{ganea2021geomol,jing2022torsional}, evaluation is conducted on
two standard GEOM subsets: \textbf{GEOM-Drugs} and \textbf{GEOM-QM9}. GEOM-Drugs
contains approximately 304k drug-like molecules (mean 44 atoms), with train/validation/test
splits of 243{,}473/30{,}433/1{,}000 molecules. GEOM-QM9 comprises approximately
120k small organic molecules (mean 11 atoms), split into 106{,}586/13{,}323/1{,}000
molecules for training/validation/testing. In practice, at most 30 conformers per
molecule are retained by keeping the top-30 ranked by Boltzmann weight for the training
and validation sets; molecules that cannot be processed by RDKit are excluded.

   \section{Supplementary Preliminaries}

This section provides additional background on conditional flow matching and energy-guided
flow matching. These details support the formulation used in the main text, where
only the resulting guided vector field is used.

\subsection{Conditional Flow Matching}

Given a source distribution $p_{0}: \mathbb{R}^{d}\to \mathbb{R}_{\geq 0}$ and a
target distribution $p_{1}: \mathbb{R}^{d}\to \mathbb{R}_{\geq 0}$, probability flow
models~\citep{lipmanflow,lipman2024flow,tong2024improving} define a time-dependent
vector field $v_{t}(x_{t}) : [0,1]\times \mathbb{R}^{d}\to \mathbb{R}^{d}$ that
transports samples from $p_{0}$ to $p_{1}$ along a probability path
$p_{t}(x_{t}) : [0,1]\times \mathbb{R}^{d}\to \mathbb{R}_{\geq 0}$. The path and
vector field satisfy the continuity equation
\begin{equation}
    \frac{\partial p_{t}(x_{t})}{\partial t}+ \nabla \cdot \left( p_{t}(x_{t})v_{t}
    (x_{t}) \right) = 0. 
\end{equation}
Flow Matching (FM)~\citep{lipmanflow,lipman2024flow} learns a parameterized
vector field $v_{\theta}(x_{t},t)$ by minimizing
\begin{equation}
    \mathcal{L}_{\mathrm{FM}}= \mathbb{E}_{t\sim\mathcal{U}(0,1),\,x_t\sim p_t}\left
    [ \left\| v_{\theta}(x_{t},t)-v_{t}(x_{t}) \right\|^{2}\right]. 
\end{equation}
In practice, the marginal vector field $v_{t}(x_{t})$ is generally intractable.
Conditional Flow Matching (CFM) instead defines a conditional probability path
$p_{t}(x_{t}\mid z)$ and a corresponding conditional vector field
$v_{t}(x_{t}\mid z)$, where $z$ denotes the conditioning variable, typically a
source--target pair $(x_{0},x_{1})$~\citep{tong2024improving}. The CFM loss is
\begin{equation}
    \mathcal{L}_{\mathrm{CFM}}= \mathbb{E}_{t\sim\mathcal{U}(0,1),\, x_t\sim p_t(x_t\mid
    z),\,z\sim p(z)}\left[ \left\| v_{\theta}(x_{t},t)-v_{t}(x_{t}\mid z) \right\|
    ^{2}\right]. 
\end{equation}
As shown in Ref.~\cite{lipmanflow}, the gradients of the marginal and conditional
objectives are equivalent:
\begin{equation}
    \nabla_{\theta}\mathcal{L}_{\mathrm{FM}}= \nabla_{\theta}\mathcal{L}_{\mathrm{CFM}}
    . 
\end{equation}
After training with $\mathcal{L}_{\mathrm{CFM}}$, samples are generated by solving
the probability-flow ODE
\begin{equation}
    dx_{t}= v_{\theta}(x_{t},t)\,dt, \qquad x_{0}\sim p_{0}(x_{0}). 
\end{equation}

\subsection{Energy-Guided Flow Matching}

Given an energy function $J(x):\mathbb{R}^{d}\rightarrow\mathbb{R}$, energy-guided
flow matching aims to generate samples from an energy-reweighted target
distribution
\begin{equation}
    p^{\prime}(x) \propto p(x)e^{-J(x)}. 
\end{equation}
This can be achieved by modifying the original vector field $v_{t}(x_{t})$ with
a guidance vector field $g_{t}(x_{t})$:
\begin{equation}
    v_{t}^{\prime}(x_{t}) = v_{t}(x_{t}) + g_{t}(x_{t}). 
\end{equation}
The following result from Ref.~\cite{feng2025guidance} characterizes the
guidance term that transports samples toward the energy-reweighted distribution.

\begin{theorem}[Energy-guided flow matching~\citep{feng2025guidance}]
     Adding $g_{t}(x_{t})$ to the original vector
    field $v_{t}(x_{t})$ yields a guided vector field $v'_{t}(x_{t})$ that generates
    the guided path
    \begin{equation}
        p'_{t}(x_{t}) = \int p_{t}(x_{t}\mid z)p'(z)\,dz,
    \end{equation}
    provided that
    \begin{align}
        g_{t}(x_{t}) & = \int \left( \mathcal{P}\frac{e^{-J(x_{1})}}{Z_{t}(x_{t})}- 1 \right) v_{t\mid z}(x_{t}\mid z) p(z\mid x_{t})\,dz, \label{eq:supp_general_guidance} \\
        Z_{t}(x_{t}) & = \int \mathcal{P}e^{-J(x_{1})}p(z\mid x_{t})\,dz . \label{eq:supp_general_guidance_normalizer}
    \end{align}
    Here $\mathcal{P}=\pi'(x_{0}\mid x_{1})/\pi(x_{0}\mid x_{1})$ denotes the reverse
    coupling ratio. When the coupling is independent, $p(z)=p_{0}(x_{0})p_{1}(x_{1}
    )$, one has $\mathcal{P}=1$.
\end{theorem}

The theorem gives the exact guidance field, but the expression is generally
intractable. The following subsections summarize two commonly used simplifications:
Gaussian-path guidance and approximate non-Gaussian guidance.

\subsection{Guidance for Gaussian-Path Flow Matching}

For independent-coupling Gaussian flow matching, the source and target are
assumed independent, $p(z)=p_{0}(x_{0})p_{1}(x_{1})$, with Gaussian source
distribution $p_{0}(x_{0})=\mathcal{N}(x_{0};\mu,\Sigma)$. This setting is closely
related to diffusion models with different noise schedules~\citep{zheng2023guided,ma2024sit}.
In this case, energy guidance can be expressed in terms of the gradient of the
time-dependent normalization term, yielding guidance forms analogous to those used
in diffusion posterior sampling~\citep{dhariwal2021diffusion,chung2022diffusion,song2023loss}.

One equivalent formulation is
\begin{equation}
    \nabla_{x_{t}}\log p_{t}^{\prime}(x_{t}) = \nabla_{x_{t}}\log p_{t}(x_{t}) -
    \nabla_{x_{t}}J(x_{t}), 
\end{equation}
as discussed in Ref.~\cite{lu2023contrastive}. Another equivalent vector-field
parameterization takes the form
\begin{equation}
    v_{t}^{\prime\prime}(x_{t}) = (1-w)v_{t}(x_{t}) + w\left[ a_{t}x_{t}+ b_{t}\nabla
    _{x_{t}}J(x_{t}) \right], 
\end{equation}
which generates
\begin{equation}
    p^{\prime\prime}(x) \propto p(x)^{1-w}e^{-wJ(x)}, 
\end{equation}
where $w$ is a guidance coefficient and $a_{t},b_{t}$ are time-dependent
schedules ~\citep{zheng2023guided}.

\subsection{Approximate Guidance for Non-Gaussian Flow-Matching Paths}

The molecular conformer generation setting considered in this work uses a non-Gaussian
source distribution, namely the Harmonic Prior. Classical Gaussian-path guidance
therefore does not directly apply. We use the approximate non-Gaussian guidance
strategy of Ref.~\cite{feng2025guidance}, which starts from the exact guidance
expression in~\autoref{eq:supp_general_guidance} and approximates it using the
concentration of $p(x_{1}\mid x_{t})$ around its mean.

First, the normalizer in~\autoref{eq:supp_general_guidance_normalizer} is approximated
by
\begin{equation}
    Z_{t}(x_{t}) = \int \mathcal{P}e^{-J(x_{1})}p(z\mid x_{t})\,dz \approx e^{-J(\hat{x}_{1})}
    , \qquad \hat{x}_{1}= \mathbb{E}_{x_{0},x_{1}\sim p(z\mid x_{t})}[x_{1}]. 
\end{equation}
The guidance field can then be approximated as
\begin{equation}
    g_{t}(x_{t}) \approx - \mathbb{E}_{z\sim p(z\mid x_{t})}\left[ (x_{1}-\hat{x}
    _{1})v_{1\mid t}(x_{t}\mid z) \right] \nabla_{\hat{x}_{1}}J(\hat{x}_{1}), \qquad
    \hat{x}_{1}= \mathbb{E}_{x_{0},x_{1}\sim p(z\mid x_{t})}[x_{1}]. 
\end{equation}

For an affine conditional path
\begin{equation}
    x_{t}= \alpha_{t}x_{1}+ \beta_{t}x_{0}+ \sigma_{t}\epsilon, \qquad \epsilon\sim
    \mathcal{N}(0,I), 
\end{equation}
with small $\sigma_{t}$ and $\dot{\sigma}_{t}$, the conditional mean $\hat{x}_{1}$
can be expressed through the $x_{1}$-parameterization of the vector field~\citep{lipman2024flow,feng2025guidance}:
\begin{equation}
    \hat{x}_{1}\approx - \frac{\dot{\beta}_{t}}{\dot{\alpha}_{t}\beta_{t}-\dot{\beta}_{t}\alpha_{t}}
    x_{t}+ \frac{\beta_{t}}{\dot{\alpha}_{t}\beta_{t}-\dot{\beta}_{t}\alpha_{t}}v
    _{t}(x_{t}). 
\end{equation}
Combining this with
\begin{equation}
    v_{1\mid t}(x_{t}\mid z) = \dot{\alpha}_{t}x_{1}+ \dot{\beta}_{t}x_{0}+ \dot{\sigma}
    _{t}\epsilon
\end{equation}
yields the approximate non-Gaussian guidance form
\begin{equation}
    g_{t}(x_{t}) \approx - \frac{\dot{\alpha}_{t}\beta_{t}-\dot{\beta}_{t}\alpha_{t}}{\beta_{t}}
    \Sigma_{1\mid t}\nabla_{\hat{x}_{1}}J(\hat{x}_{1}), 
\end{equation}
where $\Sigma_{1\mid t}$ denotes the covariance of $p(x_{1}\mid x_{t})$.
Following Ref.~\cite{feng2025guidance}, this covariance term can be absorbed
into a time-dependent scalar guidance schedule $\lambda_{t}$, leading to
\begin{equation}
    g_{t}(x_{t}) \approx - \lambda_{t}\nabla_{\hat{x}_{1}}J(\hat{x}_{1}). 
\end{equation}
The resulting guided vector field is therefore
\begin{equation}
    v_{t}^{\prime}(x_{t}) \approx v_{t}(x_{t}) - \lambda_{t}\nabla_{\hat{x}_{1}}J
    (\hat{x}_{1}), 
\end{equation}
where $\lambda_{t}$ is chosen to decay to zero as $t\to 1$.

In the main method, we instantiate this approximation with molecular
conformations by replacing $x_{t}$ with $\mathcal{C}_{t}$ and using the learned
energy model $J_{\phi}(\mathcal{C})$. This yields the energy-guided vector field
used for conformer generation:
\begin{equation}
    v_{t}^{\prime}(\mathcal{C}_{t}) \approx v_{\theta}(\mathcal{C}_{t},t) - \lambda
    _{t}\nabla_{\hat{\mathcal{C}}_{1}}J_{\phi}(\hat{\mathcal{C}}_{1}), \qquad \hat
    {\mathcal{C}}_{1}\approx \mathcal{C}_{t}+ (1-t)v_{\theta}(\mathcal{C}_{t},t).
\end{equation}
   \section{Proposed Algorithms}

We summarize the main procedures used in \ours. 
\autoref{alg:joint_training} describes the joint training of the flow-matching
generator and the learned energy model. \autoref{alg:energy_guided_sampling}
presents the energy-guided sampling procedure used for conformer generation.
\autoref{alg:energy_selection} describes the energy-based selection procedure
used in the \textsc{EnergySel} setting. Finally,
\autoref{alg:ground_state_certification} summarizes the ground-state conformation
identification procedure.

\begin{algorithm}[t]
    \small
    \caption{Joint Training of the Flow-Matching Generator and Energy Model}
    \label{alg:joint_training}
    \begin{algorithmic}[1]
    \STATE \textbf{Input:} Harmonic Prior $p_0(\mathcal{C}_0)$, conformation data distribution $p_1(\mathcal{C}_1)$, energy model $J_{\phi}(\mathcal{C})$, vector field $v_{\theta}(\mathcal{C}_t,t)$, noise scale $\sigma$, learning rates $\eta_{\mathrm{CFM}}$, $\eta_{\mathrm{EM}}$, $\eta_{\mathrm{energy}}$, and energy loss weight $\eta_{\mathrm{energy}}^{\mathrm{w}}$.
    \STATE \textbf{Output:} Trained vector-field parameters $\theta$ and energy-model parameters $\phi$.

    \STATE \textbf{Matching phase: jointly train $v_{\theta}$ and $J_{\phi}$.}
    \WHILE{not converged}
        \STATE Sample $\mathcal{C}_0 \sim p_0(\mathcal{C}_0)$, $\mathcal{C}_1 \sim p_1(\mathcal{C}_1)$, $t \sim \mathcal{U}(0,1)$, and $\epsilon \sim \mathcal{N}(0,I)$.
        \STATE Compute the interpolation
        $\mathcal{C}'_t=(1-t)\mathcal{C}_0+t\mathcal{C}_1$
        and transport direction
        $S_t=\mathcal{C}_1-\mathcal{C}_0$.
        \STATE Compute the noisy SB-CFM sample
        $\mathcal{C}_t=\mathcal{C}'_t+\sigma\sqrt{t(1-t)}\,\epsilon$.
        \STATE Compute the SB-CFM target vector field
        \[
        v_t(\mathcal{C}_t\mid \mathcal{C}_0,\mathcal{C}_1)
        =
        \frac{1-2t}{2t(1-t)}(\mathcal{C}_t-\mathcal{C}'_t)+S_t .
        \]
        \STATE Compute the flow-matching loss
        \[
        \mathcal{L}_{\mathrm{SB-CFM}}
        =
        \left\|
        v_{\theta}(\mathcal{C}_t,t)
        -
        v_t(\mathcal{C}_t\mid \mathcal{C}_0,\mathcal{C}_1)
        \right\|^2 .
        \]
        \STATE Compute the energy-matching loss
        \[
        \mathcal{L}_{\mathrm{EM}}
        =
        \left\|
        -\nabla_{\mathcal{C}'_t}J_{\phi}(\mathcal{C}'_t)
        -
        S_t
        \right\|^2 .
        \]
        \STATE Update
        $\theta \leftarrow \theta-\eta_{\mathrm{CFM}}\nabla_{\theta}\mathcal{L}_{\mathrm{SB-CFM}}$.
        \STATE Update
        $\phi \leftarrow \phi-\eta_{\mathrm{EM}}\nabla_{\phi}\mathcal{L}_{\mathrm{EM}}$.
    \ENDWHILE

    \STATE \textbf{Energy fine-tuning phase: freeze $v_{\theta}$ and refine $J_{\phi}$.}
    \WHILE{not converged}
        \STATE Sample $\mathcal{C}_0 \sim p_0(\mathcal{C}_0)$, $\mathcal{C}_1 \sim p_1(\mathcal{C}_1)$, $t \sim \mathcal{U}(0,1)$, and data conformation $\mathcal{C}\sim p_{\mathrm{data}}(\mathcal{C})$ with reference energy $E_{\mathrm{true}}(\mathcal{C})$.
        \STATE Compute $\mathcal{C}'_t=(1-t)\mathcal{C}_0+t\mathcal{C}_1$ and $S_t=\mathcal{C}_1-\mathcal{C}_0$.
        \STATE Compute
        \[
        \mathcal{L}_{\mathrm{EM}}
        =
        \left\|
        -\nabla_{\mathcal{C}'_t}J_{\phi}(\mathcal{C}'_t)
        -
        S_t
        \right\|^2 .
        \]
        \STATE Compute the supervised energy loss
        \[
        \mathcal{L}_{\mathrm{energy}}
        =
        \left|
        J_{\phi}(\mathcal{C})-E_{\mathrm{true}}(\mathcal{C})
        \right|.
        \]
        \STATE Update
        \[
        \phi \leftarrow
        \phi
        -
        \eta_{\mathrm{EM}}\nabla_{\phi}\mathcal{L}_{\mathrm{EM}}
        -
        \eta_{\mathrm{energy}}\nabla_{\phi}
        \left(
        \eta_{\mathrm{energy}}^{\mathrm{w}}\mathcal{L}_{\mathrm{energy}}
        \right).
        \]
    \ENDWHILE

    \STATE \textbf{Return:} Trained parameters $\theta$ and $\phi$.
    \end{algorithmic}
\end{algorithm}

\begin{algorithm}[t]
    \small
    \caption{Energy-Guided Sampling for Conformer Generation}
    \label{alg:energy_guided_sampling}
    \begin{algorithmic}[1]
    \STATE \textbf{Input:} Molecular graph $\mathcal{G}$, Harmonic Prior $p_0(\mathcal{C}_0\mid\mathcal{G})$, trained vector field $v_{\theta}(\mathcal{C}_t,t)$, trained energy model $J_{\phi}(\mathcal{C})$, guidance schedule $\lambda_t$, and number of ODE steps $N$.
    \STATE \textbf{Output:} Generated conformation $\hat{\mathcal{C}}$.
    \STATE Set $\Delta t = 1/N$.
    \STATE Sample initial conformation $\mathcal{C}_0 \sim p_0(\mathcal{C}_0\mid\mathcal{G})$.
    \STATE Initialize $\mathcal{C}_t \leftarrow \mathcal{C}_0$.
    \FOR{$i=0$ to $N-1$}
        \STATE Set $t=i\Delta t$.
        \STATE Estimate the endpoint
        \[
        \hat{\mathcal{C}}_1
        =
        \mathcal{C}_t+(1-t)v_{\theta}(\mathcal{C}_t,t).
        \]
        \STATE Compute the guided vector field
        \[
        v'_t(\mathcal{C}_t)
        =
        v_{\theta}(\mathcal{C}_t,t)
        -
        \lambda_t
        \nabla_{\hat{\mathcal{C}}_1}
        J_{\phi}(\hat{\mathcal{C}}_1).
        \]
        \STATE Update
        \[
        \mathcal{C}_t
        \leftarrow
        \mathcal{C}_t+v'_t(\mathcal{C}_t)\Delta t.
        \]
    \ENDFOR
    \STATE \textbf{Return:} $\hat{\mathcal{C}} \leftarrow \mathcal{C}_t$.
    \end{algorithmic}
\end{algorithm}

\begin{algorithm}[t]
    \small
    \caption{Energy-Based Selection for Conformer Generation}
    \label{alg:energy_selection}
    \begin{algorithmic}[1]
    \STATE \textbf{Input:} Molecular graph $\mathcal{G}$, trained energy model $J_{\phi}(\mathcal{C})$, energy-guided sampler in~\autoref{alg:energy_guided_sampling}, reference ensemble size $K$.
    \STATE \textbf{Output:} Selected conformer ensemble $\mathcal{S}_{\mathrm{sel}}$ of size $2K$.
    \STATE Initialize candidate set $\mathcal{S}_{\mathrm{cand}}=\emptyset$.
    \FOR{$j=1$ to $3K$}
        \STATE Generate $\hat{\mathcal{C}}^{(j)}$ using~\autoref{alg:energy_guided_sampling} with molecular graph $\mathcal{G}$.
        \STATE $\mathcal{S}_{\mathrm{cand}}\leftarrow \mathcal{S}_{\mathrm{cand}}\cup\{\hat{\mathcal{C}}^{(j)}\}$.
    \ENDFOR
    \STATE Rank all candidates in $\mathcal{S}_{\mathrm{cand}}$ by learned energy $J_{\phi}(\hat{\mathcal{C}})$ in ascending order.
    \STATE Select the $2K$ candidates with the lowest learned energy to form $\mathcal{S}_{\mathrm{sel}}$.
    \STATE \textbf{Return:} $\mathcal{S}_{\mathrm{sel}}$.
    \end{algorithmic}
\end{algorithm}

\begin{algorithm}[t]
    \small
    \caption{Ground-State Conformation Identification}
    \label{alg:ground_state_certification}
    \begin{algorithmic}[1]
    \STATE \textbf{Input:} Molecular graph $\mathcal{G}$, trained energy model $J_{\phi}(\mathcal{C})$, energy-guided sampler in~\autoref{alg:energy_guided_sampling}, ensemble size $M$, inference mode $\in\{\texttt{JustFM},\texttt{EnergyRank}\}$.
    \STATE \textbf{Output:} Predicted ground-state conformation $\hat{\mathcal{C}}^{*}$.

    \IF{mode is \texttt{JustFM}}
        \STATE Generate one conformation using~\autoref{alg:energy_guided_sampling}:
        \[
        \hat{\mathcal{C}}^{*}
        \leftarrow
        \mathrm{Sample}(\mathcal{G}).
        \]

    \ELSIF{mode is \texttt{EnergyRank}}
        \STATE Initialize candidate set $\mathcal{S}=\emptyset$.
        \FOR{$m=1$ to $M$}
            \STATE Generate $\hat{\mathcal{C}}^{(m)}$ using~\autoref{alg:energy_guided_sampling} with molecular graph $\mathcal{G}$.
            \STATE $\mathcal{S}\leftarrow \mathcal{S}\cup\{\hat{\mathcal{C}}^{(m)}\}$.
        \ENDFOR
        \STATE Select the lowest-energy conformation:
        \[
        m^{*}
        =
        \arg\min_{m\in\{1,\ldots,M\}}
        J_{\phi}(\hat{\mathcal{C}}^{(m)}).
        \]
        \STATE Set $\hat{\mathcal{C}}^{*}\leftarrow \hat{\mathcal{C}}^{(m^{*})}$.
    \ENDIF

    \STATE \textbf{Return:} $\hat{\mathcal{C}}^{*}$.
    \end{algorithmic}
\end{algorithm}
   \section{Model Architectures}

In this section, we present the learnable neural network architecture employed
in this work. To ensure consistency with the ET-Flow framework~\citep{hassan2024flow},
we adopt the architectural design of $\mathtt{TorchMD}\text{-}\mathtt{NET}$ \citep{tholke2022torchmd}.
It should be emphasized that the vast majority of the descriptions and technical
details in this section are adapted from the ET-Flow resources \citep{hassan2024flow},
with minor modifications made to suit our setting. We gratefully
acknowledge the authors of ET-Flow~\citep{hassan2024flow} for their open-source contributions.

\subsection{Architecture}
\begin{figure*}[t]
    \centering
    \makebox[.4\textwidth][c]{(a)}%
    \makebox[.6\textwidth][c]{(b)}%
    \\[0.3em]
    \includegraphics[width=\textwidth]{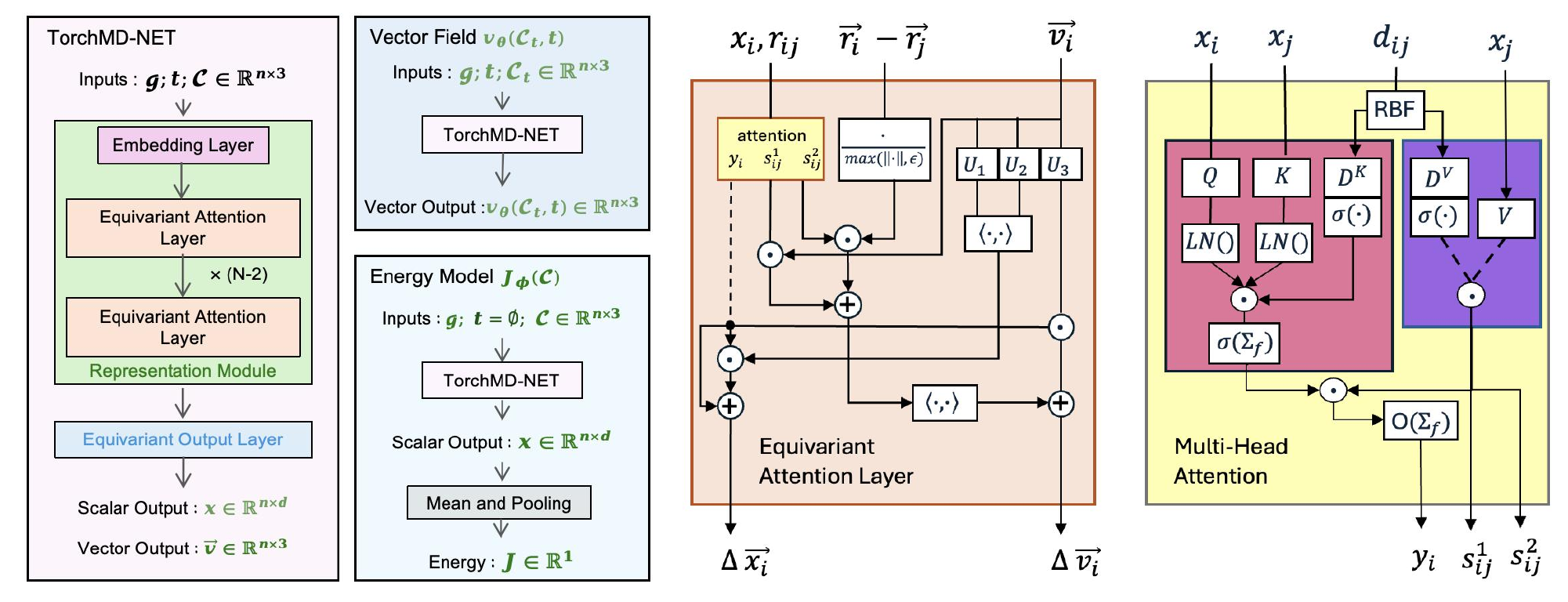}
    \caption{Model architectures in this work. (a) Following ET-Flow~\citep{hassan2024flow},
    the main learnable neural network is $\mathtt{TorchMD}\text{-}\mathtt{NET}$~\citep{tholke2022torchmd},
    modified to incorporate time as an additional input feature, as illustrated
    in the left panel. The right panel shows how $\mathtt{TorchMD}\text{-}\mathtt{NET}$
    is used as both a vector field model and an energy model. (b) Details of the
    \emph{Equivariant Attention Layer} and \emph{Multi-Head Attention}
    components in $\mathtt{TorchMD}\text{-}\mathtt{NET}$, with the illustration adapted
    from the ET-Flow~\citep{hassan2024flow} resources.}
    \label{fig: model_architectures}
\end{figure*}

 The architecture of the modified $\mathtt{TorchMD}\text{-}
\mathtt{NET}$ in \autoref{fig: model_architectures}(a) consists of two major
components, a representation layer and an output layer. For the representation
layer, a modified version of the embedding and equivariant attention-based update
layers from the equivariant transformer architecture of $\mathtt{TorchMD}\text{-}
\mathtt{NET}$ \citep{tholke2022torchmd} is used. The output layer utilizes the
gated equivariant blocks from work~\citep{schutt2021equivariant}. ET-Flow~\citep{hassan2024flow}
has made some modifications to stabilize training since it uses a larger network
than the one proposed in the $\mathtt{TorchMD}\text{-}\mathtt{NET}$ \citep{tholke2022torchmd}
paper. Additionally, since the input structures are interpolations between
structures sampled from a prior and actual conformations, it is important to ensure
the network is numerically stable when the interpolations contain two atoms very
close to each other.

\begin{table*}
    [t]
    \centering
    \small
    \setlength{\tabcolsep}{4pt}
    \begin{tabular}{@{}lll@{}}
        \toprule Name               & Description                 & Range                                                                                                                          \\
        \midrule \texttt{chirality} & Chirality Tag               & \{unspecified, tetrahedral CW \& CCW, other\}                                                                                  \\
        \texttt{degree}             & Number of bonded neighbors  & $\{x:0 \leq x \leq 10, x \in \mathbb{Z}\}$                                                                                     \\
        \texttt{charge}             & Formal charge of atom       & $\{x:-5 \leq x \leq 5, x \in \mathbb{Z}\}$                                                                                     \\
        \texttt{num\_H}             & Total Number of Hydrogens   & $\{x:0 \leq x \leq 8, x \in \mathbb{Z}\}$                                                                                      \\
        \texttt{number\_radical\_e} & Number of Radical Electrons & $\{x:0 \leq x \leq 4, x \in \mathbb{Z}\}$                                                                                      \\
        \texttt{hybridization}        & Hybridization type            & \{sp, sp\textsuperscript{2}, sp\textsuperscript{3}, sp\textsuperscript{3}d, sp\textsuperscript{3}d\textsuperscript{2}, other\} \\
        \texttt{aromatic}           & Whether in an aromatic ring  & \{True, False\}                                                                                                                \\
        \texttt{in\_ring}           & Whether in a ring           & \{True, False\}                                                                                                                \\
        \bottomrule
    \end{tabular}
    \caption{Atomic features included.}
    \label{tab:atomic}
\end{table*}

\paragraph{Embedding Layer:}
The embedding layer maps each atom's physical and chemical properties into a
learned representation space, capturing both local atomic features and geometric
neighborhood information. For the $i$-th atom in a molecule with $N$ atoms, we compute
an invariant embedding $x_{i}$ through a multilayer perceptron (MLP):
\begin{align}
    z_{i}    & = \text{embed}^{\text{int}}(z_{i}) \\
    \
 h_{i} & = \text{MLP}(z_{i})
\end{align}
where $z_{i}$ is the atomic number and $h_{i}$ represents atomic attributes. The
MLP projects atomic attributes into a feature vector of dimension $d_{h}$.

Next, we compute a neighborhood embedding $n_{i}$ that captures the local atomic
environment:
\begin{align}
    n_{i} & = \sum_{j=1}^{N}\text{embed}^{\text{nbh}}(z_{j}) \cdot g(d_{ij}, l_{ij}).
\end{align}
Here, $\text{embed}^{\text{nbh}}(z_{j})$ provides a separate embedding for
neighboring atomic numbers, $d_{ij}$ is the distance between atoms $i$ and $j$,
and $l_{ij}$ encodes edge features (either from a radius-based graph or molecular
bonds). The interaction function $g(d_{ij}, l_{ij})$ combines distance and edge information using
radial basis function (RBF) features:
\begin{align}
    g(d_{ij}, l_{ij}) & = W^{F}\left[\phi(d_{ij})e_{1}^{\text{RBF}}(d_{ij}), \ldots, \phi(d_{ij})e_{K}^{\text{RBF}}(d_{ij}), l_{ij}\right]
\end{align}
where $e_{k}^{\text{RBF}}$ are $K$ exponential radial basis functions following
\citep{unke2019physnet}, and $\phi(d_{ij})$ is a smooth cutoff function:
\begin{align}
    \phi(d_{ij}) & = \begin{cases}\frac{1}{2}\left[\cos\left(\frac{\pi d_{ij}}{d_{\text{cutoff}}}\right)+1\right],&\text{if }d_{ij}\leq d_{\text{cutoff}}\ \\ 0,&\text{otherwise}\end{cases}
\end{align}
Finally, we combine all features into the atom's embedding through a linear
projection:
\begin{equation}
    x_{i}= W^{C}\left[\text{embed}^{\text{int}}(z_{i}), h_{i}, t, n_{i}\right], 
\end{equation}
where $t$ represents the time-step, and $[\cdot,\cdot]$ denotes concatenation. The
resulting embedding $x_{i}\in \mathbb{R}^{d}$ serves as input to subsequent layers
of the network.

\paragraph{Attention Mechanism:}
The multi-head dot-product attention operation uses atom features $x_{i}$, atom attributes
$h_{i}$, time-step $t$ and inter-atomic distances $d_{ij}$ to compute attention
weights. The input atom-level features $x_{i}$ are mixed with the atom attributes
$h_{i}$ and the time-step $t$ using an MLP and then normalized using a LayerNorm~\citep{ba2016layer}.
To compute the attention matrix, the inter-atomic distances $d_{ij}$ are
projected into two filters, $D^{K}$ and $D^{V}$, as:
\begin{align}
    D^{K} & = \sigma \left(W^{D^K}e^{RBF}(d_{ij}) + b^{D^K}\right) \nonumber \\
    D^{V} & = \sigma \left(W^{D^V}e^{RBF}(d_{ij}) + b^{D^V}\right)
\end{align}
The atom level features are then linearly projected along with a LayerNorm operation
to derive the query $Q$ and key $K$ vectors. The value vector $V$ is computed
with only the linear projection of atom-level features. Applying LayerNorm on Q,
K vectors (also referred to as QK-Norm) has proven to stabilize un-normalized
values in the attention matrix~\citep{dehghani2023scaling, esser2024scaling} when
scaling networks to a large number of parameters. The $Q$ and $K$ vectors are then
used along with the distance filter $D^{K}$ for a dot-product operation over the
feature dimension:
\begin{align}
    Q = \text{LayerNorm}(W^{Q}x_{i}), \quad K & = \text{LayerNorm}(W^{K}x_{i}),\quad V = W^{V}x_{i} \\
    \text{dot}(Q, K, D^{K})                   & = \sum_{k}^{F}Q_{k}\cdot K_{k}\cdot D_{k}^{K}.
\end{align}
The attention matrix is derived by passing the above dot-product operation matrix
through a non-linearity and weighting it using a cosine cutoff $\phi(d_{ij})$ (similar
to the embedding layer) which ensures the attention weights are non-zero only when
two atoms are within a specified cutoff:
\begin{align}
    A = \text{SiLU}(\text{dot}(Q, K, D^{K})) \cdot \phi(d_{ij}).
\end{align}
Using the value vector $V$ and the distance filter $D^{V}$, we derive 3 equally sized
filters by splitting along the feature dimension,
\begin{equation}
    \label{eq:s_split}s_{ij}^{1}, s_{ij}^{2}, s_{ij}^{3}= \text{split}(V_{j}\cdot
    D^{V}_{ij}).
\end{equation}
A linear projection is then applied to combine the attention matrix and the
vectors $s_{ij}^{3}$ to derive an atom level feature
$y_{i}= W^{O}\left(\sum_{j}^{N}A_{ij}\cdot s_{ij}^{3}\right)$. The outputs of the
attention operation are $y_{i}$ (an atom level feature) and two scalar filters $s
_{ij}^{1}$ and $s_{ij}^{2}$ (edge-level features).

\paragraph{Update Layer:}
The update layer computes interactions between atoms in the attention block and
uses the outputs to update the scalar feature $x_{i}$ and the vector feature
$\vec{v}_{i}$. First, the scalar feature output $y_{i}$ from the attention
mechanism is split into three features ($q_{i}^{1}, q_{i}^{2}, q_{i}^{3}$), out of
which $q_{i}^{1}$ and $q_{i}^{2}$ are used for the scalar feature update as,
\begin{equation}
    \Delta x_{i}= q_{i}^{1}+ q_{i}^{2}\cdot \langle U_{1}\vec{v}_{i}\cdot U_{2}\vec
    {v}_{i}\rangle,
\end{equation}
where $\langle U_{1}\vec{v}_{i}\cdot U_{2}\vec{v}_{i}\rangle$ is the inner
product between linear projections of vector features $\vec{v}_{i}$ with
matrices $U_{1}, U_{2}$.

The edge vector update consists of two components. First, we compute a vector
$\vec{w}_i$, which for each atom is computed as a weighted sum of vector features
and a clamped-norm of the edge vectors over all neighbors:
\begin{equation}
    \label{eq:update_vec}\vec{w}_{i}= \sum_{j}^{N}s_{ij}^{1}\cdot \vec{v}_{j}+ s_{ij}
    ^{2}\cdot \frac{\vec{r}_{i}- \vec{r}_{j}}{\max(\lVert \vec{r}_{i}- \vec{r}_{j}\rVert,
    \epsilon)},
\end{equation}
\begin{equation}
    \Delta \vec{v}_{i}= \vec{w}_{i}+ q_{i}^{3}\cdot U_{3}\vec{v}_{i}\\
\end{equation}
where $U_{3}$ is a projection matrix over the feature dimension of
the vector feature $\vec{v}_{i}$. In this layer, we clamp the minimum value of
the norm (to $\epsilon = 0.01$) to prevent numerically large values in cases
where positions of two atoms are sampled too close from the prior.

\paragraph{$SO(3)$ Update Layer:}
We also design an $SO(3)$ equivariant architecture by adding an additional cross
product term in \autoref{eq:update_vec} as follows,
\begin{equation}
    \vec{w}_{i}= \sum_{j}^{N}s_{ij}^{1}\cdot \vec{v}_{j}+ s_{ij}^{2}\cdot \frac{\vec{r}_{i}-
    \vec{r}_{j}}{\max(\lVert \vec{r}_{i}- \vec{r}_{j}\rVert, \epsilon)}+ s_{ij}^{4}
    \cdot \left(\vec{v}_{j}\times \frac{\vec{r}_{i}- \vec{r}_{j}}{\max(\lVert
    \vec{r}_{i}- \vec{r}_{j}\rVert, \epsilon)}\right),
\end{equation}
where $s_{ij}^{4}$ is derived by modifying the split operation
\autoref{eq:s_split} in the attention layer where the value vector $V$ and
distance filter $D^{V}$ is projected into 4 equally sized filters instead of 3.

\paragraph{Output Layer:}
The output layer consists of Gated Equivariant Blocks from \citep{schutt2021equivariant}.
Given atom scalar features $x_{i}$ and vector features $\vec{v}_{i}$, the updates in each
block are defined as
\begin{align}
    x_{i, \text{updated}}, \vec{w}_{i} & = \text{split}\left(\text{MLP}\left([x_{i}, \lVert U_{1}\vec{v}_{i}\rVert]\right)\right) \\
    \vec{v}_{i, \text{updated}}        & = (U_{2}\vec{v}_{i}) \cdot \vec{w}_{i}
\end{align}
Here, the norm is taken over Cartesian components. $U_{1}$ and $U_{2}$ are linear projection matrices that act along the feature dimension.
Our modification is to use LayerNorm in the MLP to improve training stability.

\subsection{Input Featurization}

Following ET-Flow~\citep{hassan2024flow}, atomic (node) features are computed using RDKit~\citep{landrum2013rdkit}
descriptors, as detailed in \autoref{tab:atomic}. To construct the edge features
and edge index, we employ a combination of global radius-based edges and local molecular-graph
edges, similar to the approach in Ref.~\citep{jing2022torsional}.

   \section{More Details and Experiments}

\subsection{Metric Definitions}

\paragraph{Conformer ensemble generation metrics.}
Following previous studies~\citep{ganea2021geomol,xu2022geodiff,fan2024ec}, let $\mathsf{R}$
denote the reference set of $K$ conformers,
$\mathsf{R}=\{\mathcal{C}^{1},\mathcal{C}^{2},\ldots,\mathcal{C}^{K}\}$, and let
$\mathsf{G}$ denote the generated set of $2K$ conformers,
$\mathsf{G}=\{\hat{\mathcal{C}}^{1},\hat{\mathcal{C}}^{2},\ldots, \hat{\mathcal{C}}
^{2K}\}$. The root mean square deviation (RMSD) between two conformations
is defined as
\begin{equation}
    \mathrm{RMSD}(\mathcal{C}, \hat{\mathcal{C}}) = \sqrt{\frac{1}{n}\sum_{i=1}^{n}\|\mathcal{C}_{i}-\hat{\mathcal{C}}_{i}\|^{2}}
    . \label{eq: rmsd}
\end{equation}
The Recall-oriented coverage (COV-R) metric is defined as
\begin{equation}
    \mathrm{COV}(\mathsf{R},\mathsf{G}) = \frac{ \left| \left\{ \mathcal{C}^{k}\in\mathsf{R}\mid
    \min_{\hat{\mathcal{C}}^{j}\in\mathsf{G}}\mathrm{RMSD}(\mathcal{C}^{k},\hat{\mathcal{C}}^{j})
    < \delta \right\} \right| }{|\mathsf{R}|}. 
\end{equation}
The Recall-oriented average minimum RMSD (AMR-R) is defined as
\begin{equation}
    \mathrm{AMR}(\mathsf{R},\mathsf{G}) = \frac{1}{|\mathsf{R}|}\sum_{\mathcal{C}^{k}\in\mathsf{R}}
    \min_{\hat{\mathcal{C}}^{j}\in\mathsf{G}}\mathrm{RMSD}(\mathcal{C}^{k},\hat{\mathcal{C}}
    ^{j}). 
\end{equation}
Precision-oriented coverage (COV-P) and precision-oriented average
minimum RMSD (AMR-P) are obtained by swapping the roles of $\mathsf{R}$
and $\mathsf{G}$ in the above definitions.

\paragraph{Ground-state conformation identification metrics.}
Given the ground-state conformation $\mathcal{C}^{*}$ and the predicted
conformation $\hat{\mathcal{C}}^{*}$, we compute coordinate root mean
square deviation ($\mathcal{C}\text{-RMSD}$) using \autoref{eq: rmsd}. Let
$\mathbf{D}^{*}_{ij}$ and $\hat{\mathbf{D}}^{*}_{ij}$
denote the ground-truth and predicted pairwise distances between atoms $i$ and
$j$, respectively. The distance mean absolute error
($\mathbf{D}\text{-MAE}$) is defined as
\begin{equation}
    \mathbf{D}\text{-MAE}(\mathcal{C}^{*},\hat{\mathcal{C}}^{*}) = \frac{1}{n^{2}}
    \sum_{i=1}^{n}\sum_{j=1}^{n}\left| \mathbf{D}^{*}_{ij}-\hat{\mathbf{D}}^{*}_{ij}
    \right|. 
\end{equation}
The distance root mean squared error ($\mathbf{D}\text{-RMSE}$) is
defined as
\begin{equation}
    \mathbf{D}\text{-RMSE}(\mathcal{C}^{*},\hat{\mathcal{C}}^{*}) = \sqrt{ \frac{1}{n^{2}}\sum_{i=1}^{n}\sum_{j=1}^{n}\left(
    \mathbf{D}^{*}_{ij}-\hat{\mathbf{D}}^{*}_{ij}\right)^{2}}. 
\end{equation}

\subsection{Implementation Details}

\subsubsection{\texorpdfstring{Software and Computational Environment}{Software and Computational Environment}}

All experiments were run with Python 3.10.18, PyTorch 2.6.0+cu118
(CUDA 11.8), Lightning 2.5.5, torchmetrics 1.8.2, and PyTorch Geometric
2.6.1. The PyTorch Geometric environment used \texttt{torch-cluster} 1.6.3+pt26cu118,
\texttt{torch-scatter} 2.1.2+pt26cu118, \texttt{torch-sparse} 0.6.18+pt26cu118,
and \texttt{pyg\_lib} 0.4.0.

Molecular processing and analysis used RDKit 2024.3.2, datamol 0.12.5,
NetworkX 3.3, and py3Dmol 2.5.3. Numerical and statistical analyses used NumPy
1.26.4, SciPy 1.15.3, and pandas 2.3.3. Figures were generated using Matplotlib
3.10.7 and Seaborn 0.13.2.

Single-point GFN2-xTB calculations were performed using xTB 6.5.0
(commit \texttt{97a52f5}, conda-forge build compiled on 15 May 2022) with the
\texttt{--gfn 2} option. GFN2-xTB is the GFN2 parameterization implemented in xTB,
rather than a separate software package.

\subsubsection{Reflow Technique}
 Following AvgFlow~\citep{cao2025efficient}, which
adopts a Reflow technique~\citep{liu2022flow} to improve the conformation sampling
quality with a single ODE step, we also implement a Reflow procedure in our
experiments. Specifically, after the main training phase is completed, we first randomly
sample a fixed number of prior samples $\mathcal{C}_{0}^{\prime}$ from the harmonic
prior distribution for each molecule in the training and validation sets. We
then apply a fixed number of ODE steps to generate the corresponding
conformation samples $\mathcal{C}_{1}^{\prime}$. Finally, the coupled pairs
$(\mathcal{C}_{0}^{\prime}, \mathcal{C}_{1}^{\prime})$ are used to further fine-tune
the trained model with the reflow loss:
\begin{equation}
    \mathcal{C}_{t}^{\prime}= (1 - t)\mathcal{C}_{0}^{\prime}+ t\mathcal{C}_{1}^{\prime}
    + \sigma \sqrt{t(1 - t)}\cdot \epsilon, \quad \epsilon \sim \mathcal{N}(0, I)
    .
\end{equation}
\begin{equation}
    v_{t}(\mathcal{C}_{t}^{\prime}| \mathcal{C}_{0}^{\prime}, \mathcal{C}_{1}^{\prime}
    ) = \frac{1 - 2t}{2t(1 - t)}(\sigma \sqrt{t(1 - t)}\cdot \epsilon) + (\mathcal{C}
    _{1}^{\prime}- \mathcal{C}_{0}^{\prime}).
\end{equation}
\begin{equation}
    \mathcal{L}_{\textit{Reflow}}= \mathbb{E}_{\mathcal{C}_t^{\prime} \sim p_t(\mathcal{C}_t^{\prime}
    | \mathcal{C}_0^{\prime}, \mathcal{C}_1^{\prime}), t \sim \mathcal{U}(0, 1)}\left
    [\|v_{\theta}(\mathcal{C}_{t}^{\prime}, t) - v_{t}(\mathcal{C}_{t}^{\prime}|
    \mathcal{C}_{0}^{\prime}, \mathcal{C}_{1}^{\prime})\|^{2}\right].
\end{equation}

It is worth noting that we apply the reflow technique only to the vector field
model $v_{\theta}$, i.e., the unguided path. After reflow fine-tuning, we use the
same guided sampling algorithm described in~\autoref{alg:energy_guided_sampling}
for inference as before, but employing the newly reflowed vector field model.

\subsubsection{Training Details}
For GEOM-Drugs, during the first \textit{matching phase}, \oursregular is trained
for 500 epochs with a batch size of 24, capped at 5{,}000 batches per epoch, on 6
NVIDIA A100 GPUs. During the second \textit{energy fine-tuning phase},
\oursregular is again trained for 500 epochs with a batch size of 48, limited to
5{,}000 batches per epoch, on 4 NVIDIA A100 GPUs. We use the AdamW optimizer with
a cosine-annealed learning rate schedule ranging from $1 \times 10^{-8}$ to $5 \times
10^{-4}$ for both phases. For GEOM-QM9, in the first \textit{matching phase}, \oursregular
is trained for 1{,}000 epochs with a batch size of 48, capped at 1{,}000 batches
per epoch, on 8 NVIDIA A100 GPUs. In the second \textit{energy fine-tuning phase},
\oursregular is trained for 1{,}000 epochs with a batch size of 64, again limited
to 1{,}000 batches per epoch, on 4 NVIDIA A100 GPUs. For GEOM-QM9, we also use
the AdamW optimizer with a cosine-annealed learning rate schedule ranging from
$1 \times 10^{-8}$ to $7 \times 10^{-4}$. For both datasets, the best checkpoint
is selected based on the lowest validation loss.

\subsubsection{Reproduction Issues of ET-Flow}
 As \oursregular builds upon ET-Flow~\cite{hassan2024flow},
we have made every effort to reproduce the reported results of ET-Flow on GEOM-Drugs
and GEOM-QM9. However, as shown in Table~1 and Table~2 of the main manuscript, we observe a noticeable performance gap between our reproduced
results and those originally reported, especially on GEOM-Drugs. Several
relevant issues have been raised in ET-Flow's official GitHub repository (\href{https://github.com/shenoynikhil/ETFlow.git}{https://github.com/shenoynikhil/ETFlow.git}),
and the authors have acknowledged these concerns. We thank the authors for their
helpful communication and support during our reproduction attempts, and we leave
a more detailed investigation of this gap as an open problem for future work.

\subsubsection{Choice of the Guidance Schedule}

As formulated in Eq.~12 of the main manuscript, we perform energy-guided sampling
using the guided vector field $v^{\prime}_{t}(\mathcal{C}_{t})$. The guidance
strength $\lambda_{t}$ is chosen as a time-dependent decay factor that vanishes
as $t \to 1$. In our experiments, we adopt a simple quadratic schedule, as summarized
in \autoref{tab: guidance_schedule}. Further discussion of this choice is provided
in the ``Observations on the Guidance Schedule'' subsection.

\begin{table}[t]
    \centering
    \begin{tabular}{lcccc}
        \hline
                   & 1 ODE step     & 2 ODE steps    & 5 ODE steps    & 50 ODE steps   \\
        \hline
        GEOM-Drugs & $0.5(1-t)^{2}$ & $0.3(1-t)^{2}$ & $0.2(1-t)^{2}$ & $0.1(1-t)^{2}$ \\
        GEOM-QM9   & $0.4(1-t)^{2}$ & $0.3(1-t)^{2}$ & $0.2(1-t)^{2}$ & $0.1(1-t)^{2}$ \\
        \hline
    \end{tabular}
    \caption{Guidance schedule $\lambda_{t}$ under different numbers of ODE
    sampling steps.}
    \label{tab: guidance_schedule}
\end{table}

\subsection{More Experimental Results}

\subsubsection{Observations on the Guidance Schedule}

\begin{figure*}[t]
    \centering
    \makebox[\textwidth]{\footnotesize{(a) 5 ODE steps sampling on GEOM-QM9}}
    \begin{subfigure}
        {\textwidth}
        \centering
        \begin{adjustbox}
            {max width=0.85\textwidth}
            \begin{tabular}{cccccccccc}
                \toprule \multirow{2}{*}{$\lambda_{t}$}                                                      & \multicolumn{2}{c}{COV-R $\uparrow$} & \multicolumn{2}{c}{AMR-R $\downarrow$} & \multicolumn{2}{c}{COV-P $\uparrow$} & \multicolumn{2}{c}{AMR-P $\downarrow$} & $J_{\phi}$ $\downarrow$ \\
                \cmidrule(lr){2-3}\cmidrule(lr){4-5}\cmidrule(lr){6-7}\cmidrule(lr){8-9}\cmidrule(lr){10-10} & mean                                 & median                                 & mean                                 & median                                 & mean                   & median & mean  & median & mean  \\
                \midrule $0.1(1-t)^{2}$                                                                      & 96.11                                & 100.00                                 & 0.084                                & 0.036                                  & 94.82                  & 100.00 & 0.098 & 0.044  & 0.880 \\
                $0.2(1-t)^{2}$                                                                               & 95.97                                & 100.00                                 & 0.083                                & 0.037                                  & 95.22                  & 100.00 & 0.093 & 0.042  & 0.611 \\
                $0.3(1-t)^{2}$                                                                               & 95.53                                & 100.00                                 & 0.084                                & 0.032                                  & 95.56                  & 100.00 & 0.085 & 0.038  & 0.406 \\
                $0.4(1-t)^{2}$                                                                               & 95.64                                & 100.00                                 & 0.083                                & 0.033                                  & 95.83                  & 100.00 & 0.081 & 0.033  & 0.243 \\
                $0.5(1-t)^{2}$                                                                               & 95.01                                & 100.00                                 & 0.086                                & 0.035                                  & 95.86                  & 100.00 & 0.081 & 0.032  & 0.148 \\
                \bottomrule
            \end{tabular}
        \end{adjustbox}
    \end{subfigure}
    \begin{subfigure}
        {\textwidth}
        \centering
        \includegraphics[width=0.9\linewidth]{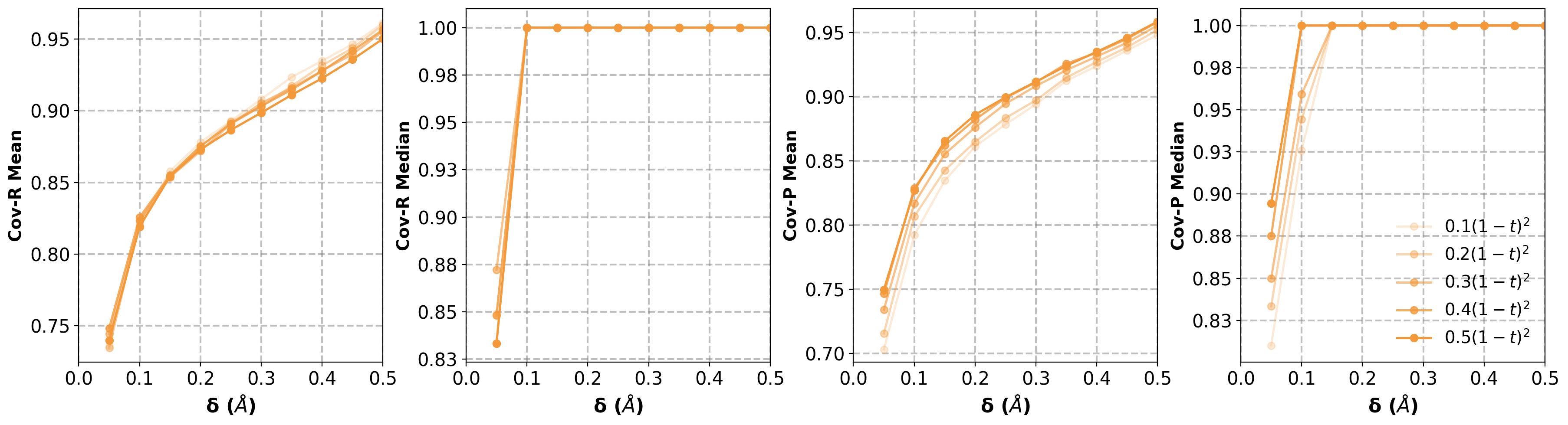}
    \end{subfigure}
    \makebox[\textwidth]{\footnotesize{(b) 50 ODE steps sampling on GEOM-QM9}}
    \begin{subfigure}
        {\textwidth}
        \centering
        \begin{adjustbox}
            {max width=0.85\textwidth}
            \begin{tabular}{cccccccccc}
                \toprule \multirow{2}{*}{$\lambda_{t}$}                                                      & \multicolumn{2}{c}{COV-R $\uparrow$} & \multicolumn{2}{c}{AMR-R $\downarrow$} & \multicolumn{2}{c}{COV-P $\uparrow$} & \multicolumn{2}{c}{AMR-P $\downarrow$} & $J_{\phi}$ $\downarrow$ \\
                \cmidrule(lr){2-3}\cmidrule(lr){4-5}\cmidrule(lr){6-7}\cmidrule(lr){8-9}\cmidrule(lr){10-10} & mean                                 & median                                 & mean                                 & median                                 & mean                   & median & mean  & median & mean   \\
                \midrule $0.1(1-t)^{2}$                                                                      & 95.89                                & 100.00                                 & 0.080                                & 0.032                                  & 95.33                  & 100.00 & 0.085 & 0.034  & 0.166  \\
                $0.2(1-t)^{2}$                                                                               & 95.23                                & 100.00                                 & 0.080                                & 0.031                                  & 94.96                  & 100.00 & 0.084 & 0.029  & 0.097  \\
                $0.3(1-t)^{2}$                                                                               & 95.67                                & 100.00                                 & 0.077                                & 0.030                                  & 95.64                  & 100.00 & 0.078 & 0.028  & 0.051  \\
                $0.4(1-t)^{2}$                                                                               & 95.13                                & 100.00                                 & 0.081                                & 0.030                                  & 95.60                  & 100.00 & 0.076 & 0.026  & 0.009  \\
                $0.5(1-t)^{2}$                                                                               & 95.39                                & 100.00                                 & 0.083                                & 0.031                                  & 95.61                  & 100.00 & 0.078 & 0.031  & -0.023 \\
                \bottomrule
            \end{tabular}
        \end{adjustbox}
    \end{subfigure}
    \begin{subfigure}
        {\textwidth}
        \centering
        \includegraphics[width=0.9\linewidth]{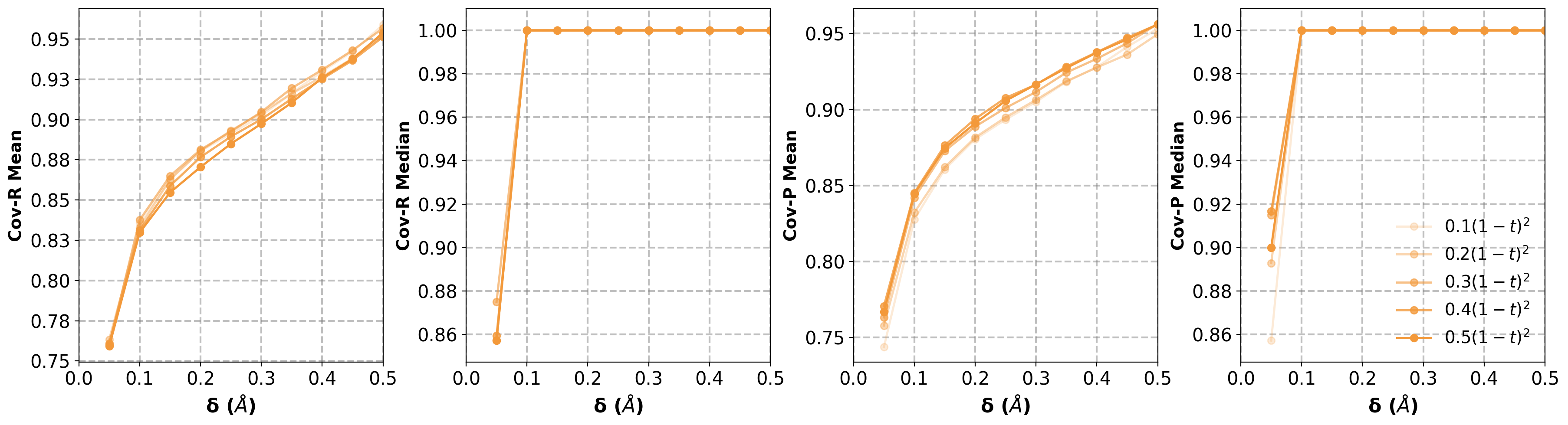}
    \end{subfigure}
    \makebox[\textwidth]{\footnotesize{(c) 5 ODE steps sampling on GEOM-Drugs}}
    \begin{subfigure}
        {\textwidth}
        \centering
        \begin{adjustbox}
            {max width=0.85\textwidth}
            \begin{tabular}{cccccccccc}
                \toprule \multirow{2}{*}{$\lambda_{t}$}                                                      & \multicolumn{2}{c}{COV-R $\uparrow$} & \multicolumn{2}{c}{AMR-R $\downarrow$} & \multicolumn{2}{c}{COV-P $\uparrow$} & \multicolumn{2}{c}{AMR-P $\downarrow$} & $J_{\phi}$ $\downarrow$ \\
                \cmidrule(lr){2-3}\cmidrule(lr){4-5}\cmidrule(lr){6-7}\cmidrule(lr){8-9}\cmidrule(lr){10-10} & mean                                 & median                                 & mean                                 & median                                 & mean                   & median & mean  & median & mean  \\
                \midrule $0.1(1-t)^{2}$                                                                      & 78.62                                & 84.21                                  & 0.492                                & 0.474                                  & 68.49                  & 73.82  & 0.620 & 0.565  & 9.197 \\
                $0.2(1-t)^{2}$                                                                               & 77.24                                & 82.31                                  & 0.499                                & 0.479                                  & 69.97                  & 76.47  & 0.607 & 0.541  & 7.254 \\
                $0.3(1-t)^{2}$                                                                               & 76.36                                & 80.86                                  & 0.508                                & 0.485                                  & 71.15                  & 77.57  & 0.598 & 0.530  & 5.753 \\
                $0.4(1-t)^{2}$                                                                               & 75.74                                & 80.00                                  & 0.516                                & 0.496                                  & 72.23                  & 78.92  & 0.916 & 0.515  & 4.631 \\
                $0.5(1-t)^{2}$                                                                               & 74.47                                & 78.15                                  & 0.530                                & 0.502                                  & 72.51                  & 79.57  & 0.605 & 0.510  & 3.811 \\
                \bottomrule
            \end{tabular}
        \end{adjustbox}
    \end{subfigure}
    \begin{subfigure}
        {\textwidth}
        \centering
        \includegraphics[width=0.9\linewidth]{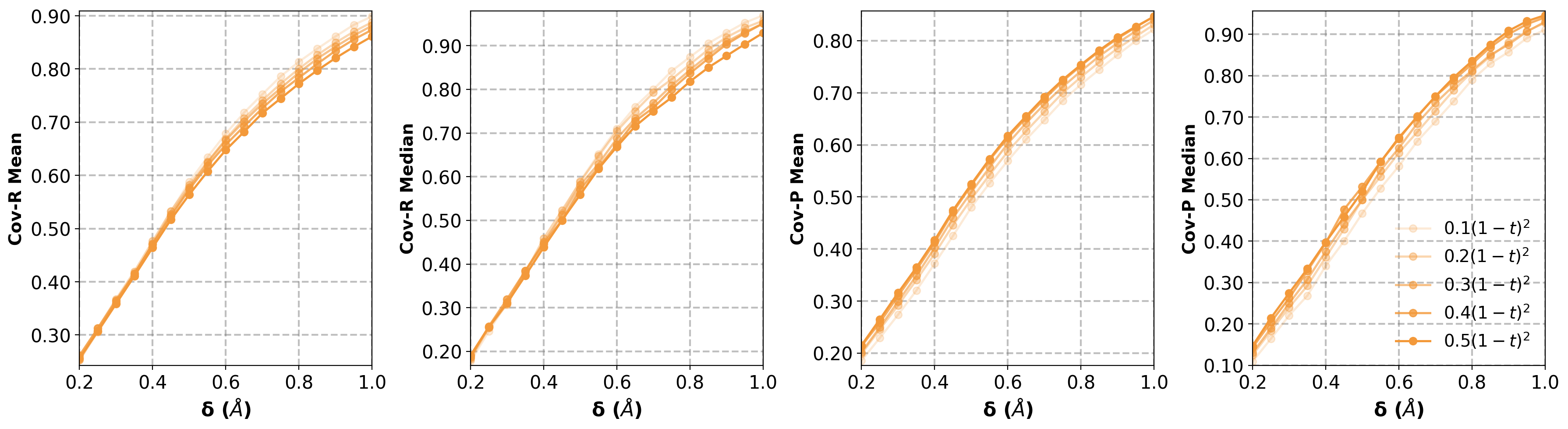}
    \end{subfigure}
    \caption{ Ablation study of guidance strengths $\lambda_{t}$ for 5-step (a) and
    50-step (b) ODE sampling on GEOM-QM9, and for 5-step (c) ODE sampling on
    GEOM-Drugs. The table reports Recall and Precision metrics at a fixed RMSD threshold
    of $\delta = 0.5$~\AA\ for GEOM-QM9 and $\delta = 0.75$~\AA\ for GEOM-Drugs,
    together with the mean predicted energy $J_{\phi}$. The plots depict how these
    metrics vary as a function of the RMSD threshold $\delta$. }
    \label{fig: ablation_guidance_schedule_qm9_so3}
\end{figure*}

In the ``Choice of the Guidance Schedule'' subsection, we outline the guidance schedule
employed in our experiments. A natural question arises as to why relatively
small guidance strengths are selected for the 5-step and 50-step ODE samplers. To
justify this design choice, we perform an ablation study over a range of
guidance strengths $\lambda_{t}$ under both settings. The quantitative results and
metric curves are summarized in \autoref{fig: ablation_guidance_schedule_qm9_so3}.

The ablation results reveal a trade-off governed by the guidance strength.
Stronger guidance generally reduces COV-R and increases COV-P, although individual
AMR statistics do not change monotonically. The coverage curves show that precision
gains are most pronounced at small RMSD thresholds $\delta$.
At the same time, the mean predicted energy $J_{\phi}$
decreases markedly, indicating that stronger guidance biases the sampling
process toward lower-energy regions of the conformational landscape. Consequently,
higher-energy conformations become underrepresented, while the generated structures
align more closely with low-energy reference conformations.

This behavior suggests that increasing the guidance strength makes the sampler more
conservative: it preferentially generates energetically favorable
structures—thereby enhancing Precision—but at the expense of structural
diversity, which leads to reduced Recall across RMSD thresholds. Because our
primary objective is to attain high-quality conformations with very few ODE
steps, we adopt moderate guidance strengths for the 5-step and 50-step settings.
This choice yields a more balanced trade-off between Recall and Precision, enabling
fair comparison with existing baselines while still leveraging the benefits of energy-guided
sampling.

\subsubsection{Why Energy-Matching Training Is Necessary for the Energy Model}

Since our primary objective is to employ the energy-guided vector field
$v^{\prime}_{t}(\mathcal{C}_{t}, t)$ to yield the guided probability path
$p^{\prime}_{t}(\mathcal{C}_{t})$, rather than relying on the original unguided vector
field $v_{\theta}(\mathcal{C}_{t}, t)$ that produces the unguided path $p_{t}(\mathcal{C}
_{t})$, it is essential that the learned energy model provides accurate and well-calibrated
gradients. Both vector fields transport the same source distribution (the
Harmonic prior $p_{0}(\mathcal{C}_{0})$) but toward different targets,
$p_{1}(\mathcal{C}_{1})\, e^{-J_{\phi}(\mathcal{C}_{1})}$ for the guided case
and $p_{1}(\mathcal{C}_{1})$ for the unguided case. While the energy regression loss
$\mathcal{L}_{\text{energy}}$ fits the energies of data conformations, it does not
directly constrain energy gradients along prior-to-data interpolation paths. In
contrast, the \emph{Energy Matching} loss $\mathcal{L}_{\mathrm{em}}$ aligns the
negative energy gradient with the transport direction from Harmonic-prior samples
to data conformations.

\begin{figure*}[t]
    \centering
    \makebox[\textwidth]{\footnotesize{(a) 2 ODE steps sampling on GEOM-QM9}}
    \begin{subfigure}
        {\textwidth}
        \centering
        \begin{adjustbox}
            {max width=0.85\textwidth}
            \begin{tabular}{lcccccccc}
                \toprule \multirow{2}{*}{methods}                                                                      & \multicolumn{2}{c}{COV-R $\uparrow$} & \multicolumn{2}{c}{AMR-R $\downarrow$} & \multicolumn{2}{c}{COV-P $\uparrow$} & \multicolumn{2}{c}{AMR-P $\downarrow$} \\
                \cmidrule(lr){2-3}\cmidrule(lr){4-5}\cmidrule(lr){6-7}\cmidrule(lr){8-9}                               & mean                                 & median                                 & mean                                 & median                                & mean  & median & mean  & median \\
                \midrule $v_{(}\mathcal{C}_{t}, t)$                                                                    & 97.18                                & 100.00                                 & 0.107                                & 0.062                                 & 93.54 & 100.00 & 0.151 & 0.107  \\
                $v^{\prime}_{t}(\mathcal{C}_{t}, t)\ \&\ \mathcal{L}_{\mathrm{energy}}$                                & 96.73                                & 100.00                                 & 0.112                                & 0.068                                 & 91.14 & 100.00 & 0.174 & 0.122  \\
                $v^{\prime}_{t}(\mathcal{C}_{t}, t)\ \&\ \mathcal{L}_{\mathrm{energy}}\ \&\ \mathcal{L}_{\mathrm{em}}$ & 95.14                                & 100.00                                 & 0.109                                & 0.060                                 & 95.18 & 100.00 & 0.118 & 0.067  \\
                \bottomrule
            \end{tabular}
        \end{adjustbox}
    \end{subfigure}
    \begin{subfigure}
        {\textwidth}
        \centering
        \includegraphics[width=0.9\linewidth]{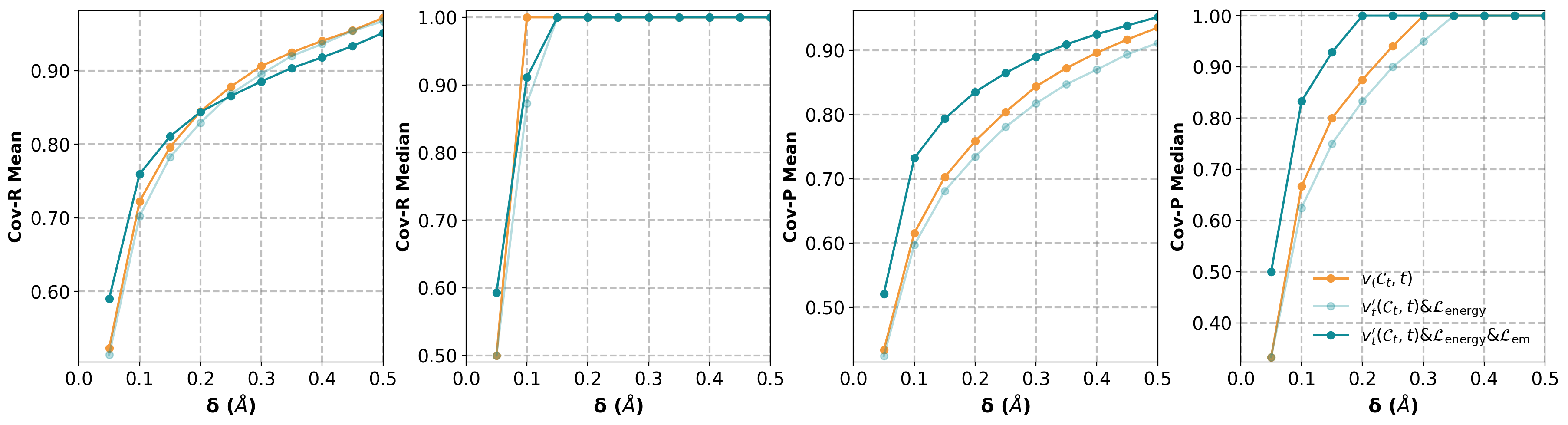}
    \end{subfigure}
    \makebox[\textwidth]{\footnotesize{(b) 5 ODE steps sampling on GEOM-QM9}}
    \begin{subfigure}
        {\textwidth}
        \centering
        \begin{adjustbox}
            {max width=0.85\textwidth}
            \begin{tabular}{lcccccccc}
                \toprule \multirow{2}{*}{methods}                                                                      & \multicolumn{2}{c}{COV-R $\uparrow$} & \multicolumn{2}{c}{AMR-R $\downarrow$} & \multicolumn{2}{c}{COV-P $\uparrow$} & \multicolumn{2}{c}{AMR-P $\downarrow$} \\
                \cmidrule(lr){2-3}\cmidrule(lr){4-5}\cmidrule(lr){6-7}\cmidrule(lr){8-9}                               & mean                                 & median                                 & mean                                 & median                                & mean  & median & mean  & median \\
                \midrule $v_{(}\mathcal{C}_{t}, t)$                                                                    & 96.02                                & 100.00                                 & 0.084                                & 0.038                                 & 94.48 & 100.00 & 0.103 & 0.053  \\
                $v^{\prime}_{t}(\mathcal{C}_{t}, t)\ \&\ \mathcal{L}_{\mathrm{energy}}$                                & 95.82                                & 100.00                                 & 0.084                                & 0.037                                 & 94.59 & 100.00 & 0.100 & 0.048  \\
                $v^{\prime}_{t}(\mathcal{C}_{t}, t)\ \&\ \mathcal{L}_{\mathrm{energy}}\ \&\ \mathcal{L}_{\mathrm{em}}$ & 96.28                                & 100.00                                 & 0.081                                & 0.037                                 & 95.39 & 100.00 & 0.088 & 0.038  \\
                \bottomrule
            \end{tabular}
        \end{adjustbox}
    \end{subfigure}
    \begin{subfigure}
        {\textwidth}
        \centering
        \includegraphics[width=0.9\linewidth]{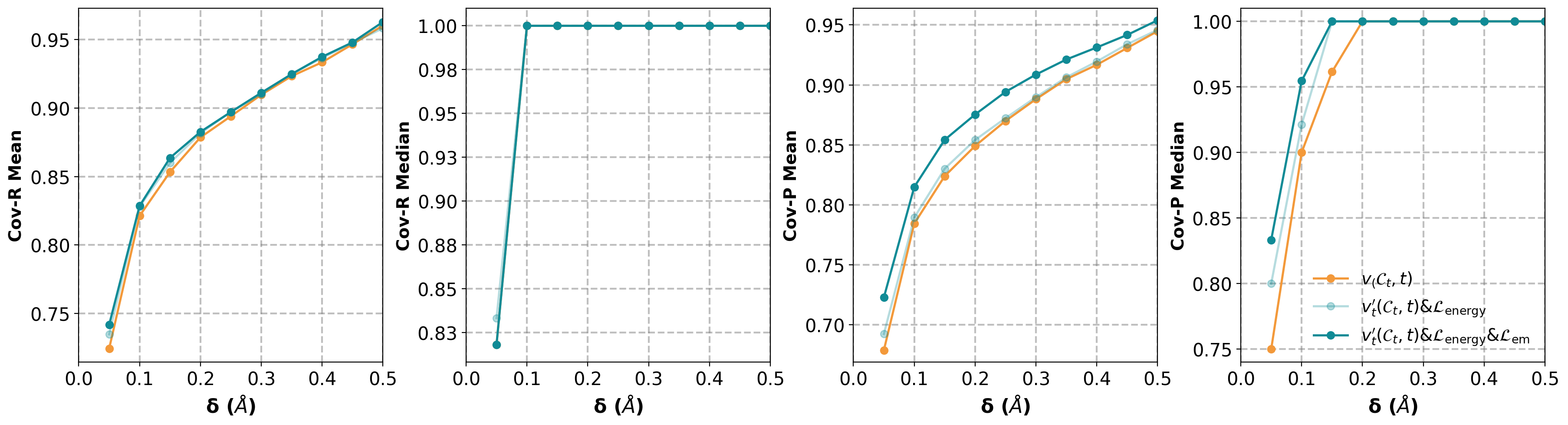}
    \end{subfigure}
    \caption{ Ablation study on the necessity of \emph{Energy Matching} training.
    For 2-step (a) and 5-step (b) ODE sampling on GEOM-QM9, the table reports
    Recall and Precision metrics at a fixed RMSD threshold $\delta = 0.5$~\AA, and
    the plots depict how these metrics vary as a function of the RMSD threshold $\delta$.
    }
    \label{fig:energy_matching_necessity}
    
\end{figure*}

The necessity of \emph{Energy Matching}~\citep{balcerak2025energy} is confirmed
empirically in ~\autoref{fig:energy_matching_necessity}. For both the
2-step and 5-step ODE samplers, using only the unguided vector field $v(\mathcal{C}
_{t}, t)$ leads to high Recall (COV-R and AMR-R) but noticeably weaker Precision
(COV-P and AMR-P), indicating poor energetic consistency. Adding guidance via $v^{\prime}
_{t}(\mathcal{C}_{t}, t)$ with only $\mathcal{L}_{\text{energy}}$ gives inconsistent
changes in performance. Training with both $\mathcal{L}_{\text{energy}}$ and
$\mathcal{L}_{\mathrm{em}}$ improves Precision-oriented metrics in both settings.
At 2 steps, these gains are accompanied by lower mean COV-R and higher mean AMR-R
than the unguided baseline; at 5 steps, both Recall- and Precision-oriented metrics
improve. These results support the role of \emph{Energy Matching} in learning
energy gradients for effective guided ODE sampling.

\subsubsection{Ablation Studies on the Choice of the Vector Field for Sampling}

\begin{figure*}[t]
    \centering
    \makebox[\textwidth]{\footnotesize{(a) 2 ODE steps sampling on GEOM-QM9}}
    \begin{subfigure}
        {\textwidth}
        \centering
        \begin{adjustbox}
            {max width=0.85\textwidth}
            \begin{tabular}{lccccccccc}
                \toprule \multirow{2}{*}{Vector field}                                                                          & \multicolumn{2}{c}{COV-R $\uparrow$} & \multicolumn{2}{c}{AMR-R $\downarrow$} & \multicolumn{2}{c}{COV-P $\uparrow$} & \multicolumn{2}{c}{AMR-P $\downarrow$} & $J_{\phi}$ $\downarrow$ \\
                \cmidrule(lr){2-3}\cmidrule(lr){4-5}\cmidrule(lr){6-7}\cmidrule(lr){8-9}\cmidrule(lr){10-10}                    & mean                                 & median                                 & mean                                 & median                                 & mean                   & median & mean  & median & mean   \\
                \midrule $v_{\theta}(\mathcal{C}_{t}, t)$                                                                       & 96.46                                & 100.00                                 & 0.110                                & 0.066                                  & 93.56                  & 100.00 & 0.150 & 0.110  & 10.182 \\
                $-\ \nabla_{\mathcal{C}_t}J_{\phi}(\mathcal{C}_{t})$                                                            & 94.82                                & 100.00                                 & 0.167                                & 0.130                                  & 89.67                  & 100.00 & 0.246 & 0.208  & 36.150 \\
                $v_{\theta}(\mathcal{C}_{t}, t) - \lambda_{t}\cdot \nabla_{\hat{\mathcal{C}}_1}J_{\phi}(\hat{\mathcal{C}}_{1})$ & 96.33                                & 100.00                                 & 0.103                                & 0.055                                  & 95.74                  & 100.00 & 0.116 & 0.066  & 5.579  \\
                \bottomrule
            \end{tabular}
        \end{adjustbox}
    \end{subfigure}
    \begin{subfigure}
        {\textwidth}
        \centering
        \includegraphics[width=0.9\linewidth]{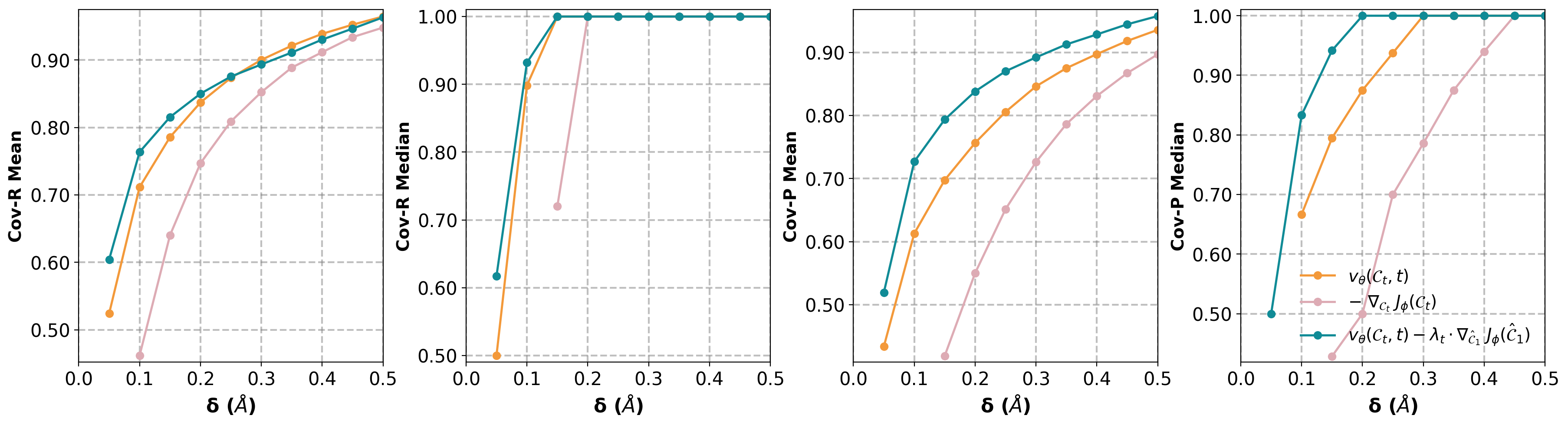}
    \end{subfigure}
    \makebox[\textwidth]{\footnotesize{(b) 5 ODE steps sampling on GEOM-QM9}}
    \begin{subfigure}
        {\textwidth}
        \centering
        \begin{adjustbox}
            {max width=0.85\textwidth}
            \begin{tabular}{lccccccccc}
                \toprule \multirow{2}{*}{Vector field}                                                                          & \multicolumn{2}{c}{COV-R $\uparrow$} & \multicolumn{2}{c}{AMR-R $\downarrow$} & \multicolumn{2}{c}{COV-P $\uparrow$} & \multicolumn{2}{c}{AMR-P $\downarrow$} & $J_{\phi}$ $\downarrow$ \\
                \cmidrule(lr){2-3}\cmidrule(lr){4-5}\cmidrule(lr){6-7}\cmidrule(lr){8-9}\cmidrule(lr){10-10}                    & mean                                 & median                                 & mean                                 & median                                 & mean                   & median & mean  & median & mean   \\
                \midrule $v_{\theta}(\mathcal{C}_{t}, t)$                                                                       & 96.58                                & 100.00                                 & 0.079                                & 0.037                                  & 95.00                  & 100.00 & 0.097 & 0.048  & 1.218  \\
                $-\ \nabla_{\mathcal{C}_t}J_{\phi}(\mathcal{C}_{t})$                                                            & 95.50                                & 100.00                                 & 0.148                                & 0.116                                  & 89.30                  & 100.00 & 0.246 & 0.204  & 16.338 \\
                $v_{\theta}(\mathcal{C}_{t}, t) - \lambda_{t}\cdot \nabla_{\hat{\mathcal{C}}_1}J_{\phi}(\hat{\mathcal{C}}_{1})$ & 95.92                                & 100.00                                 & 0.082                                & 0.034                                  & 95.32                  & 100.00 & 0.089 & 0.038  & 0.608  \\
                \bottomrule
            \end{tabular}
        \end{adjustbox}
    \end{subfigure}
    \begin{subfigure}
        {\textwidth}
        \centering
        \includegraphics[width=0.9\linewidth]{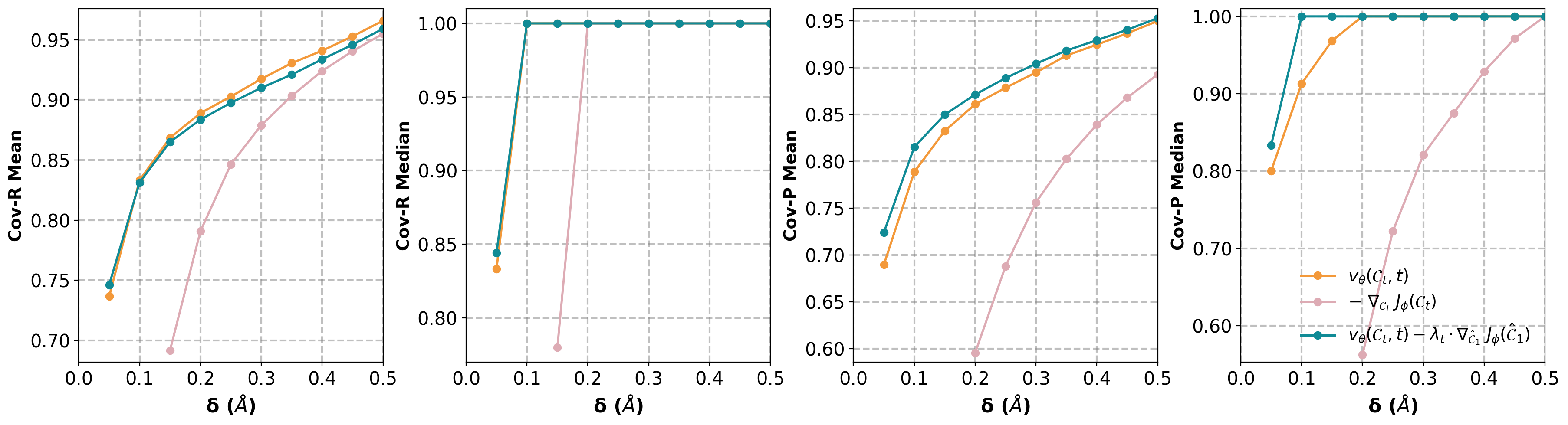}
    \end{subfigure}
    \makebox[\textwidth]{\footnotesize{(c) 5 ODE steps sampling on GEOM-Drugs}}
    \begin{subfigure}
        {\textwidth}
        \centering
        \begin{adjustbox}
            {max width=0.85\textwidth}
            \begin{tabular}{lccccccccc}
                \toprule \multirow{2}{*}{Vector field}                                                                          & \multicolumn{2}{c}{COV-R $\uparrow$} & \multicolumn{2}{c}{AMR-R $\downarrow$} & \multicolumn{2}{c}{COV-P $\uparrow$} & \multicolumn{2}{c}{AMR-P $\downarrow$} & $J_{\phi}$ $\downarrow$ \\
                \cmidrule(lr){2-3}\cmidrule(lr){4-5}\cmidrule(lr){6-7}\cmidrule(lr){8-9}\cmidrule(lr){10-10}                    & mean                                 & median                                 & mean                                 & median                                 & mean                   & median & mean  & median & mean   \\
                \midrule $v_{\theta}(\mathcal{C}_{t}, t)$                                                                       & 78.88                                & 85.71                                  & 0.488                                & 0.463                                  & 66.73                  & 71.78  & 0.643 & 0.580  & 12.174 \\
                $-\ \nabla_{\mathcal{C}_t}J_{\phi}(\mathcal{C}_{t})$                                                            & 73.62                                & 80.22                                  & 0.574                                & 0.554                                  & 53.68                  & 51.68  & 3.451 & 0.760  & 81.589 \\
                $v_{\theta}(\mathcal{C}_{t}, t) - \lambda_{t}\cdot \nabla_{\hat{\mathcal{C}}_1}J_{\phi}(\hat{\mathcal{C}}_{1})$ & 77.69                                & 83.33                                  & 0.497                                & 0.469                                  & 70.41                  & 76.01  & 0.599 & 0.531  & 7.244  \\
                \bottomrule
            \end{tabular}
        \end{adjustbox}
    \end{subfigure}
    \begin{subfigure}
        {\textwidth}
        \centering
        \includegraphics[width=0.9\linewidth]{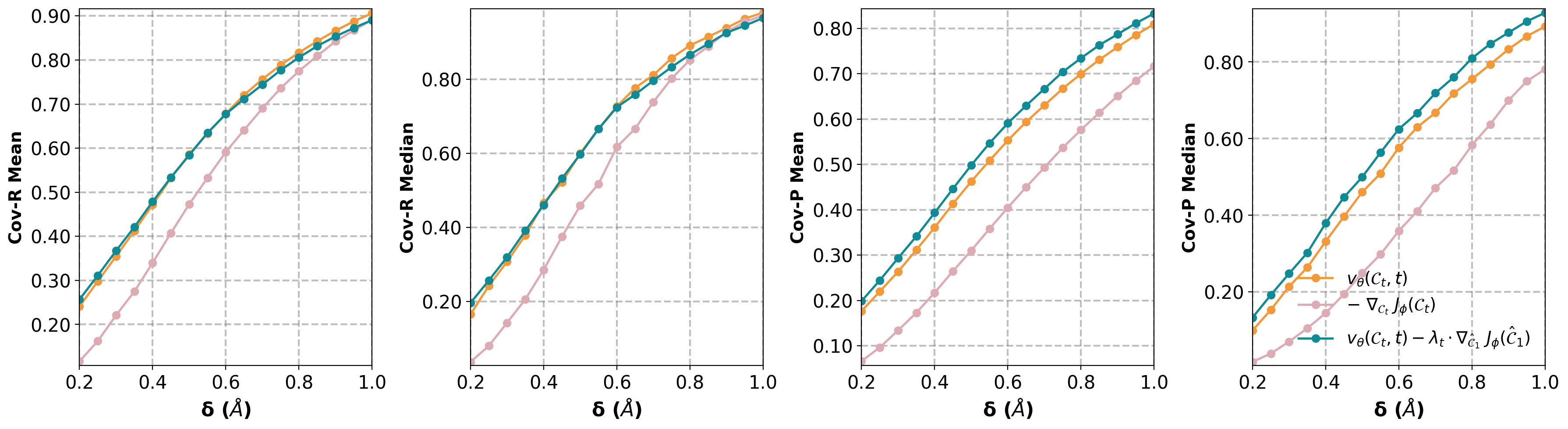}
    \end{subfigure}
    \caption{ Ablation study of different types of vector fields for 2-step (a) and
    5-step (b) ODE sampling on GEOM-QM9, and for 5-step (c) ODE sampling on GEOM-Drugs.
    The table reports Recall and Precision metrics at a fixed RMSD threshold of $\delta
    = 0.5$~\AA\ for GEOM-QM9 and $\delta = 0.75$~\AA\ for GEOM-Drugs, together
    with the mean predicted energy $J_{\phi}$. The plots depict how these metrics
    vary as a function of the RMSD threshold $\delta$. }
    \label{fig: ablation_vector_filed_types}
\end{figure*}

To further validate the robustness of our proposed guided vector field $v^{\prime}$,
we conduct ablation studies comparing three vector fields for ODE sampling: (1) the
unguided vector field $v_{\theta}(\mathcal{C}_{t}, t)$, corresponding to the
pure flow-matching framework in ET-Flow~\cite{hassan2024flow}; (2) the pure energy-gradient
field $-\nabla_{\mathcal{C}_{t}}J_{\phi}(\mathcal{C}_{t})$, representing a
straight-path flow-matching formulation as in Ref.~\cite{liu2022flow} with a \emph{time-independent}
vector field, or equivalently, deterministic Langevin dynamics~\cite{welling2011bayesian}
without noise (i.e., Gradient Descent); and (3) our proposed energy-guided vector
field $v_{\theta}(\mathcal{C}_{t}, t) - \lambda_{t}\, \nabla_{\hat{\mathcal{C}}_{1}}
J_{\phi}(\hat{\mathcal{C}}_{1})$. The quantitative results and metric
trajectories are presented in~\autoref{fig: ablation_vector_filed_types}.

The results in~\autoref{fig: ablation_vector_filed_types} show that the guided field
improves COV-P and AMR-P relative to the unguided field $v_{\theta}(\mathcal{C}_{t},t)$
in all three tabulated settings. Recall-oriented results show a trade-off: mean
AMR-R improves for 2-step GEOM-QM9 sampling but increases slightly for 5-step
sampling on GEOM-QM9 and GEOM-Drugs, while mean COV-R decreases in all three
settings. The pure energy-gradient field of the energy-based model (EBM),
$-\nabla_{\mathcal{C}_{t}}J_{\phi}(\mathcal{C}_{t})$, produces lower coverage and
larger AMR errors than the guided field. The guided field also yields lower mean
predicted energies $J_{\phi}$ than both alternatives in each setting. Together,
these results show improved precision and a shift toward lower predicted energies,
with trade-offs in recall.

\subsubsection{Energy-guided sampling improves ensemble-level property
consistency}

RMSD-based metrics assess geometric agreement between generated and reference conformations,
but they do not fully capture whether a generated ensemble preserves chemically
relevant properties. We therefore further evaluate ensemble-level consistency using
Boltzmann-weighted molecular properties.

Following established conformer-generation protocols~\citep{jing2022torsional,wang2023swallowing,hassan2024flow},
we evaluate a random subset of 100 molecules from the GEOM-Drugs test set. For
each molecule with $K$ reference conformers, we generate $\min(2K,32)$ conformers
and compare ensemble properties computed from the generated conformations with those
of the reference ensemble. Specifically, we report median absolute errors for the
ensemble energy ($E$), dipole moment ($\mu$), highest occupied molecular
orbital--lowest unoccupied molecular orbital (HOMO--LUMO) gap ($\Delta \epsilon$),
and minimum energy ($E_{\min}$). Properties are evaluated both before and after
GFN2-xTB~\citep{bannwarth2019gfn2} geometry relaxation, which separates the
quality of the raw generated conformations from the quality obtained after local
structural refinement.

As shown in~\autoref{tab:ensemble_properties_relaxation}, pre-relaxation errors are
large for methods with available results, consistent with the sensitivity of
quantum-chemical properties to local geometric distortions. Because pre-relaxation
results are unavailable for several baselines, we focus this comparison on the
reproduced ET-Flow baseline. Before relaxation, \ours reduces errors on the
energy-related quantities $E$, $\Delta\epsilon$, and $E_{\min}$ relative to
reproduced ET-Flow, suggesting that energy-guided sampling already shifts generated
conformations toward more physically plausible regions before post-processing.

After GFN2-xTB relaxation, all methods show substantially reduced property
errors. In this setting, \ours achieves the lowest errors on $E$, $\mu$, and
$E_{\min}$, while remaining competitive on $\Delta\epsilon$. These results
indicate that the generated ensembles preserve not only geometric structure but
also energetic and electronic characteristics of the reference ensembles. The pre-relaxation
improvements further support the role of the learned energy model within the generative
dynamics, rather than merely as a post-generation correction.

\begin{table*}
    [t]
    \caption{Median ensemble-property errors between generated and reference
    conformers on a 100-molecule subset of GEOM-Drugs, evaluated before and after
    GFN2-xTB geometry relaxation. $E$, $\Delta \epsilon$, and $E_{\min}$ are
    reported in kcal/mol, and $\mu$ is reported in debye. Lower values indicate
    better agreement with the reference ensemble. Best and second-best results are
    highlighted in bold and underlined within each evaluation setting.}
    \label{tab:ensemble_properties_relaxation}
    \centering
    \begin{adjustbox}
        {max width=\textwidth}
        \begin{tabular}{lcccccccc}
            \toprule \multirow{2}{*}{Method}     & \multicolumn{4}{c}{Without relaxation} & \multicolumn{4}{c}{With relaxation} \\
            \cmidrule(lr){2-5}\cmidrule(lr){6-9} & $E$                                    & $\mu$                              & $\Delta \epsilon$          & $E_{\min}$                 & $E$                       & $\mu$                       & $\Delta \epsilon$           & $E_{\min}$                  \\
            \midrule RDKit                       & 39.08                                  & 1.40                               & 5.04                       & 39.14                      & 0.81                      & 0.52                        & 0.75                        & 1.16                        \\
            OMEGA                                & 16.47                                  & 0.78                               & 3.25                       & 16.45                      & 0.68                      & 0.66                        & 0.68                        & 0.69                        \\
            GeoMol                               & 43.27                                  & 1.22                               & 7.36                       & 43.68                      & 0.42                      & 0.34                        & 0.59                        & 0.40                        \\
            GeoDiff                              & 18.82                                  & 1.34                               & 4.96                       & 19.43                      & 0.31                      & 0.35                        & 0.89                        & 0.39                        \\
            Torsional Diff.                      & 36.91                                  & 0.92                               & 4.93                       & 36.94                      & 0.22                      & 0.35                        & 0.54                        & 0.13                        \\
            MCF                                  & --                                     & --                                 & --                         & --                         & 0.68$\pm$0.06             & 0.28$\pm$0.05               & 0.63$\pm$0.05               & 0.04$\pm$0.00               \\
            ET-Flow                              & --                                     & --                                 & --                         & --                         & \underline{0.18$\pm$0.01} & 0.18$\pm$0.01               & 0.35$\pm$0.06               & 0.02$\pm$0.00               \\
            $\text{ET-Flow}_{\text{reproduced}}$ & \underline{2.36$\pm$0.132}             & \textbf{0.270$\pm$0.025}           & \underline{1.00$\pm$0.017} & \underline{2.36$\pm$0.198} & 0.135$\pm$0.036           & \underline{0.166$\pm$0.015} & \textbf{0.263$\pm$0.030}    & \underline{0.018$\pm$0.006} \\
            \ours                                & \textbf{2.21$\pm$0.085}                & \underline{0.291$\pm$0.050}        & \textbf{0.952$\pm$0.081}   & \textbf{2.22$\pm$0.141}    & \textbf{0.126$\pm$0.044}  & \textbf{0.164$\pm$0.023}    & \underline{0.276$\pm$0.055} & \textbf{0.016$\pm$0.003}    \\
            \bottomrule
        \end{tabular}
    \end{adjustbox}
\end{table*}

\subsubsection{Energy Landscape Analysis and Guided Conformational Sampling}

We further analyze whether the learned energy model captures meaningful features
of the molecular energy landscape and whether this landscape contributes to
guided conformational sampling. This analysis complements the main results by providing
mechanistic evidence that the learned energy function is not only useful for ranking
generated conformations, but also shapes the sampling process toward lower-energy
regions.

We first compare three variants of \oursregular on GEOM-Drugs with 5 ODE
sampling steps: (1) an unguided baseline (ET-Flow; \textbf{w/o guidance}); (2) a
guided model in which the energy model $J_{\phi}$ is trained only with the \emph{Energy
Matching} loss $\mathcal{L}_{\mathrm{em}}$ (\ours: \textbf{guidance \&
$\mathcal{L}_{\mathrm{em}}$}); and (3) the full guided model trained with both
$\mathcal{L}_{\mathrm{em}}$ and the energy regression loss
$\mathcal{L}_{\mathrm{energy}}$ (\ours: \textbf{guidance \&
$\mathcal{L}_{\mathrm{em}}$ \& $\mathcal{L}_{\mathrm{energy}}$}). For each model,
we generate one conformation per molecule and evaluate it against the ground-state
structure. Each experiment is repeated 10 times to obtain stable statistics.

As shown in~\autoref{fig: ground_state_guidance_ablation_drugs_o3}, the full
model achieves the lowest median error and the narrowest error distribution on
the main $\mathcal{C}\text{-RMSD}$ metric, compared with both the unguided baseline
and the Energy-Matching-only variant. Similar trends are observed for
$\mathbf{D}\text{-MAE}$ and $\mathbf{D}\text{-RMSE}$, where the full model shifts
the error distributions toward smaller values. These results suggest that the
combined training objective improves the ability of $J_{\phi}$ to guide sampling
toward conformations closer to the ground-state structure.

We next visualize the learned energy landscape for six randomly selected
molecules in~\autoref{fig: landscape_show}. As shown in \autoref{fig: landscape_show}(b),
after normalization to $[0,1]$, the predicted energy values follow the true
energy profiles of the ground-truth conformations, indicating that the learned energy
model captures meaningful relative variations within each molecular conformational
ensemble.

Finally, \autoref{fig: landscape_show}(c) and~\autoref{fig: landscape_show}(d)
compare the predicted energy distributions of conformations generated with 2 and
5 ODE sampling steps. In both settings, energy guidance shifts the generated
ensembles toward lower predicted energies and reduces the spread of the energy
distribution. This trend becomes more pronounced with additional sampling steps,
suggesting that the learned energy landscape progressively steers the sampling
process toward lower-energy regions of conformational space.

\clearpage

\begingroup
\setlength{\parskip}{3.5pt}
\setlength{\textfloatsep}{7pt plus 2pt minus 2pt}
\setlength{\floatsep}{7pt plus 2pt minus 2pt}
\setlength{\intextsep}{7pt plus 2pt minus 2pt}
\setlength{\abovecaptionskip}{4pt}
\setlength{\belowcaptionskip}{0pt}

\subsubsection{\texorpdfstring{Molecular Factors Affecting Learned-Energy/xTB Rank Correlation}{Molecular Factors Affecting Learned-Energy/xTB Rank Correlation}}

To characterize the observed molecule-to-molecule variation in
learned-energy/xTB agreement, we performed an exploratory analysis on the 100
GEOM-Drugs molecules used in the energy-correlation experiment, each represented
by 50 reference conformers. For every molecular graph, we computed RDKit
descriptors covering size, flexibility, ring structure, aromaticity, and fraction
Csp$^{3}$. Ranking fidelity was measured by the molecule-wise Spearman
correlation between learned energy $J_{\phi}$ and GFN2-xTB single-point energy,
and descriptor--fidelity associations were assessed by across-molecule Spearman
correlations with permutation tests and false-discovery-rate correction.

\begin{figure*}[!htbp]
    \centering
    \includegraphics[width=0.80\textwidth]{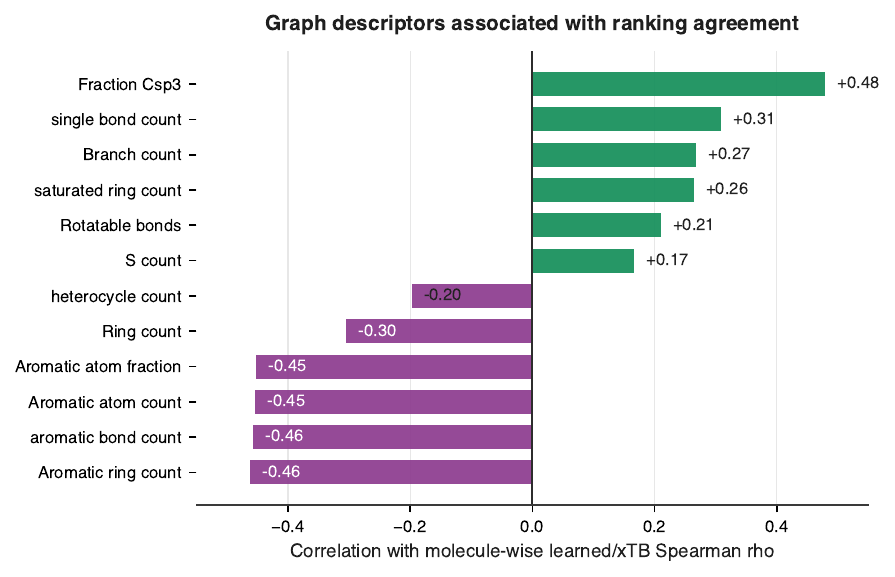}
    \caption{Univariate associations between molecular graph descriptors
    and molecule-wise learned-energy/xTB ranking fidelity ($n=100$ molecules; 50
    conformers per molecule). Bars report across-molecule Spearman correlations;
    fraction Csp$^{3}$ is most positively associated with fidelity, whereas
    aromaticity descriptors are most negatively associated.}
    \label{fig:molecular_factor_descriptor_bar}
\end{figure*}

Across the 100 molecules, learned and xTB energies are positively rank
correlated in nearly all cases: the median molecule-wise Spearman coefficient is
$0.823$, with 96 molecules above $0.5$ and 57 at or above $0.8$. The strength of
this agreement nevertheless varies systematically with molecular topology
(\autoref{fig:molecular_factor_descriptor_bar}). Fraction Csp$^{3}$ has the
largest positive association ($r_s=0.480$), whereas aromatic-ring count and
aromatic-atom fraction are negatively associated ($r_s=-0.462$ and $-0.452$);
all three remain significant after correction ($q=0.002$). Heavy-atom count is
essentially unassociated with ranking fidelity ($r_s=-0.059$), and the weak
rotatable-bond trend ($r_s=0.210$) does not remain significant ($q=0.087$),
indicating that size and nominal flexibility alone do not explain the
between-molecule variation.

\begin{figure*}[!htbp]
    \centering
    \includegraphics[width=0.91\textwidth]{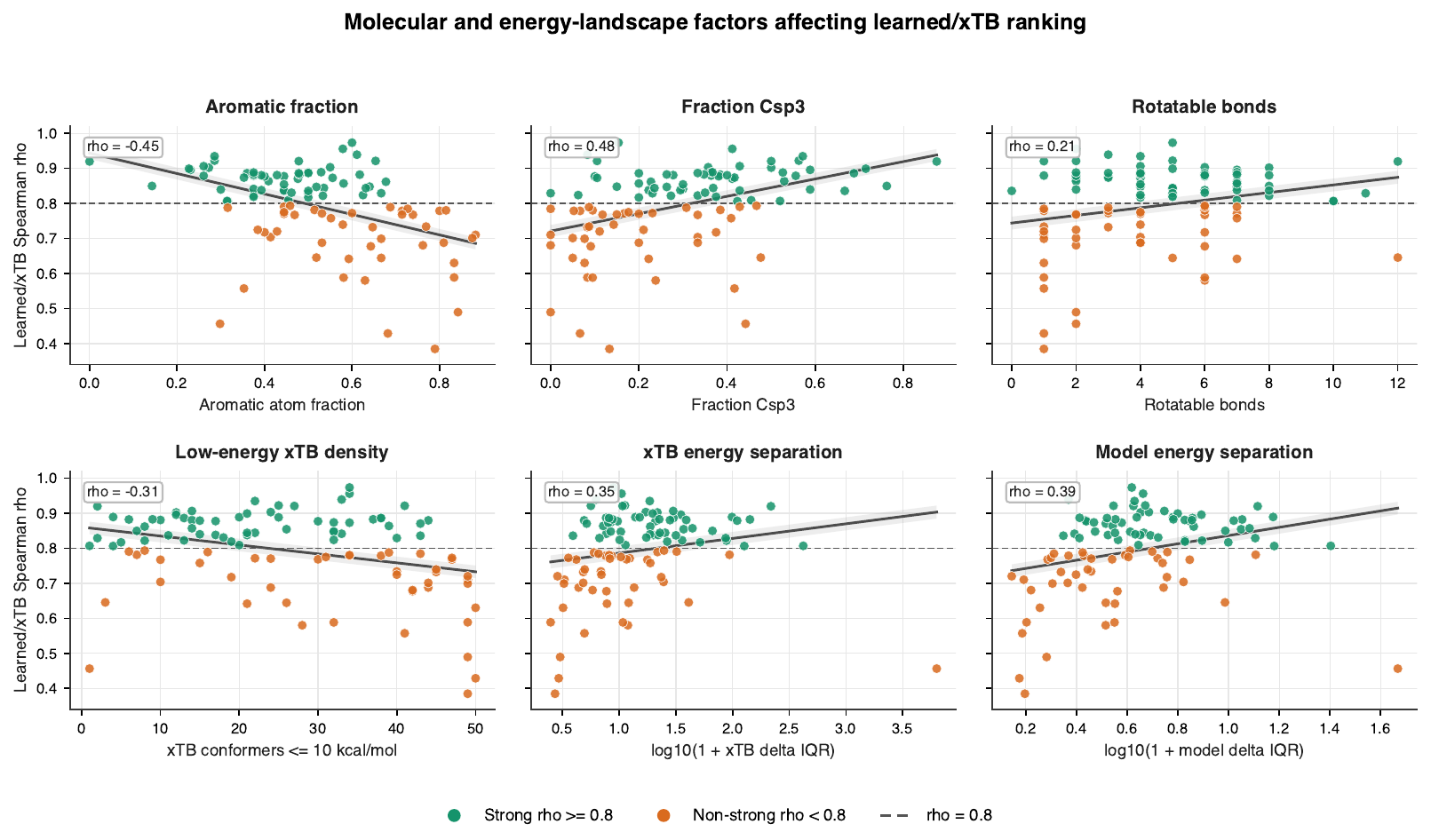}
    \caption{Molecule-level relationships between representative graph or
    energy-landscape descriptors and learned-energy/xTB ranking fidelity. Points
    denote molecules and fitted lines summarize univariate trends: fraction
    Csp$^{3}$ and energy-spread IQRs are positively associated with fidelity,
    whereas aromaticity and dense low-energy conformer populations are
    negatively associated.}
    \label{fig:molecular_factor_scatter_grid}
\end{figure*}

A complementary source of variation is the intrinsic separability of the
conformational energy landscape (\autoref{fig:molecular_factor_scatter_grid}).
The IQRs of the learned and xTB relative-energy distributions are positively
associated with ranking fidelity ($r_s=0.389$ and $0.346$, respectively), whereas
the number of conformers within 10~kcal/mol of the xTB minimum is negatively
associated with it ($r_s=-0.310$). This pattern is consistent with a resolution
effect: narrow energy windows make pairwise orderings sensitive to small
prediction errors, whereas broader energetic separation provides a better
resolved ranking signal. This interpretation concerns relative ordering and does
not imply that narrow landscapes necessarily have large absolute energy errors.

\begin{figure*}[!htbp]
    \centering
    \includegraphics[width=0.90\textwidth]{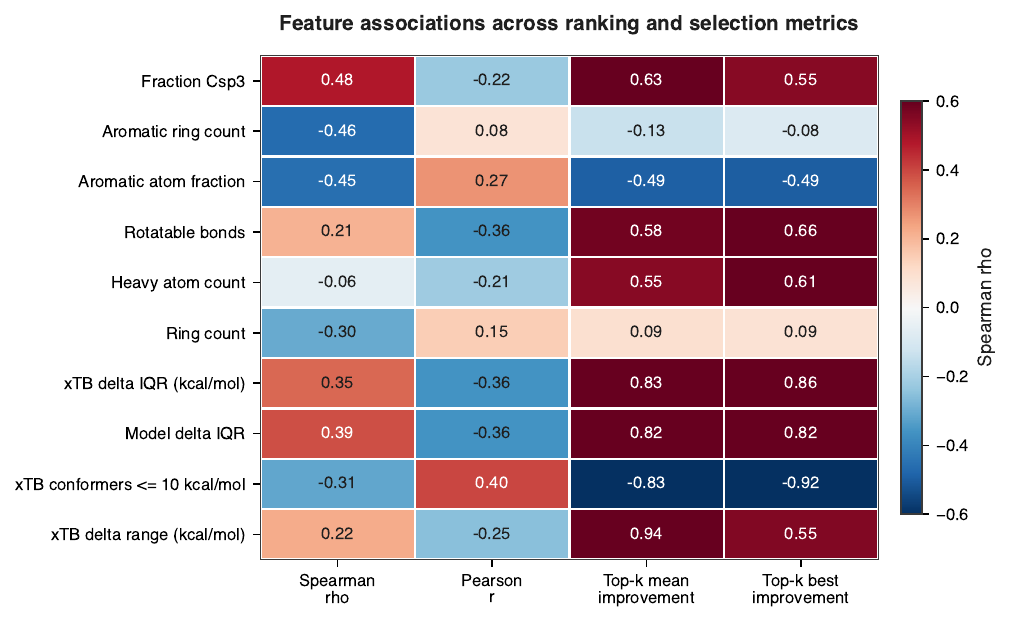}
    \caption{Across-molecule Spearman associations between selected graph
    and energy-landscape descriptors and four complementary evaluation metrics.
    The first two columns quantify full-ensemble rank and linear agreement; the
    final two quantify the improvement obtained by selecting conformers with the
    learned energy rather than at random. The distinct association patterns show
    that global energy agreement and practical top-$k$ selection benefit are
    related but non-equivalent objectives.}
    \label{fig:molecular_factor_correlation_heatmap}
\end{figure*}

The broader comparison in
\autoref{fig:molecular_factor_correlation_heatmap} also distinguishes global
agreement from downstream selection utility. In particular, energy-spread
descriptors are strongly associated with the magnitude of top-$k$ improvement,
whereas their associations with Spearman and Pearson agreement are more moderate.
Thus, learned-energy selection can remain useful even when a single global
correlation coefficient is imperfect. Overall, ranking fidelity is associated
with both molecular topology and energy-landscape separability: aromatic and
ring-rich graphs tend to show lower rank agreement, sp$^{3}$-rich graphs tend to
show higher agreement, and densely clustered low-energy conformers form a less
well-resolved ranking problem than energetically separated conformers. These
exploratory, univariate associations from 100 molecules should not be interpreted
as causal effects; rather, they indicate that residual learned-energy/xTB
heterogeneity reflects both molecular structure and landscape resolution.

\raggedbottom
\subsubsection{\texorpdfstring{Zero-Shot Out-of-Distribution Evaluation}{Zero-Shot Out-of-Distribution Evaluation}}

We assess zero-shot generalization by applying a checkpoint trained
exclusively on GEOM-Drugs to GEOM-QM9 and GEOM-XL, without fine-tuning or other
parameter updates. \textbf{GEOM-Drugs $\rightarrow$ GEOM-QM9} tests transfer to
substantially smaller molecules than those in the training distribution.

\begin{table}[H]
    \centering
    \caption{Zero-shot conformer generation from GEOM-Drugs to GEOM-QM9.
    Models are trained on GEOM-Drugs and evaluated directly on GEOM-QM9 without
    fine-tuning. \textbf{FM} denotes unguided flow matching
    (ET-Flow~\citep{hassan2024flow}); \ours denotes energy-guided sampling. The
    RMSD threshold is $\delta=0.5$~\AA.}
    \label{tab:ood_drugs_to_qm9}
    \begin{adjustbox}{max width=\textwidth}
        \renewcommand{\arraystretch}{1.1}
        \begin{tabular}{lccccccccc}
            \toprule
            \multirow{2}{*}{Method} & \multirow{2}{*}{ODE steps} &
            \multicolumn{2}{c}{COV-R $\uparrow$} &
            \multicolumn{2}{c}{AMR-R $\downarrow$} &
            \multicolumn{2}{c}{COV-P $\uparrow$} &
            \multicolumn{2}{c}{AMR-P $\downarrow$} \\
            \cmidrule(lr){3-4}\cmidrule(lr){5-6}\cmidrule(lr){7-8}\cmidrule(lr){9-10}
            & & mean & median & mean & median & mean & median & mean & median \\
            \midrule
            FM & 1 & 39.46 & 11.06 & 0.535 & 0.557 & 28.22 & 3.57 & 0.621 & 0.643 \\
            \textbf{\oursregular (Ours)} & 1 & 72.83 & 100.00 & 0.375 & 0.369 & 59.85 & 64.15 & 0.472 & 0.458 \\
            FM & 2 & 85.80 & 100.00 & 0.278 & 0.248 & 65.54 & 72.48 & 0.425 & 0.396 \\
            \textbf{\oursregular (Ours)} & 2 & 84.13 & 100.00 & 0.261 & 0.223 & 67.77 & 75.00 & 0.394 & 0.352 \\
            FM & 5 & 86.66 & 100.00 & 0.228 & 0.177 & 68.32 & 75.00 & 0.375 & 0.334 \\
            \textbf{\oursregular (Ours)} & 5 & 85.52 & 100.00 & 0.229 & 0.174 & 68.31 & 76.62 & 0.368 & 0.321 \\
            \bottomrule
        \end{tabular}
    \end{adjustbox}
\end{table}

As shown in~\autoref{tab:ood_drugs_to_qm9}, the GEOM-Drugs-trained model
retains substantial zero-shot capability on the smaller-molecule distribution,
especially under the tightest integration budget. At one ODE step, \oursregular
raises COV-R from 39.46\% to 72.83\% and COV-P from 28.22\% to 59.85\%, while
reducing AMR-R from 0.535 to 0.375 and AMR-P from 0.621 to 0.472. At two steps,
guidance further lowers both AMR metrics and improves COV-P; at five steps, the
samplers are closely matched in coverage and AMR-R while \oursregular retains
lower AMR-P.

\textbf{GEOM-Drugs $\rightarrow$ GEOM-XL:} GEOM-XL comprises 102
exceptionally large molecules that lie outside the GEOM-Drugs training
distribution and therefore probes extrapolation to a molecular-size regime not
observed during training. Following prior GEOM-XL evaluations of Torsional
Diffusion~\citep{jing2022torsional}, MCF~\citep{wang2023swallowing}, and
ET-Flow~\citep{hassan2024flow}, \autoref{tab:ood_geom_xl} reports their
published baselines alongside our controlled zero-shot results.

\begin{table}[!tbp]
    \centering
    \caption{Zero-shot conformer generation from GEOM-Drugs to GEOM-XL.
    Published baseline results are taken from Torsional
    Diffusion~\citep{jing2022torsional}, MCF~\citep{wang2023swallowing}, and
    ET-Flow~\citep{hassan2024flow}.}
    \label{tab:ood_geom_xl}
    \begin{adjustbox}{max width=0.95\textwidth}
        \renewcommand{\arraystretch}{1.1}
        \begin{tabular}{lcccccc}
            \toprule
            \multirow{2}{*}{Method} & \multirow{2}{*}{ODE steps} &
            \multicolumn{2}{c}{AMR-R $\downarrow$} &
            \multicolumn{2}{c}{AMR-P $\downarrow$} &
            \multirow{2}{*}{\# mols} \\
            \cmidrule(lr){3-4}\cmidrule(lr){5-6}
            & & mean & median & mean & median & \\
            \midrule
            \multicolumn{7}{l}{\emph{Published GEOM-XL results}} \\
            GeoDiff & -- & 2.92 & 2.62 & 3.35 & 3.15 & -- \\
            GeoMol & -- & 2.47 & 2.39 & 3.30 & 3.14 & -- \\
            Torsional Diff. & -- & 2.05 & 1.86 & 2.94 & 2.78 & -- \\
            MCF-S (13M) & -- & 2.22 & 1.97 & 3.17 & 2.81 & 102 \\
            MCF-B (64M) & -- & 2.01 & 1.70 & 3.03 & 2.64 & 102 \\
            MCF-L (242M) & -- & 1.97 & 1.60 & 2.94 & 2.43 & 102 \\
            ET-Flow (8.3M) & -- & 2.31 & 1.93 & 3.31 & 2.84 & 102 \\
            Torsional Diff. & -- & 1.93 & 1.86 & 2.84 & 2.71 & 77 \\
            MCF-S (13M) & -- & 2.02 & 1.87 & 2.90 & 2.69 & 77 \\
            MCF-B (64M) & -- & 1.71 & 1.61 & 2.69 & 2.44 & 77 \\
            MCF-L (242M) & -- & 1.64 & 1.51 & 2.57 & 2.26 & 77 \\
            ET-Flow (8.3M) & -- & 2.00 & 1.80 & 2.96 & 2.63 & 75 \\
            \midrule
            \multicolumn{7}{l}{\emph{Our controlled zero-shot evaluation}} \\
            $\text{ET-Flow}_{\text{reproduced}}$ (8.3M) & 1 & 2.212 & 2.133 & 3.071 & 2.996 & 71 \\
            \textbf{\oursregular (Ours, 16.6M)} & 1 & \textbf{2.100} & \textbf{2.024} & \textbf{3.053} & \textbf{2.918} & 71 \\
            $\text{ET-Flow}_{\text{reproduced}}$ (8.3M) & 2 & 1.991 & 1.967 & \textbf{2.989} & \textbf{2.843} & 59 \\
            \textbf{\oursregular (Ours, 16.6M)} & 2 & \textbf{1.983} & \textbf{1.939} & 3.068 & 2.996 & \textbf{63} \\
            $\text{ET-Flow}_{\text{reproduced}}$ (8.3M) & 5 & 2.003 & 1.942 & 3.108 & 2.986 & 64 \\
            \textbf{\oursregular (Ours, 16.6M)} & 5 & \textbf{1.956} & \textbf{1.933} & \textbf{3.054} & \textbf{2.983} & \textbf{67} \\
            $\text{ET-Flow}_{\text{reproduced}}$ (8.3M) & 50 & 2.022 & 1.993 & 3.169 & \textbf{2.999} & \textbf{73} \\
            \textbf{\oursregular (Ours, 16.6M)} & 50 & \textbf{1.977} & \textbf{1.949} & \textbf{3.144} & 3.017 & 69 \\
            \bottomrule
        \end{tabular}
    \end{adjustbox}
    \vspace{0.25em}
    \begin{minipage}{0.95\textwidth}
        \scriptsize \textbf{Table notes.} Bold values indicate the better result
        within each matched ODE-step pair in our controlled zero-shot evaluation.
        \# mols denotes the number of successfully evaluated molecules. Counts
        below 102 indicate generation or evaluation failures for some large
        molecules, as also reported by MCF~\citep{wang2023swallowing} and
        ET-Flow~\citep{hassan2024flow}.
    \end{minipage}
\end{table}

The GEOM-XL transfer is substantially more challenging, as reflected by
the smaller number of molecules that can be successfully generated and
evaluated. Nevertheless, relative to
$\text{ET-Flow}_{\text{reproduced}}$, \oursregular improves all four AMR
statistics at one step while retaining the same 71 valid molecules, with mean
AMR-R and AMR-P decreasing by 5.1\% and 0.6\%. At five steps, \oursregular
reduces mean AMR-R and AMR-P by 2.3\% and 1.7\% and evaluates three additional
molecules. The two- and 50-step settings show mixed trade-offs, so the controlled
comparison indicates the most consistent benefit from energy guidance in the
low-step regime, with more metric-dependent gains as the integration budget
increases.

\FloatBarrier
\subsubsection{\texorpdfstring{Sampling Efficiency and Energy-Guidance Overhead}{Sampling Efficiency and Energy-Guidance Overhead}}

We distinguish conformer quality under a fixed ODE budget, the cost of
one guided step, and the sampling-stage latency needed to reach a target quality.
Our few-step claim concerns quality under a restricted integration budget and
does not imply that a guided step is cheaper: each guided step evaluates both the
learned energy and its coordinate gradient. We therefore conducted a controlled
wall-clock study to test whether fewer required steps can offset this overhead.

For the matched-step comparison, FM was obtained by disabling energy
guidance in the same trained model while holding molecular inputs, test indices,
generated conformer counts, ODE solver, and integration-step budget fixed. After
warm-up, we timed only the CUDA-synchronized sampling call, excluding data
loading, batch construction, host-to-device transfer, file I/O, and metric
evaluation. For each molecule with $K$ references, both samplers generated $2K$
conformers. GEOM-QM9 covers all 998 valid test molecules over three repeats;
GEOM-Drugs uses one complete pass over a reproducible 100-molecule subset.

\begin{table}[H]
    \centering
    \caption{Sampling-only wall-clock latency and energy-guidance
    overhead. Panel (a) isolates matched-step guidance cost; panel (b) compares
    quality-comparable operating points.}
    \label{tab:sampling_efficiency_guidance_overhead}
    \renewcommand{\arraystretch}{1.04}
    \textbf{(a) Fixed-step guidance overhead}\\[0.2em]
    \begin{adjustbox}{max width=\textwidth}
        \begin{tabular}{lcccc}
            \toprule
            Dataset & ODE steps & FM time (s/mol.) & \oursregular time (s/mol.) &
            Overhead (\oursregular/FM) \\
            \midrule
            \multirow{4}{*}{GEOM-QM9}
            & 1  & $0.0468 \pm 0.0007$ & $0.1309 \pm 0.0014$ & $2.80\times$ \\
            & 2  & $0.0823 \pm 0.0009$ & $0.2490 \pm 0.0027$ & $3.03\times$ \\
            & 5  & $0.1837 \pm 0.0004$ & $0.6063 \pm 0.0026$ & $3.30\times$ \\
            & 50 & $1.7498 \pm 0.0120$ & $6.0640 \pm 0.0144$ & $3.47\times$ \\
            \midrule
            \multirow{4}{*}{GEOM-Drugs}
            & 1  & $0.5239$  & $1.2245$  & $2.34\times$ \\
            & 2  & $0.8363$  & $2.1691$  & $2.59\times$ \\
            & 5  & $1.7544$  & $5.0232$  & $2.86\times$ \\
            & 50 & $15.4513$ & $58.1521$ & $3.76\times$ \\
            \bottomrule
        \end{tabular}
    \end{adjustbox}

    \vspace{0.45em}
    \textbf{(b) Quality-comparable few-step speedup}\\[0.2em]
    \begin{adjustbox}{max width=\textwidth}
        \begin{tabular}{lccccc}
            \toprule
            Dataset & FM steps & FM time (s/mol.) & \oursregular steps &
            \oursregular time (s/mol.) & Speedup \\
            \midrule
            GEOM-QM9   & 50 & 1.7498  & 5 & 0.6063 & $2.89\times$ \\
            GEOM-Drugs & 50 & 15.4513 & 5 & 5.0232 & $3.08\times$ \\
            \bottomrule
        \end{tabular}
    \end{adjustbox}

    \vspace{0.25em}
    \begin{minipage}{0.97\textwidth}
        \scriptsize
        \textit{Notes.} Latency denotes the arithmetic mean of the
        per-molecule sampling-call times. For GEOM-QM9, values are the mean
        $\pm$ standard deviation of three run-level means, each computed over
        all 998 molecules. The GEOM-Drugs values are from a single complete pass
        over the fixed 100-molecule subset; run-to-run uncertainty is therefore
        not estimated. The overhead in panel (a) is computed as
        $T_{\mathrm{EnFlow},N}/T_{\mathrm{FM},N}$, and the speedup in panel (b) as
        $T_{\mathrm{FM},50}/T_{\mathrm{EnFlow},5}$; both are ratios of the
        corresponding arithmetic means. Reflow is omitted from the timing table
        because it modifies training only and leaves the one-step \oursregular
        inference graph and operation count unchanged. Absolute latencies should
        be compared only within each dataset because molecular sizes and the
        numbers of generated conformers differ between GEOM-QM9 and GEOM-Drugs.
    \end{minipage}
\end{table}

\begin{figure*}[!tbp]
    \centering
    \includegraphics[width=0.90\textwidth]{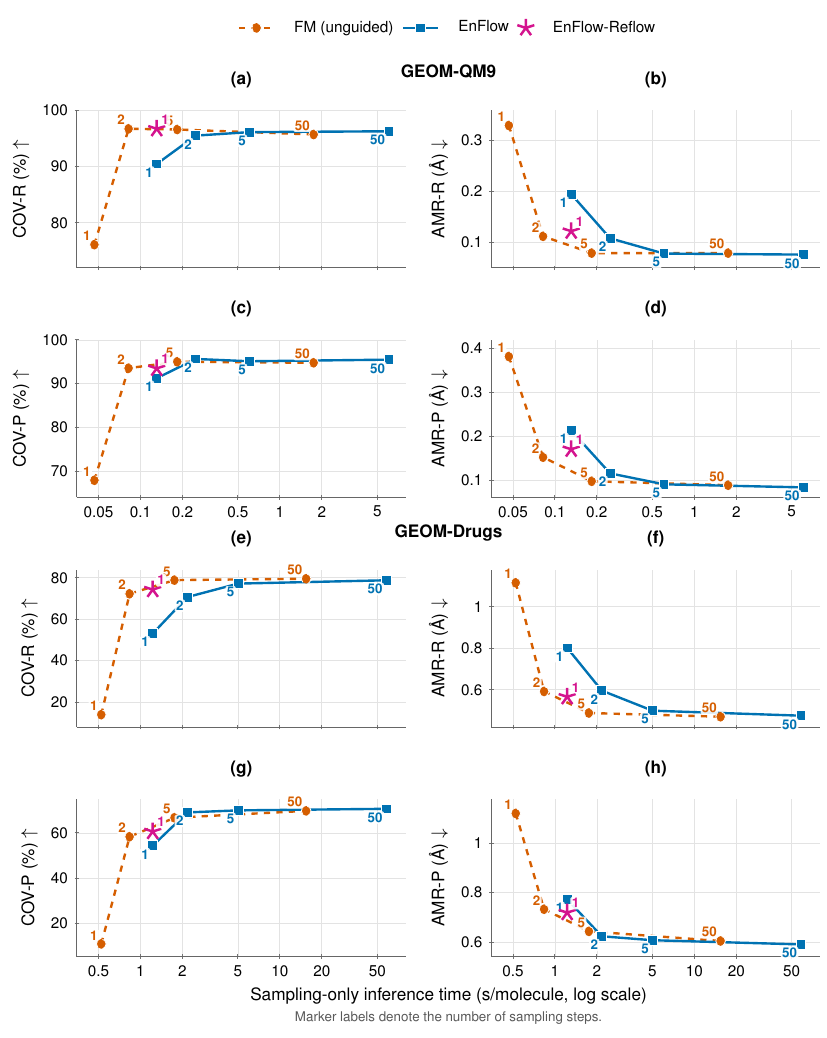}
    \caption{Quality--latency trade-offs for conformer generation on
    GEOM-QM9 (a--d) and GEOM-Drugs (e--h). Each point pairs a mean
    conformer-generation metric with mean sampling-only latency; marker labels
    denote ODE steps. COV metrics are higher-is-better and AMR metrics are
    lower-is-better. FM is unguided sampling, and EnFlow-Reflow is plotted at the
    measured one-step \oursregular latency because Reflow changes training but
    not the one-step inference graph. Coverage uses $0.5$~\AA{} for GEOM-QM9 and
    $0.75$~\AA{} for GEOM-Drugs.}
    \label{fig:inference_quality_tradeoff_combined}
\end{figure*}

\autoref{tab:sampling_efficiency_guidance_overhead}(a) shows that the
\oursregular/FM latency ratio ranges from $2.80\times$ to $3.47\times$ on
GEOM-QM9 and from $2.34\times$ to $3.76\times$ on GEOM-Drugs. Guidance therefore
does not reduce the cost of an individual integration step; its advantage is to
improve conformer quality under a restricted ODE budget. Practical sampling
efficiency should thus be assessed at quality-comparable operating points.

As shown in
\autoref{tab:sampling_efficiency_guidance_overhead}(b), five-step \oursregular
closely matches 50-step FM on GEOM-QM9
($96.11/0.078/95.11/0.090$ versus $95.69/0.079/94.77/0.088$ for
COV-R/AMR-R/COV-P/AMR-P), while reducing latency from $1.7498$ to $0.6063$
seconds per molecule ($2.89\times$). On GEOM-Drugs, five-step \oursregular is
quality-comparable, but not uniformly superior, to 50-step FM
($77.20/0.499/70.00/0.607$ versus $79.54/0.470/69.79/0.604$), and reduces
latency from $15.4513$ to $5.0232$ seconds per molecule ($3.08\times$). Thus,
the reduced step count offsets the guidance overhead at these operating points;
\autoref{fig:inference_quality_tradeoff_combined} shows the complete trade-off.

Because Reflow changes training but not the one-step inference graph, its
operating points use the measured one-step \oursregular latency. Relative to
one-step \oursregular, EnFlow-Reflow improves COV-R/AMR-R/COV-P/AMR-P from
$90.45/0.195/91.23/0.213$ to $96.70/0.122/93.48/0.170$ on GEOM-QM9 and from
$53.10/0.802/54.30/0.773$ to $74.30/0.566/60.60/0.719$ on GEOM-Drugs, without
increasing the inference operation count. Overall, guidance incurs a per-step
cost but can improve sampling-stage efficiency by reducing the steps required
for high-quality generation. The reported $2.89\times$ and $3.08\times$ gains
refer specifically to the quality-comparable operating points above; among the
evaluated budgets, five-step \oursregular offers the most favorable empirical
quality--latency balance.

\clearpage
\endgroup

   \begin{figure*}[t]
   \centering
   \vspace{-20pt}
   \par
   \smallskip
   \makebox[\textwidth]{\footnotesize{(a)}}
   \includegraphics[width=\textwidth]{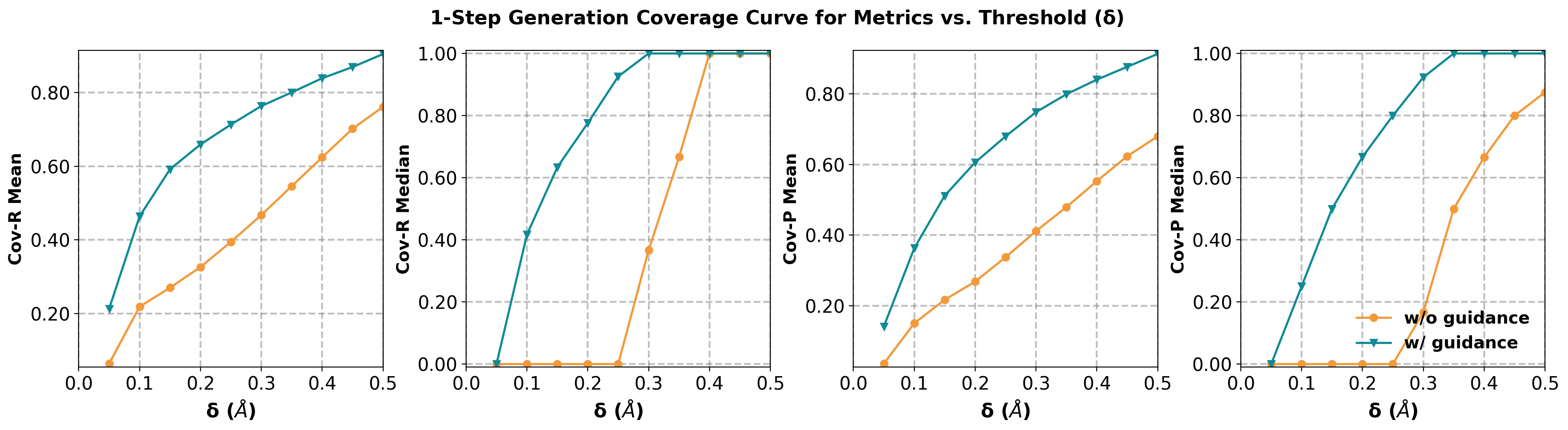}
   \vspace{-20pt}
   \par
   \smallskip
   \makebox[\textwidth]{\footnotesize{(b)}}
   \includegraphics[width=\textwidth]{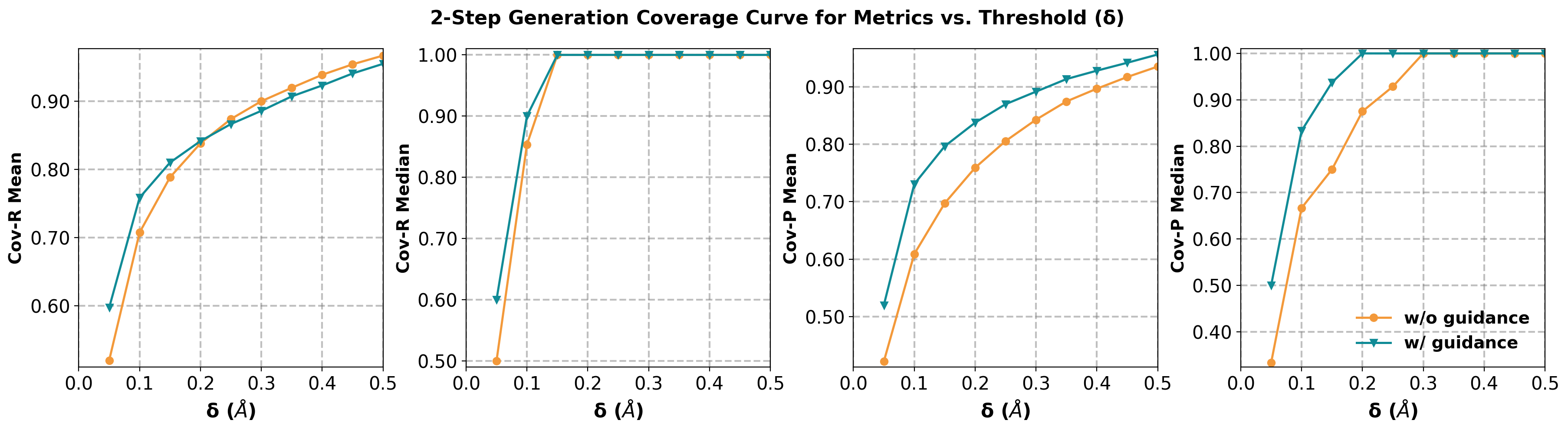}
   \vspace{-20pt}
   \par
   \smallskip
   \makebox[\textwidth]{\footnotesize{(c)}}
   \includegraphics[width=\textwidth]{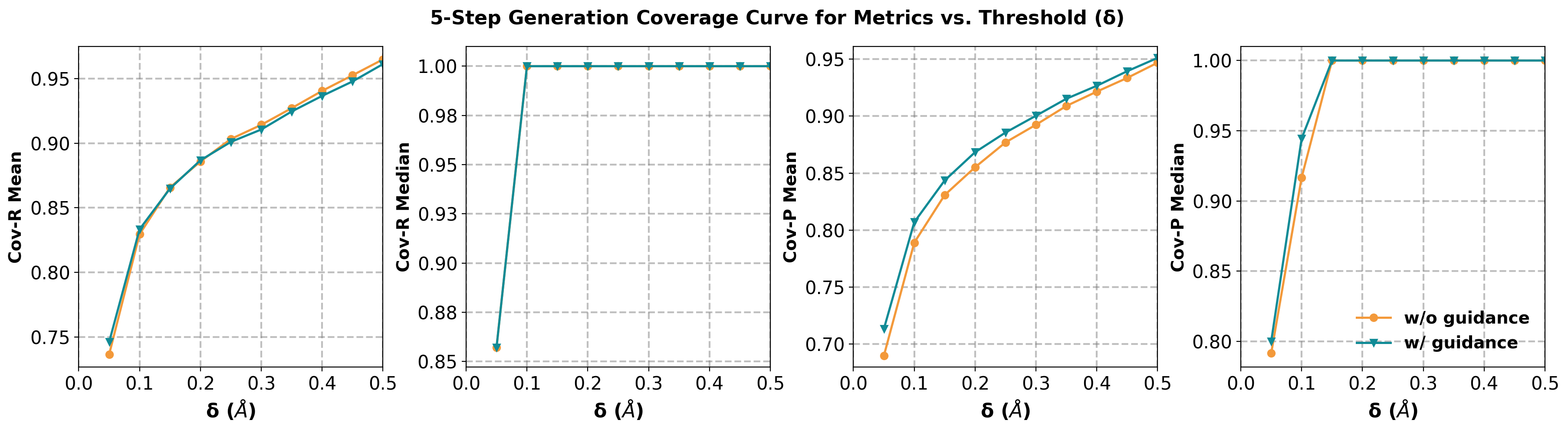}
   \vspace{-15pt}
   \caption{Ablation study of coverage (\%) vs.\ threshold $\boldsymbol{\delta}$
   on the GEOM-QM9 dataset for ODE sampling with 1 (a), 2 (b) and 5 (c) steps, comparing
   the unguided baseline (ET-Flow; \textbf{w/o guidance}) with the guided model (\ours;
   \textbf{w/ guidance}).}
   
\end{figure*}

\begin{figure*}[t]
   \centering
   \vspace{-20pt}
   \par
   \smallskip
   \makebox[\textwidth]{\footnotesize{(a)}}
   \includegraphics[width=\textwidth]{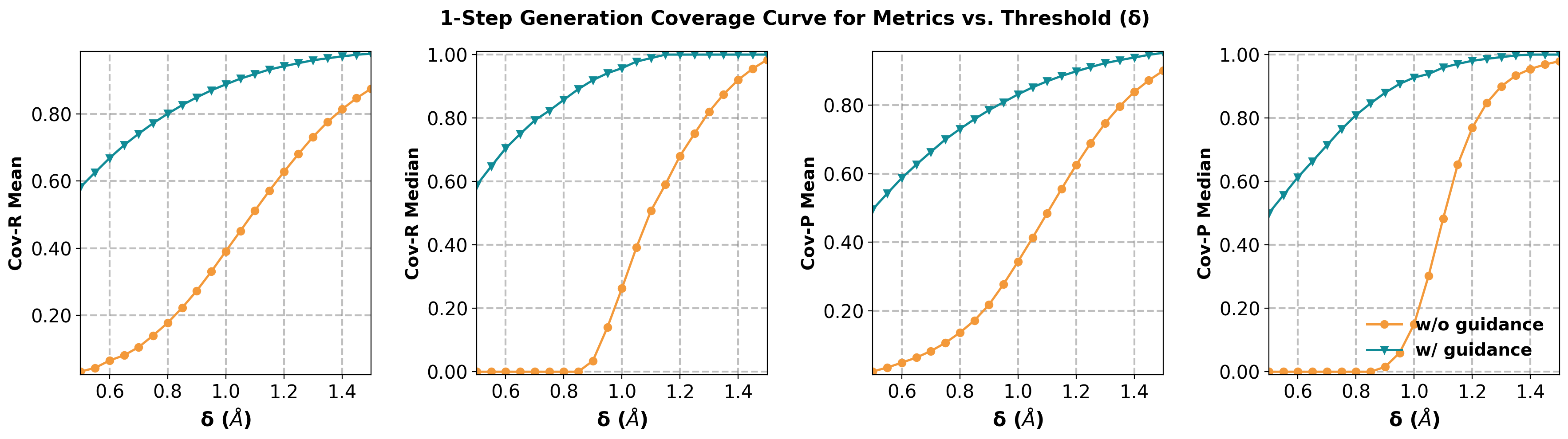}
   \vspace{-20pt}
   \par
   \smallskip
   \makebox[\textwidth]{\footnotesize{(b)}}
   \includegraphics[width=\textwidth]{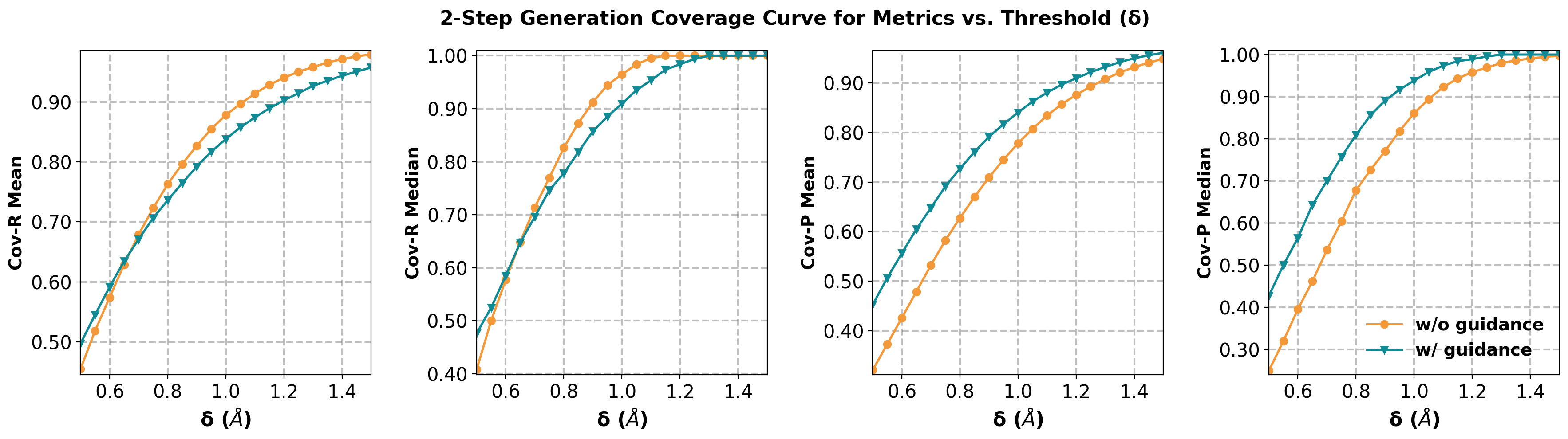}
   \vspace{-20pt}
   \par
   \smallskip
   \makebox[\textwidth]{\footnotesize{(c)}}
   \includegraphics[width=\textwidth]{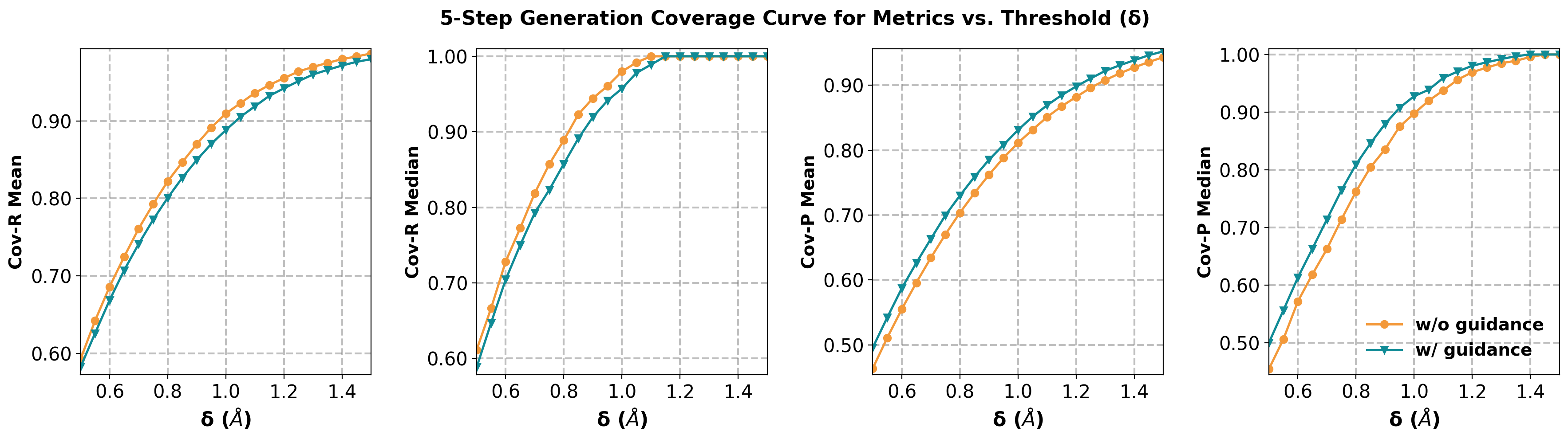}
   \vspace{-20pt}
   \par
   \smallskip
   \makebox[\textwidth]{\footnotesize{(d)}}
   \includegraphics[width=\textwidth]{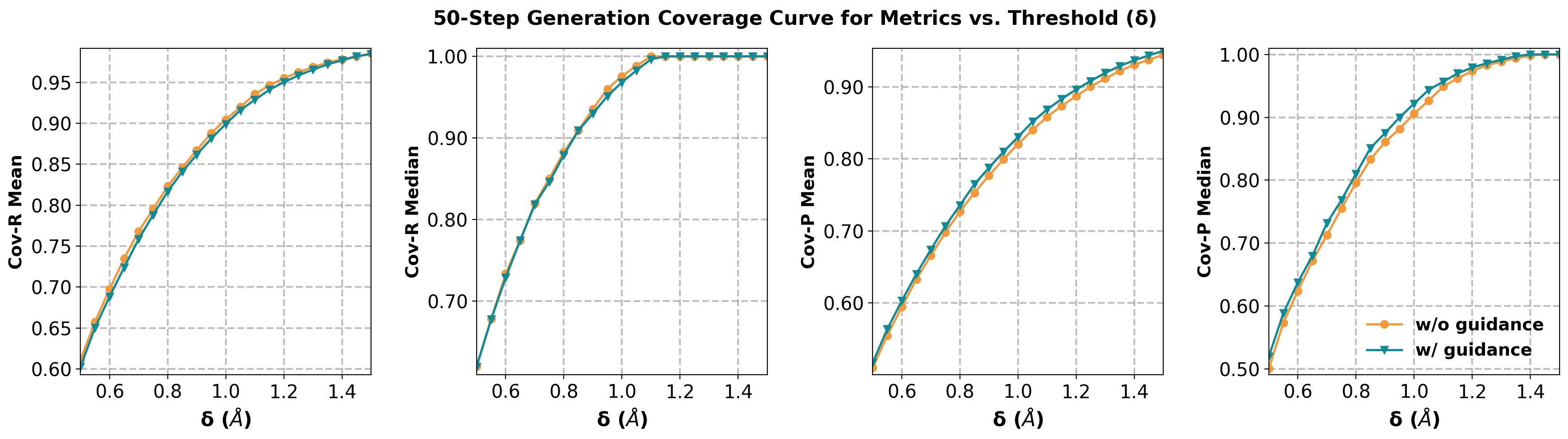}
   \vspace{-15pt}
   \caption{Ablation study of coverage (\%) vs.\ threshold $\boldsymbol{\delta}$
   on the GEOM-Drugs dataset for ODE sampling with 1 (a), 2 (b), 5 (c), and 50 (d)
   steps, comparing the unguided baseline (ET-Flow; \textbf{w/o guidance}) with the
   guided model (\ours; \textbf{w/ guidance}).}
   
\end{figure*}

\begin{figure*}[t]
   \centering
   \includegraphics[width=\dimexpr\textwidth/3\relax]{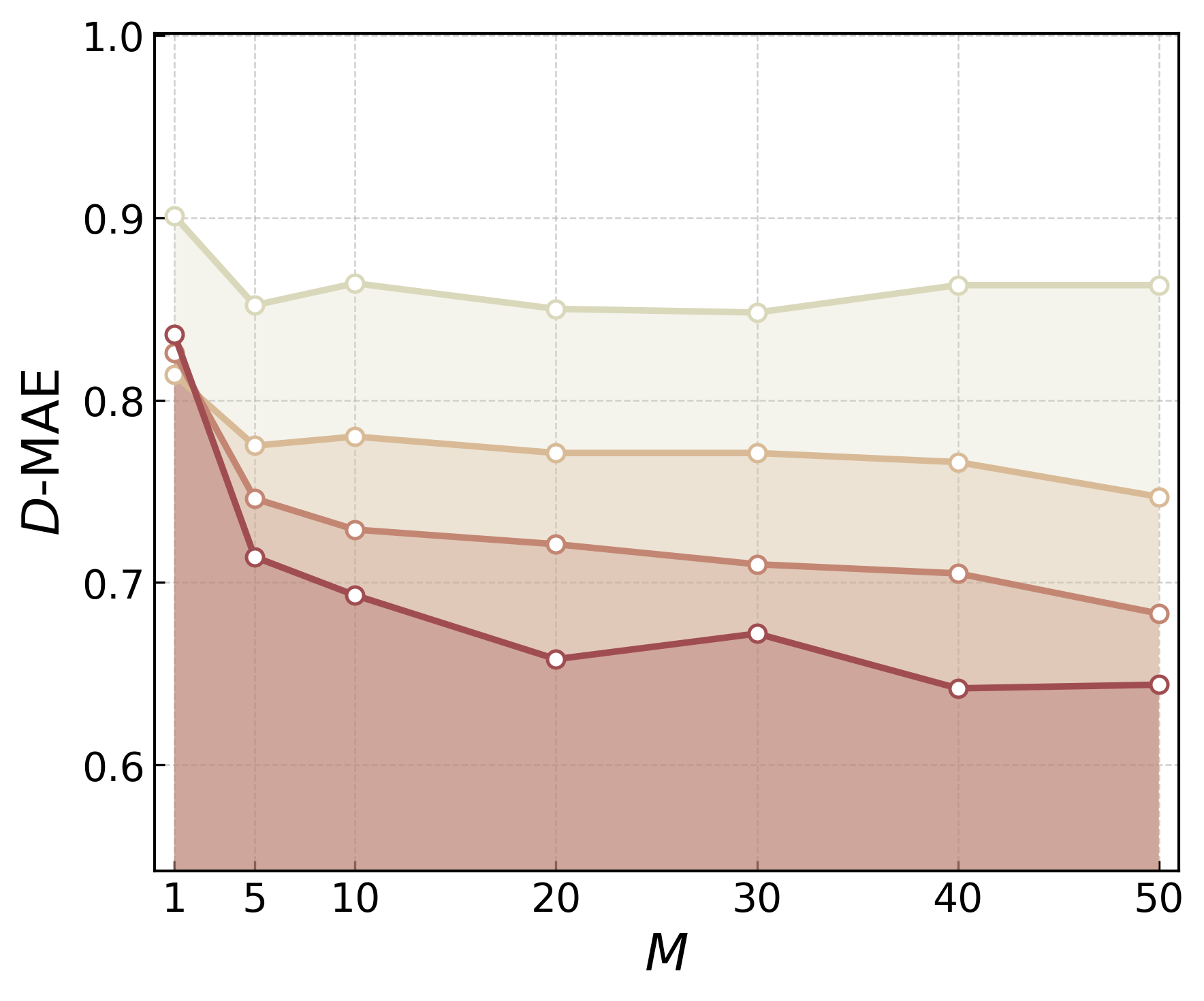}%
   \includegraphics[width=\dimexpr\textwidth/3\relax]{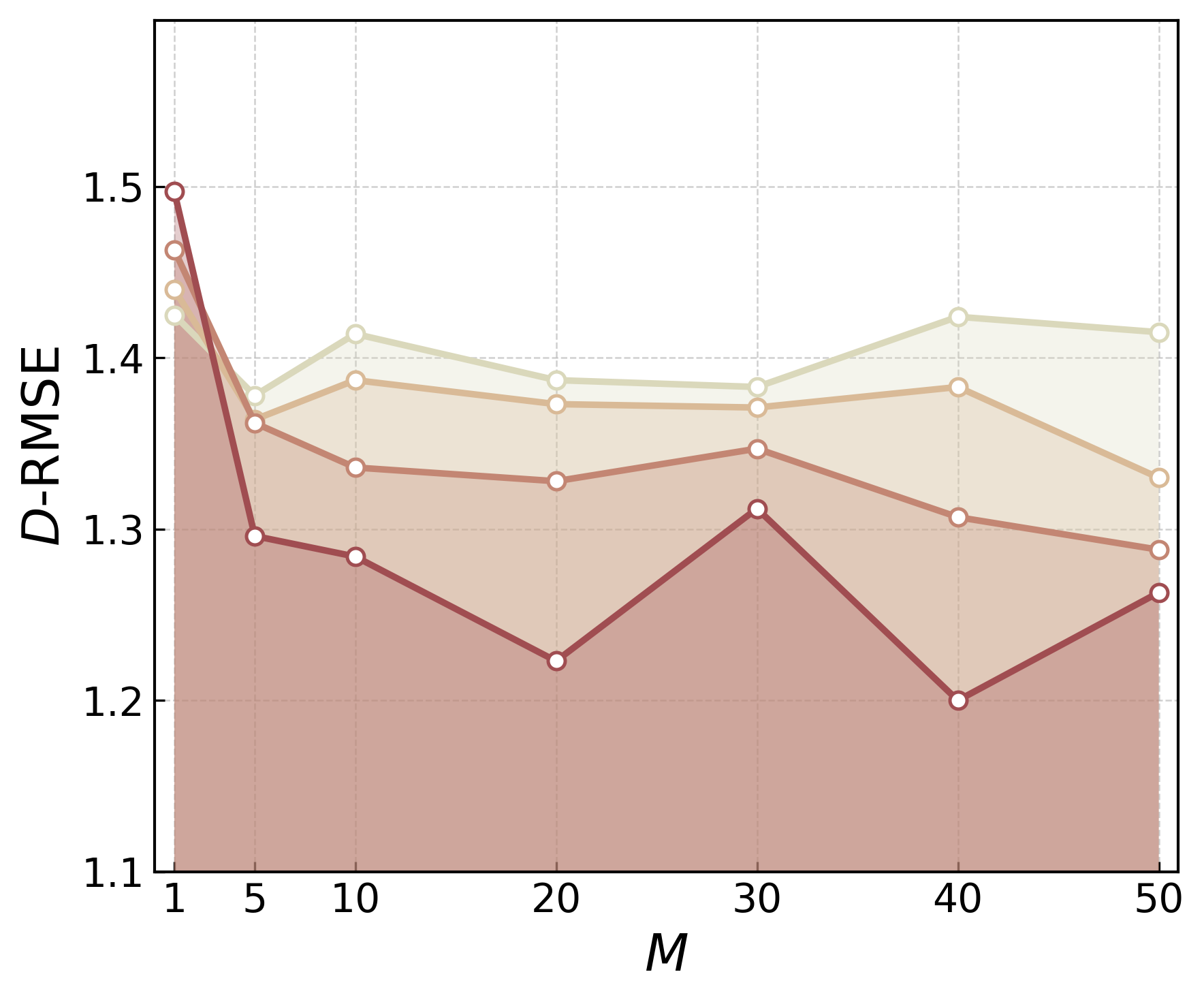}%
   \includegraphics[width=\dimexpr\textwidth/3\relax]{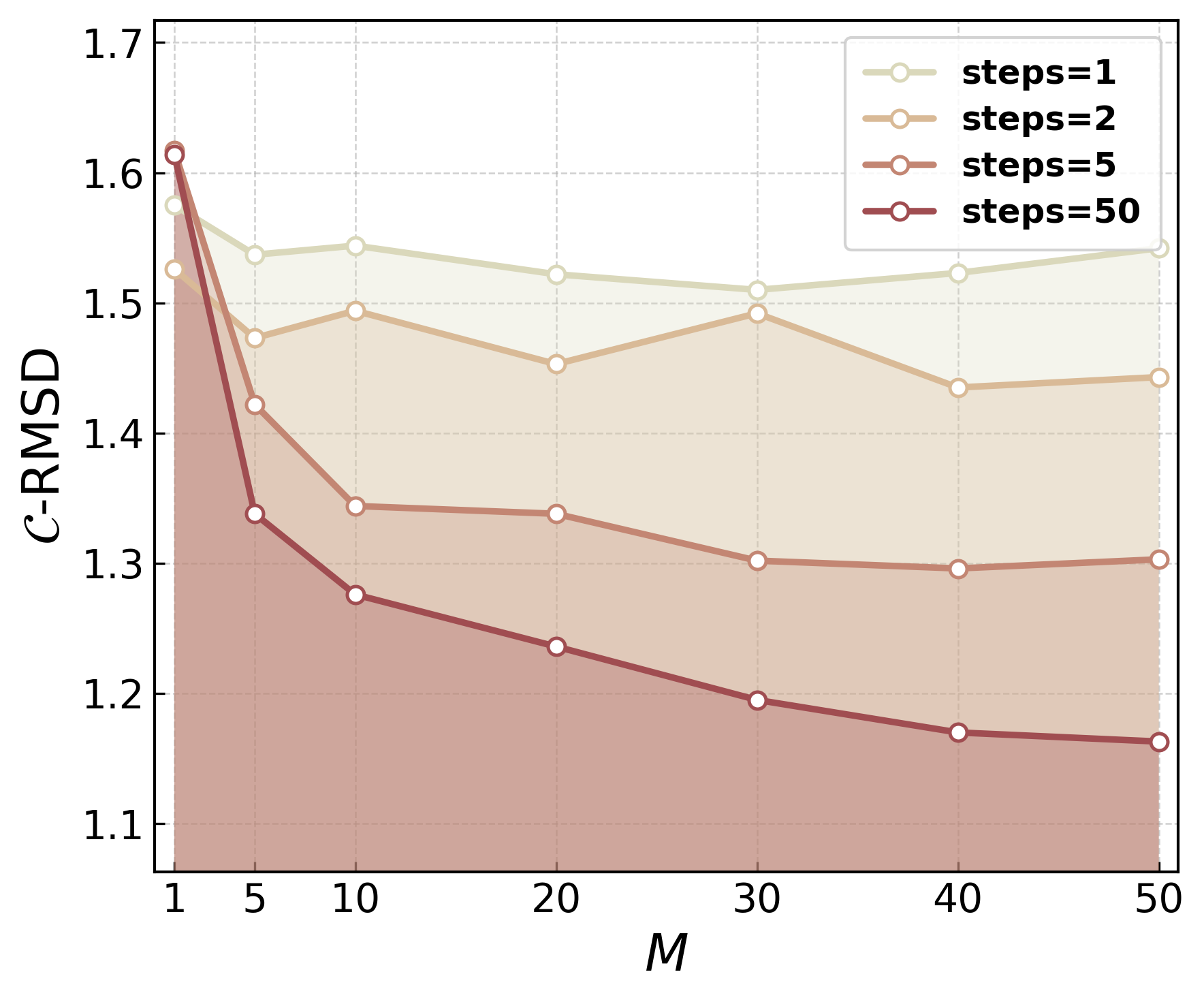}
   \caption{Ablation of \texttt{EnergyRank} mode for ground-state conformation
   prediction on the GEOM-Drugs dataset. Effect of ensemble size
   $M = 1, 5, 10, 20, 50$ under $1, 2, 5,$ and $50$ ODE sampling steps. From left
   to right: $\mathbf{D}\text{-MAE}$ (Å), $\mathbf{D}\text{-RMSE}$ (Å),
   $\mathcal{C}\text{-RMSD}$ (Å).}
   
\end{figure*}

\begin{figure*}[t]
   \centering
   \includegraphics[width=.32\textwidth]{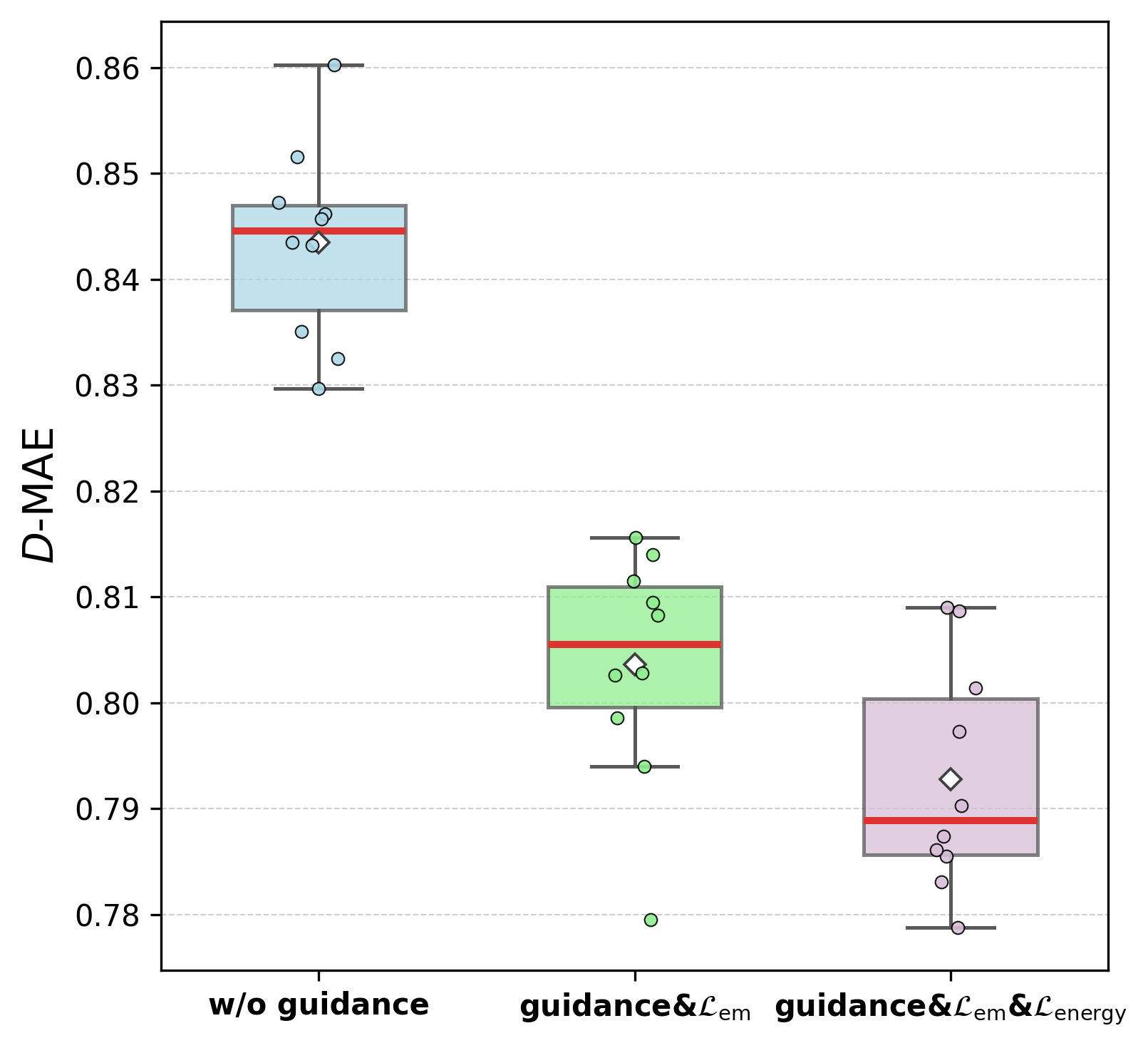}%
   \hfill
   \includegraphics[width=.32\textwidth]{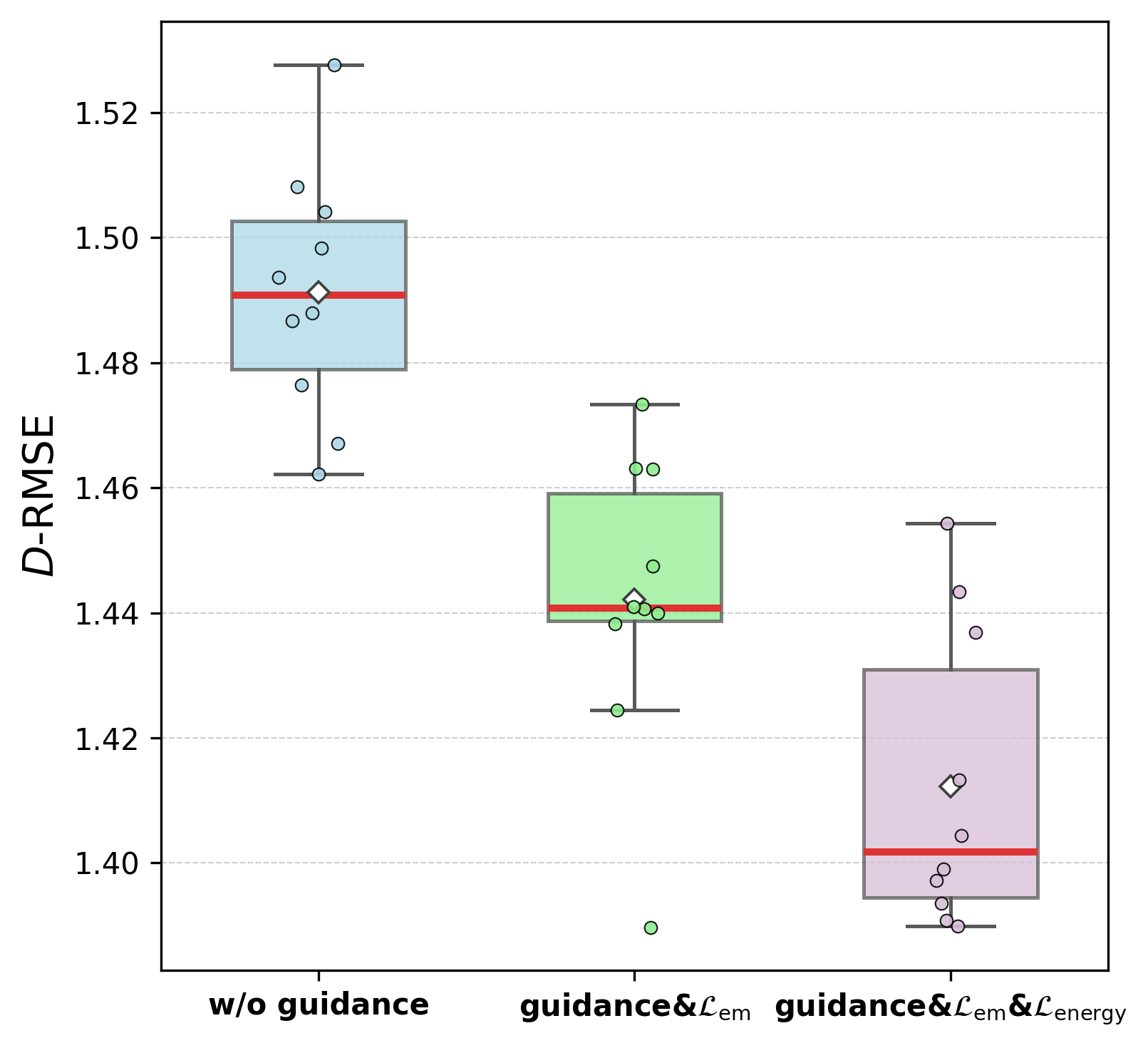}%
   \hfill
   \includegraphics[width=.32\textwidth]{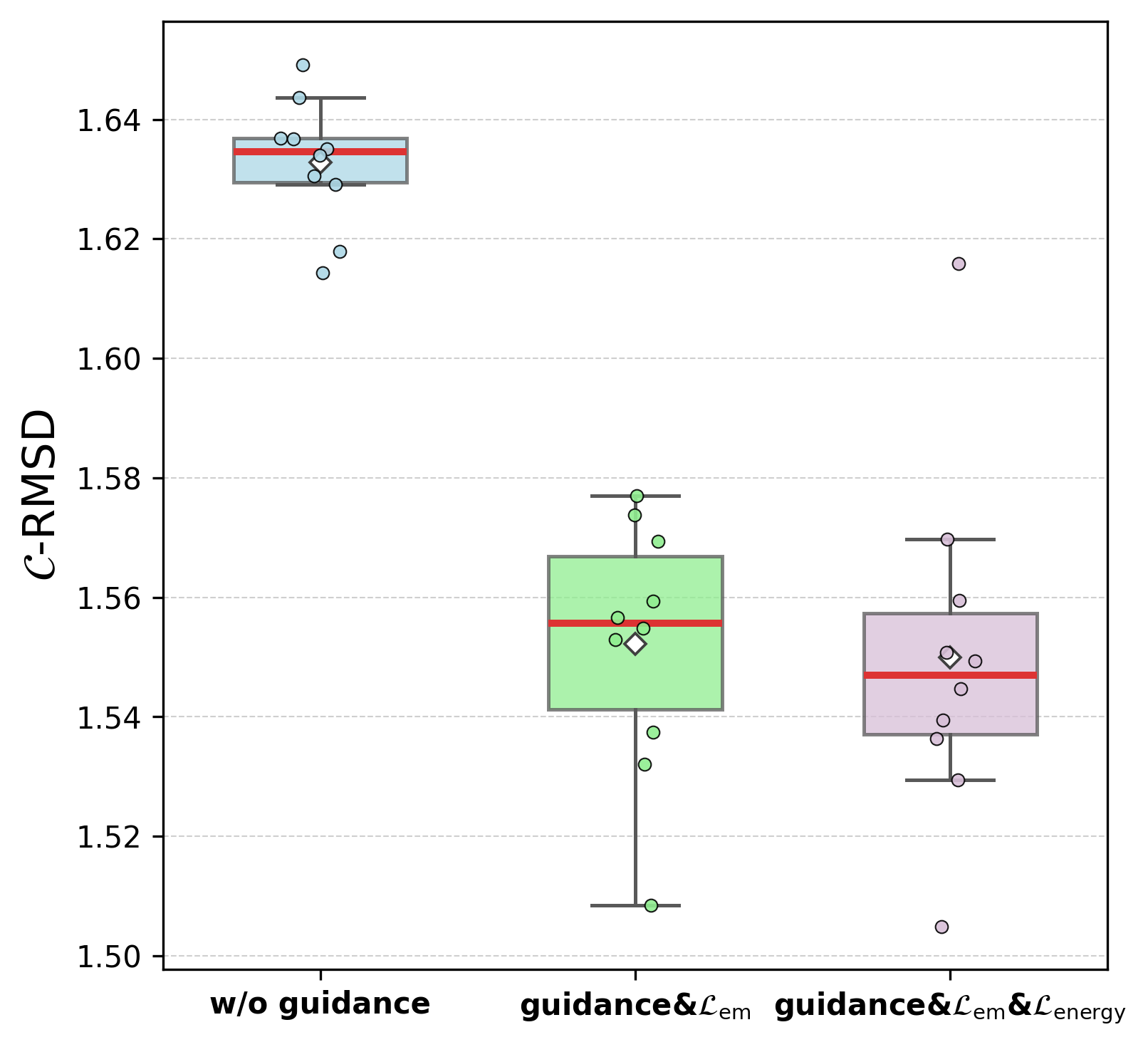}
   \caption{With \texttt{JustFM} mode and 5-step ODE sampling, boxplots of
   ground-state conformation prediction performance under three settings: (1) unguided
   baseline (ET-Flow; \textbf{w/o guidance}); (2) guided model with energy
   matching only (\textbf{guidance} \& $\mathcal{L}_{\mathrm{em}}$); and (3)
   fully guided model with energy matching and energy fine-tuning (\ours;
   \textbf{guidance} \& $\mathcal{L}_{\mathrm{em}}$ \& $\mathcal{L}_{\mathrm{energy}}$).}
   \label{fig: ground_state_guidance_ablation_drugs_o3}
\end{figure*}

\begin{figure*}[p]
   \centering
   \setlength{\tabcolsep}{0pt}
   \renewcommand{\arraystretch}{0}
   \begin{tabular*}{\textwidth}{@{\extracolsep{\fill}}cccc@{}}
      (a) & (b) & (c) & (d) \\[3pt]
      \includegraphics[width=.215\textwidth]{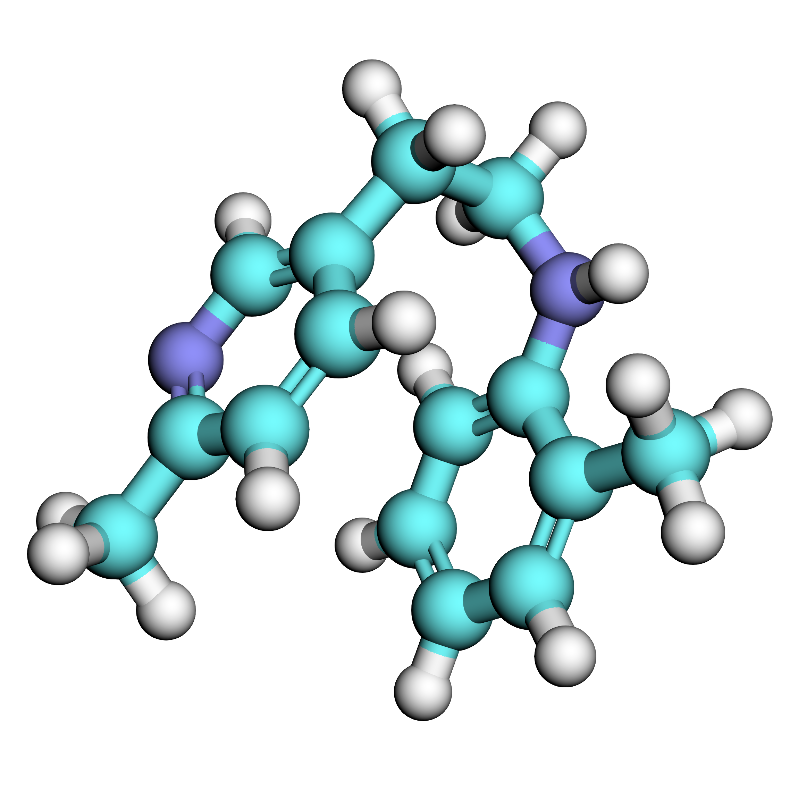} &
      \includegraphics[width=.215\textwidth]{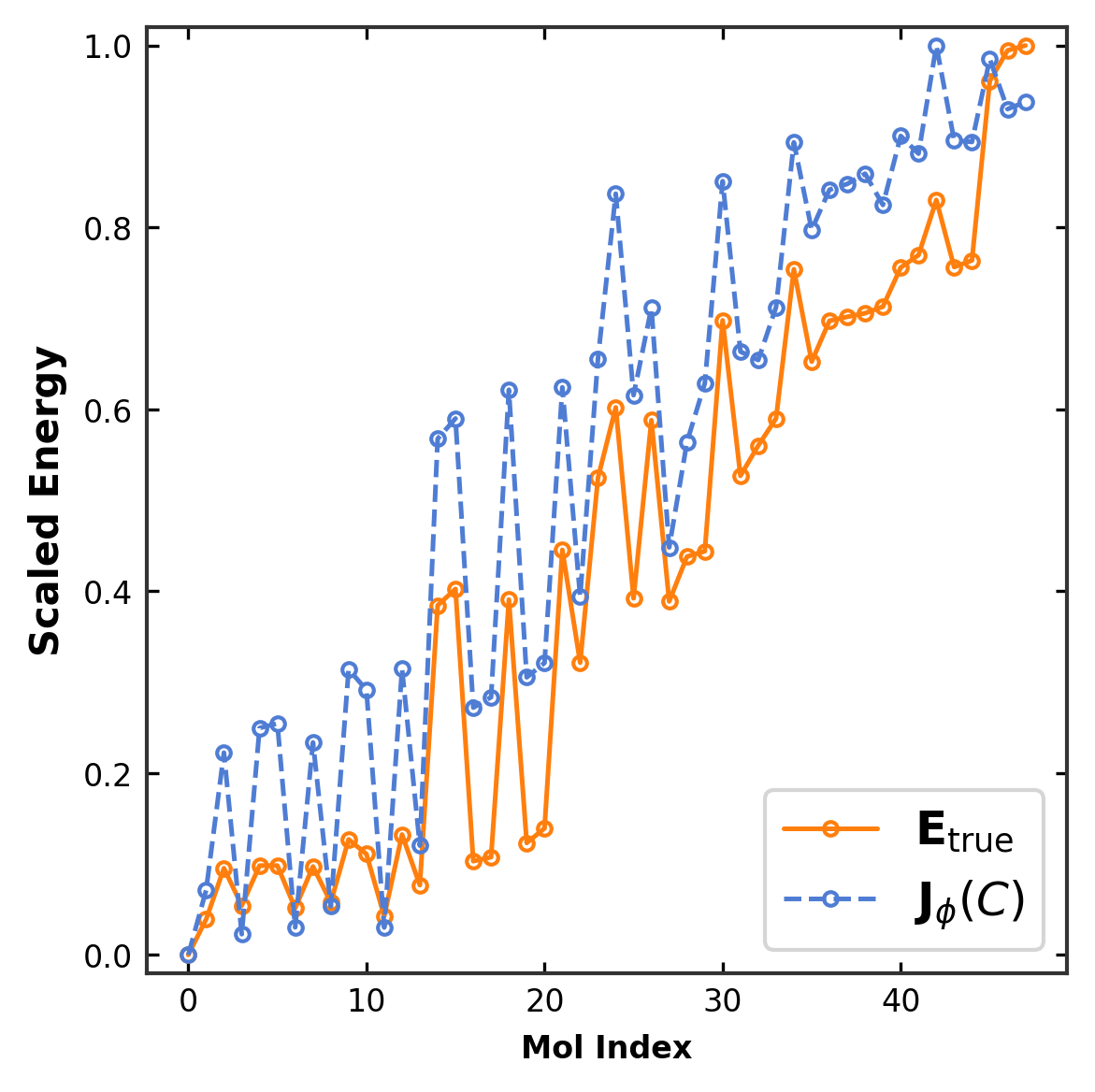} &
      \includegraphics[width=.215\textwidth]{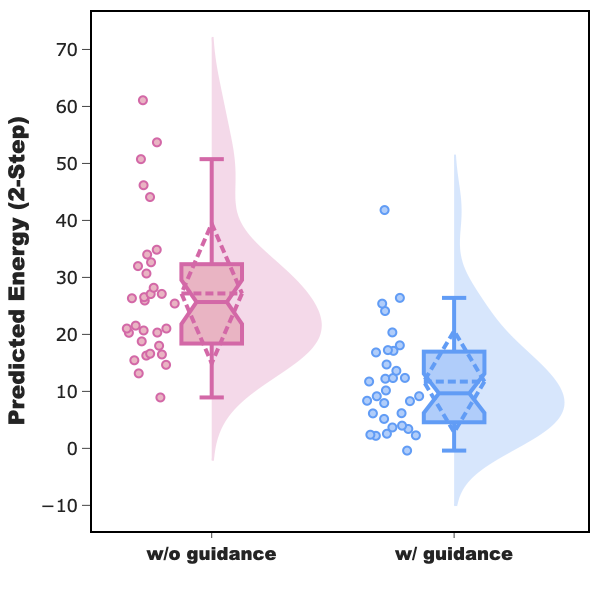} &
      \includegraphics[width=.215\textwidth]{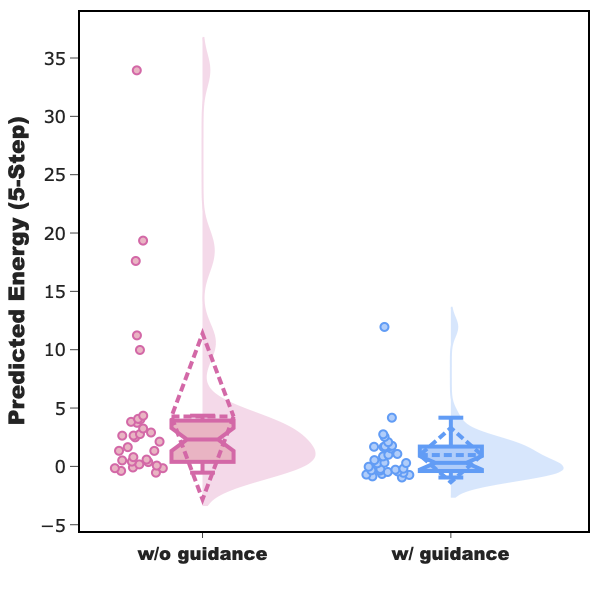} \\[1.5pt]
      \includegraphics[width=.215\textwidth]{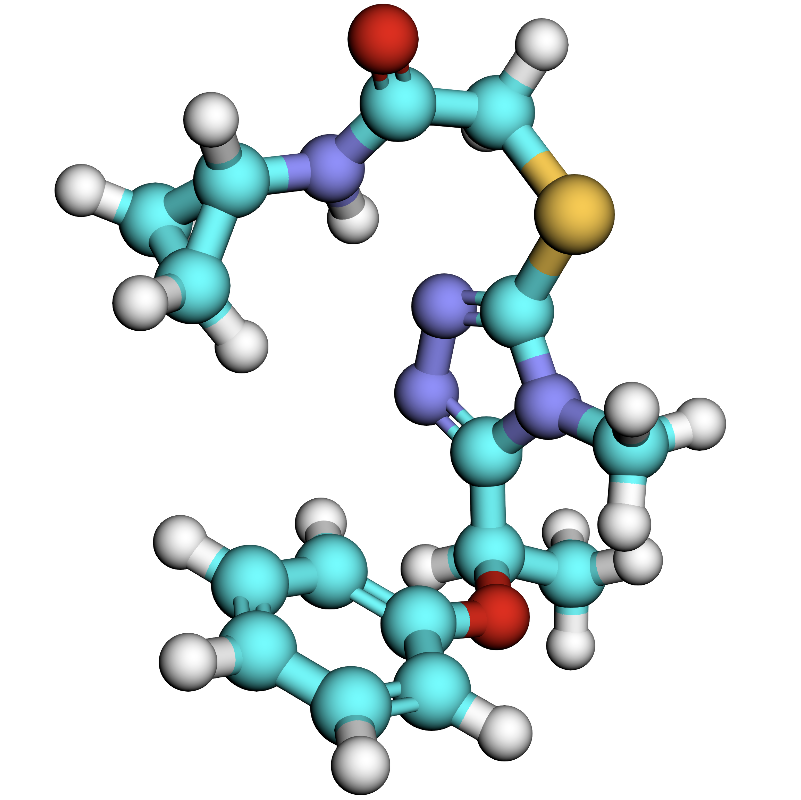} &
      \includegraphics[width=.215\textwidth]{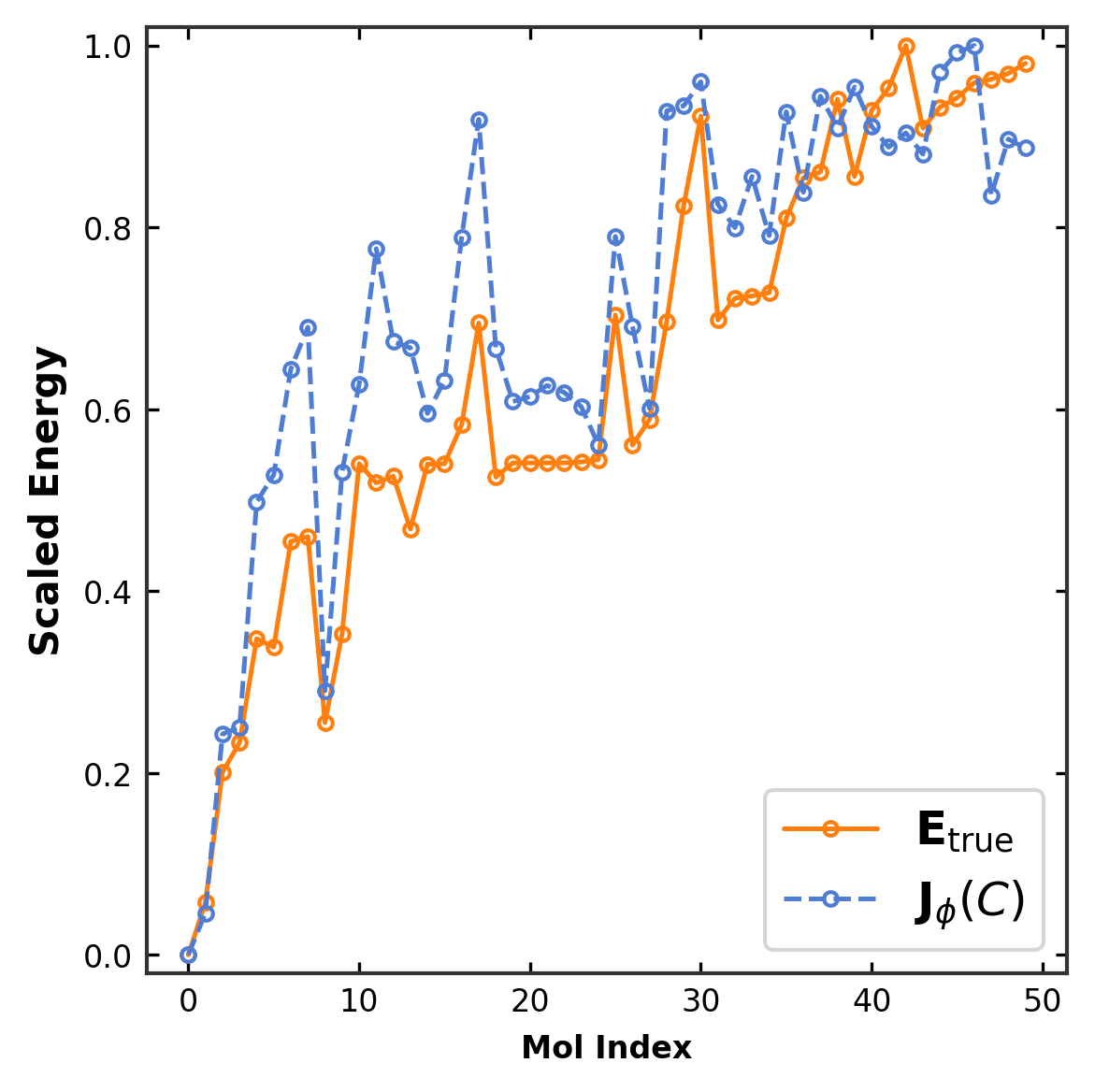} &
      \includegraphics[width=.215\textwidth]{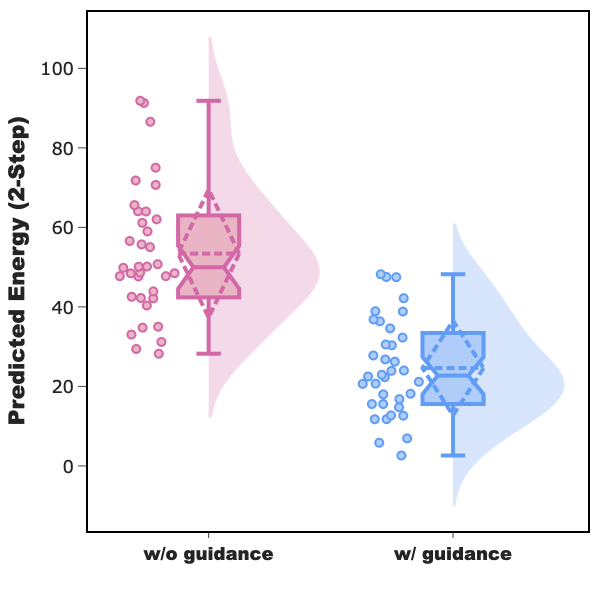} &
      \includegraphics[width=.215\textwidth]{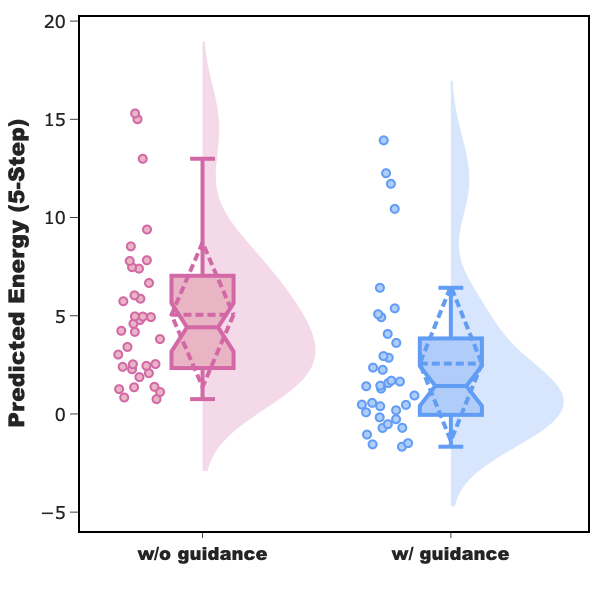} \\[1.5pt]
      \includegraphics[width=.215\textwidth]{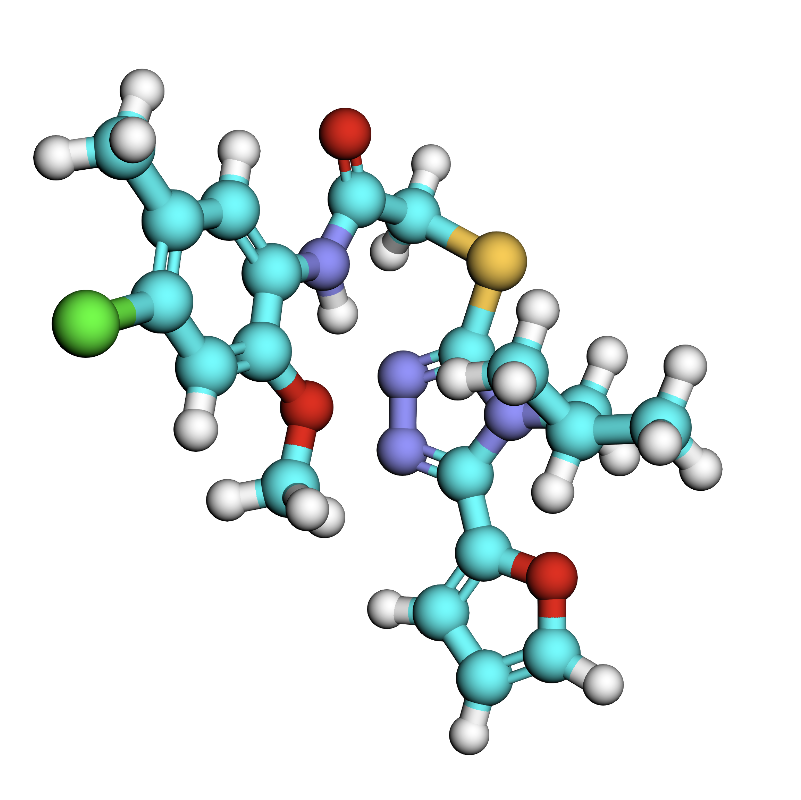} &
      \includegraphics[width=.215\textwidth]{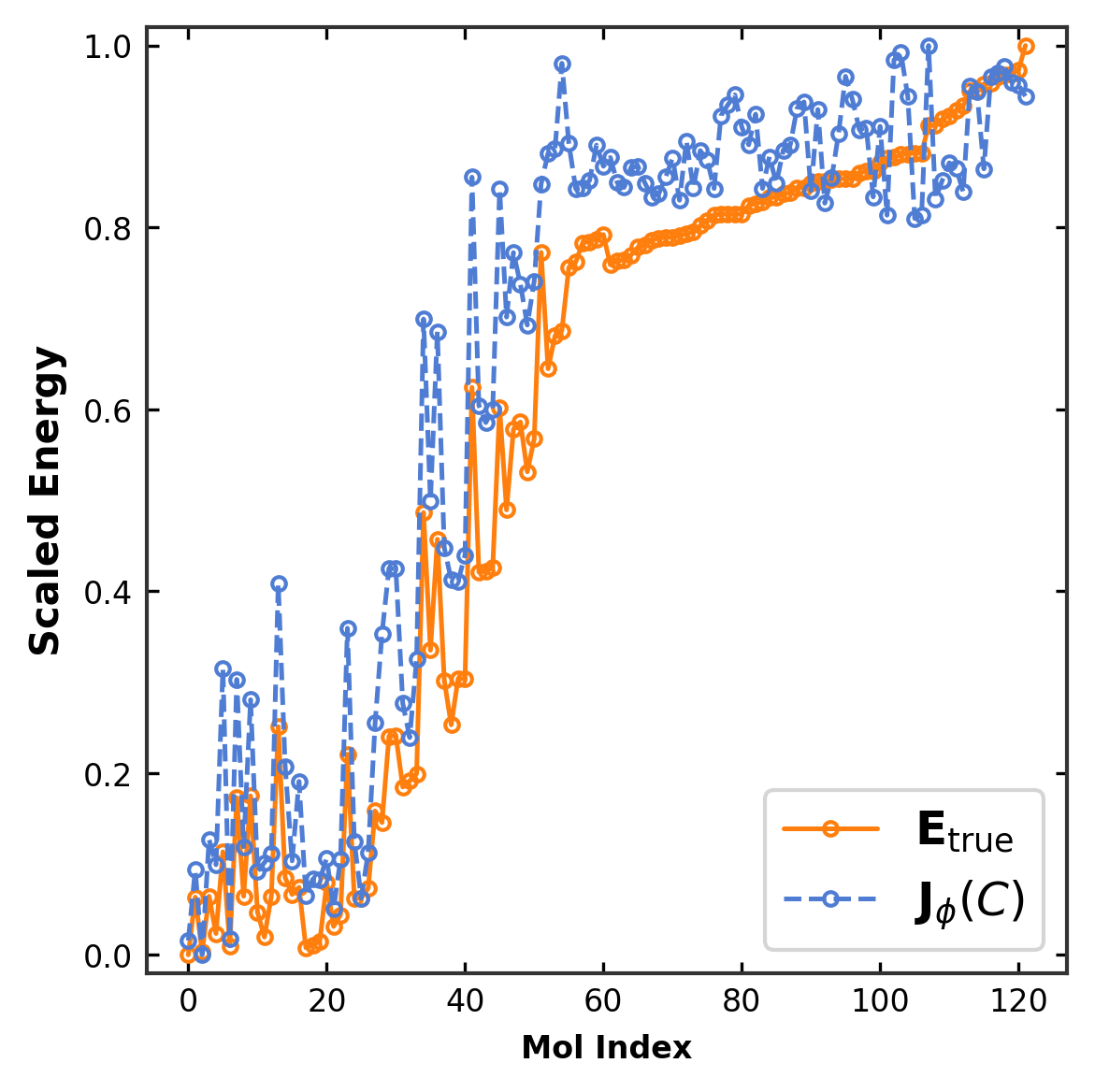} &
      \includegraphics[width=.215\textwidth]{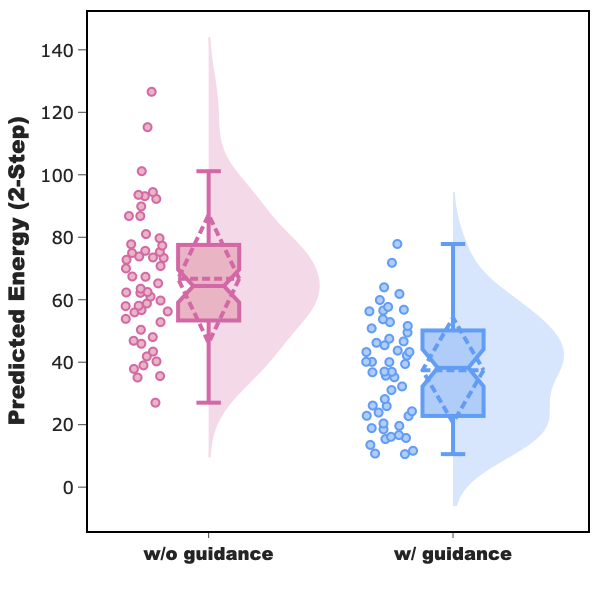} &
      \includegraphics[width=.215\textwidth]{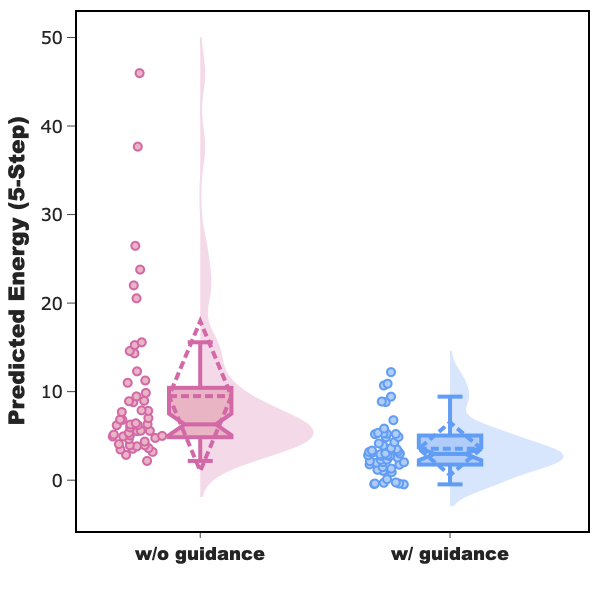} \\[1.5pt]
      \includegraphics[width=.215\textwidth]{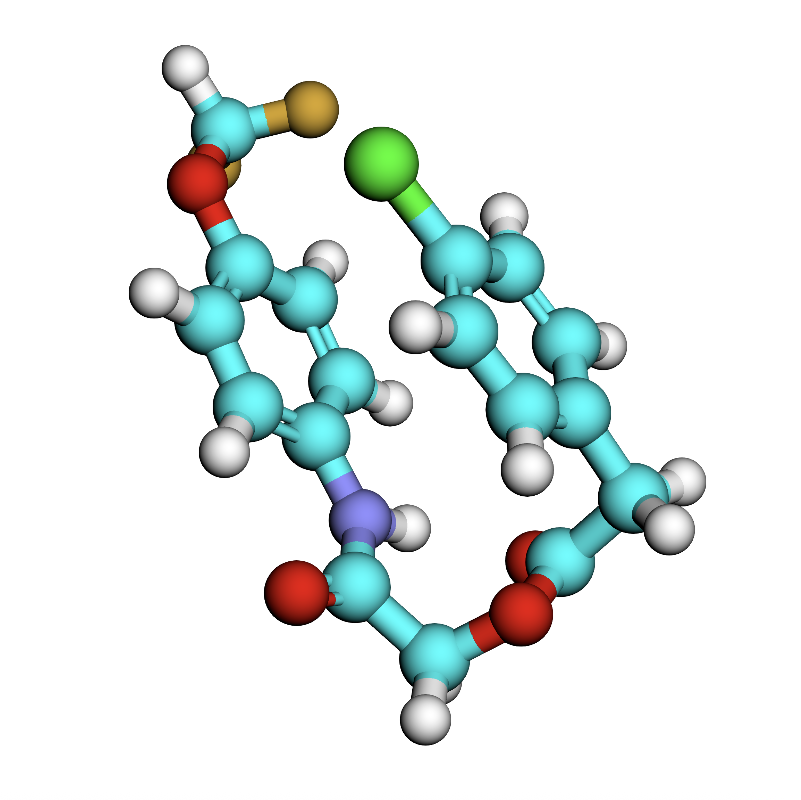} &
      \includegraphics[width=.215\textwidth]{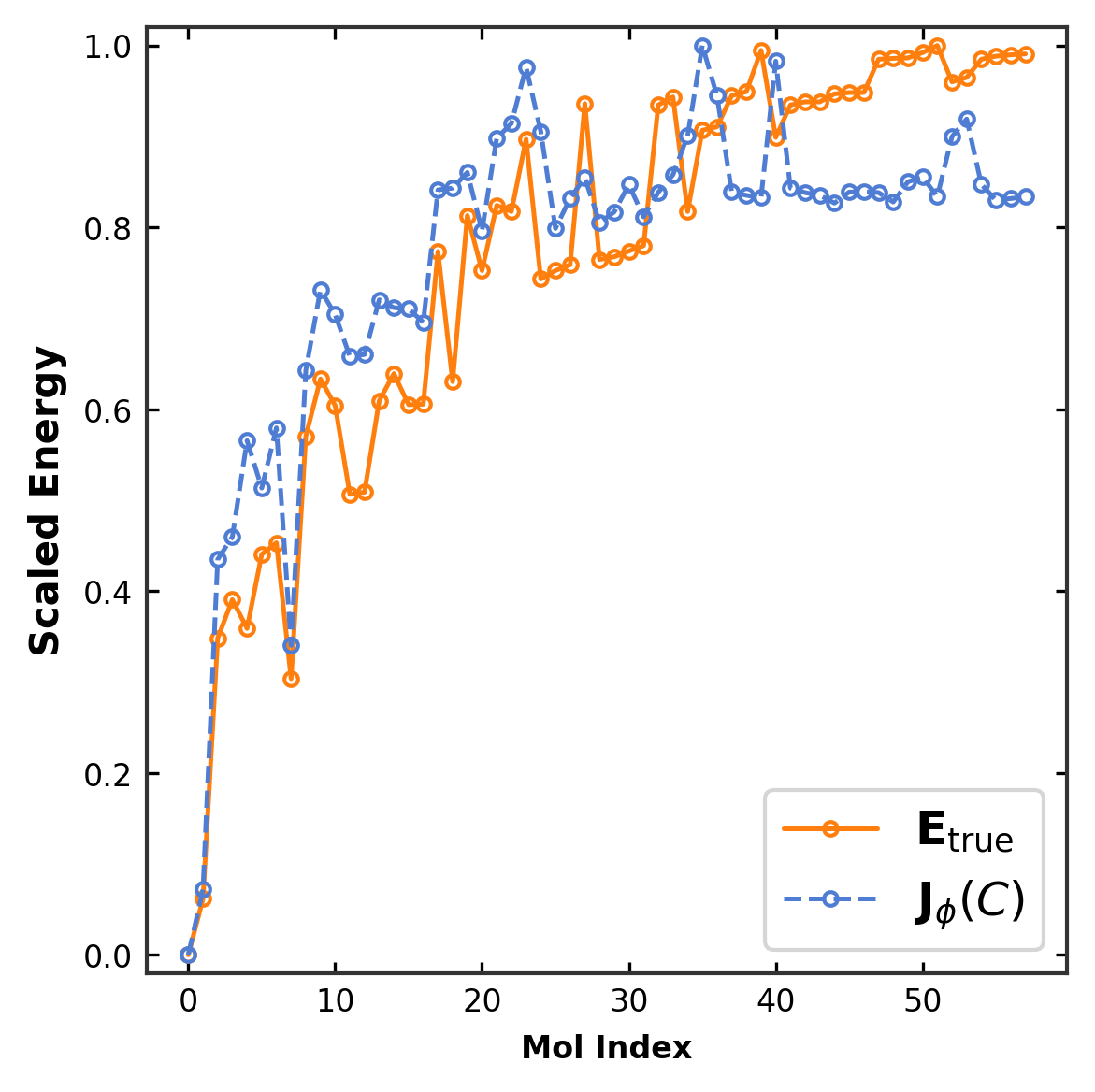} &
      \includegraphics[width=.215\textwidth]{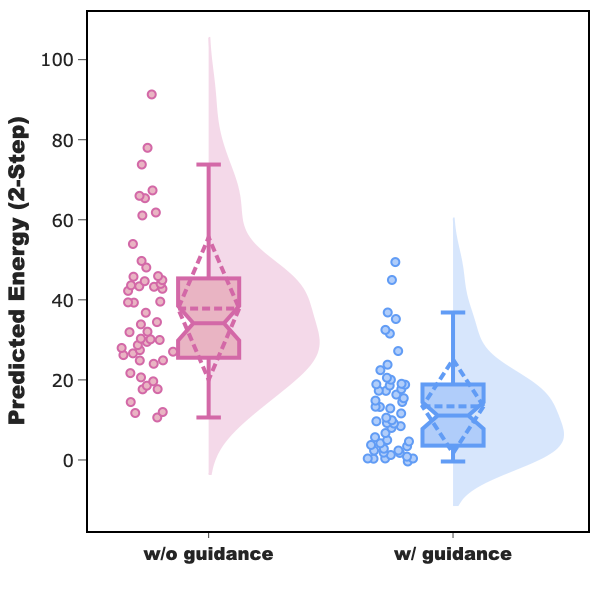} &
      \includegraphics[width=.215\textwidth]{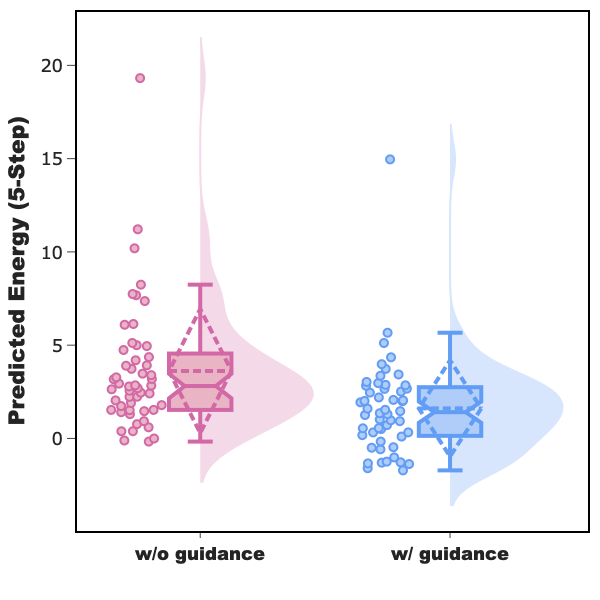} \\[1.5pt]
      \includegraphics[width=.215\textwidth]{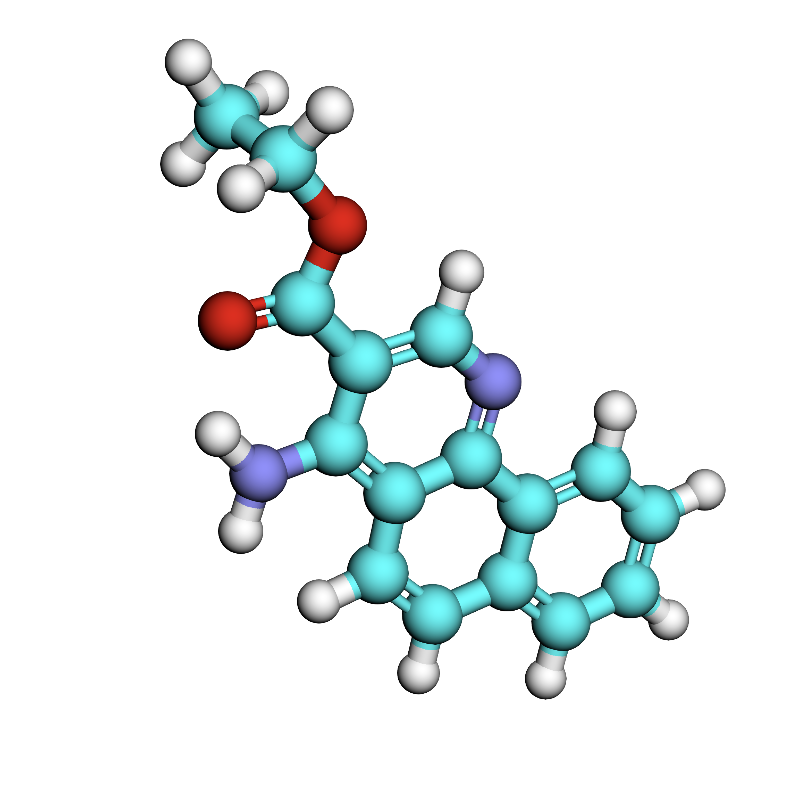} &
      \includegraphics[width=.215\textwidth]{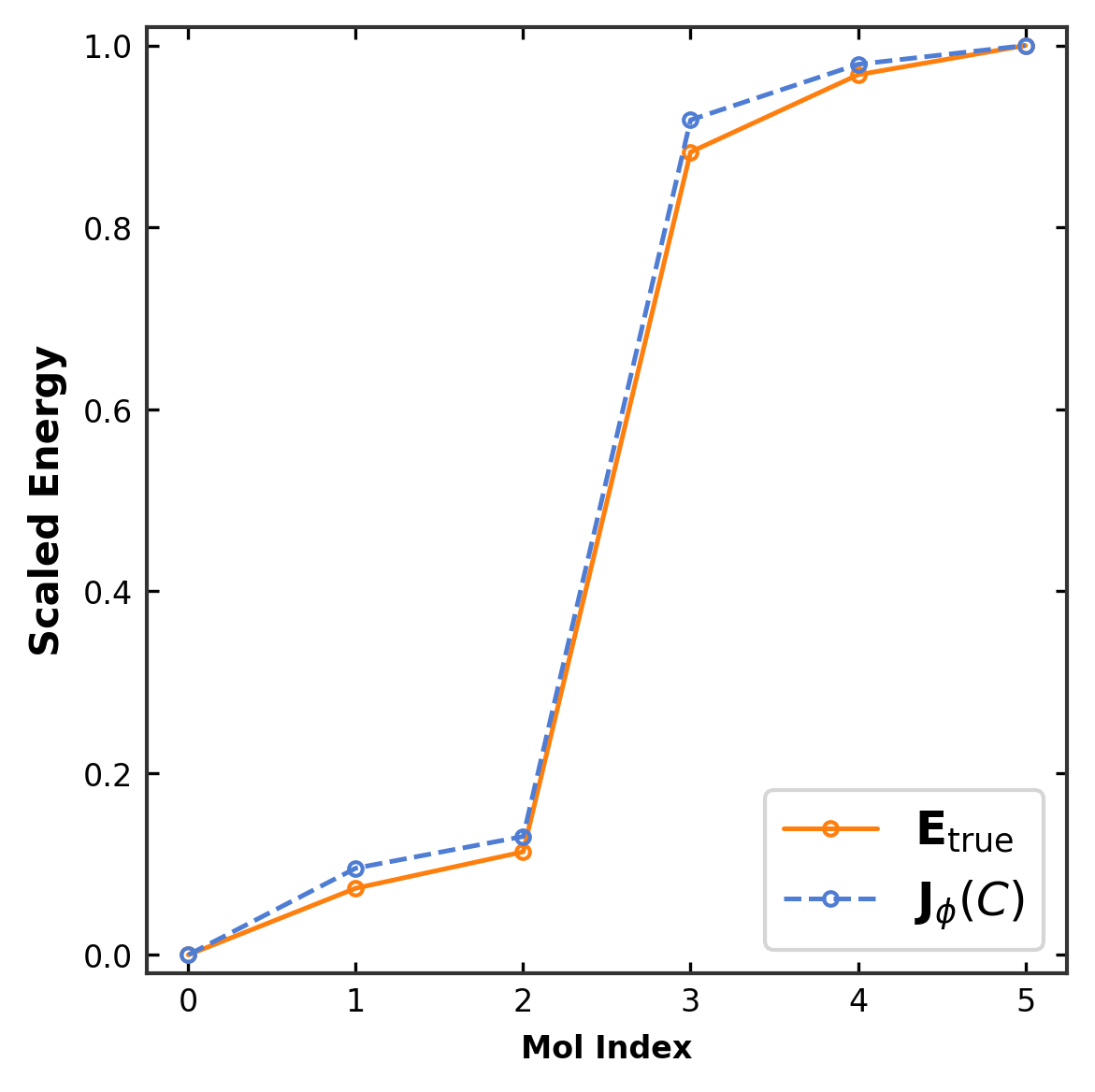} &
      \includegraphics[width=.215\textwidth]{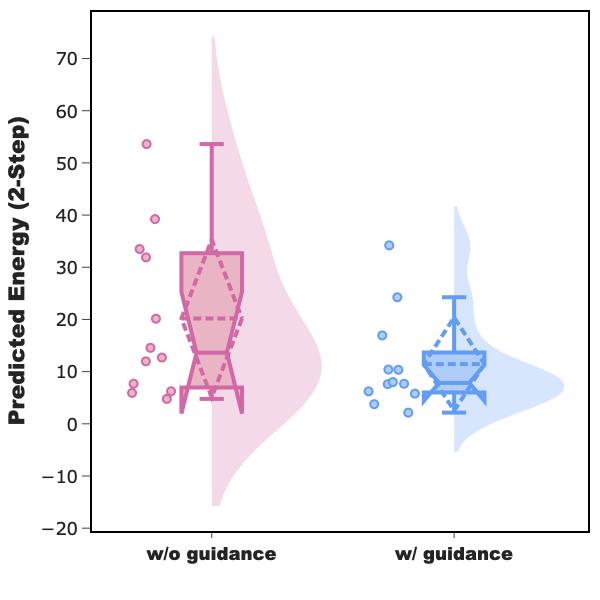} &
      \includegraphics[width=.215\textwidth]{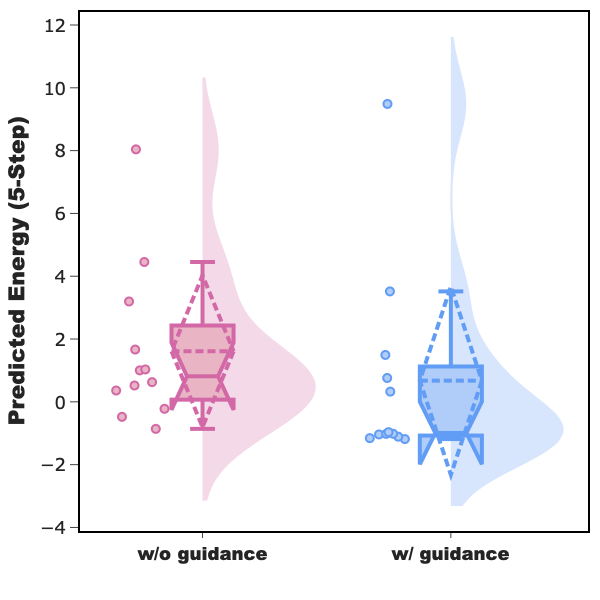} \\[1.5pt]
      \includegraphics[width=.215\textwidth]{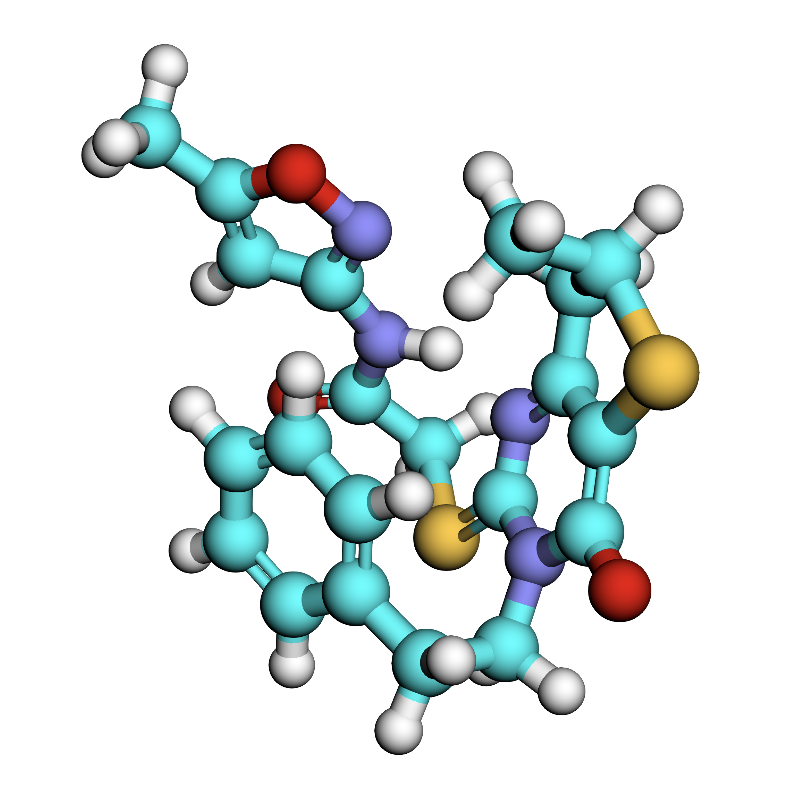} &
      \includegraphics[width=.215\textwidth]{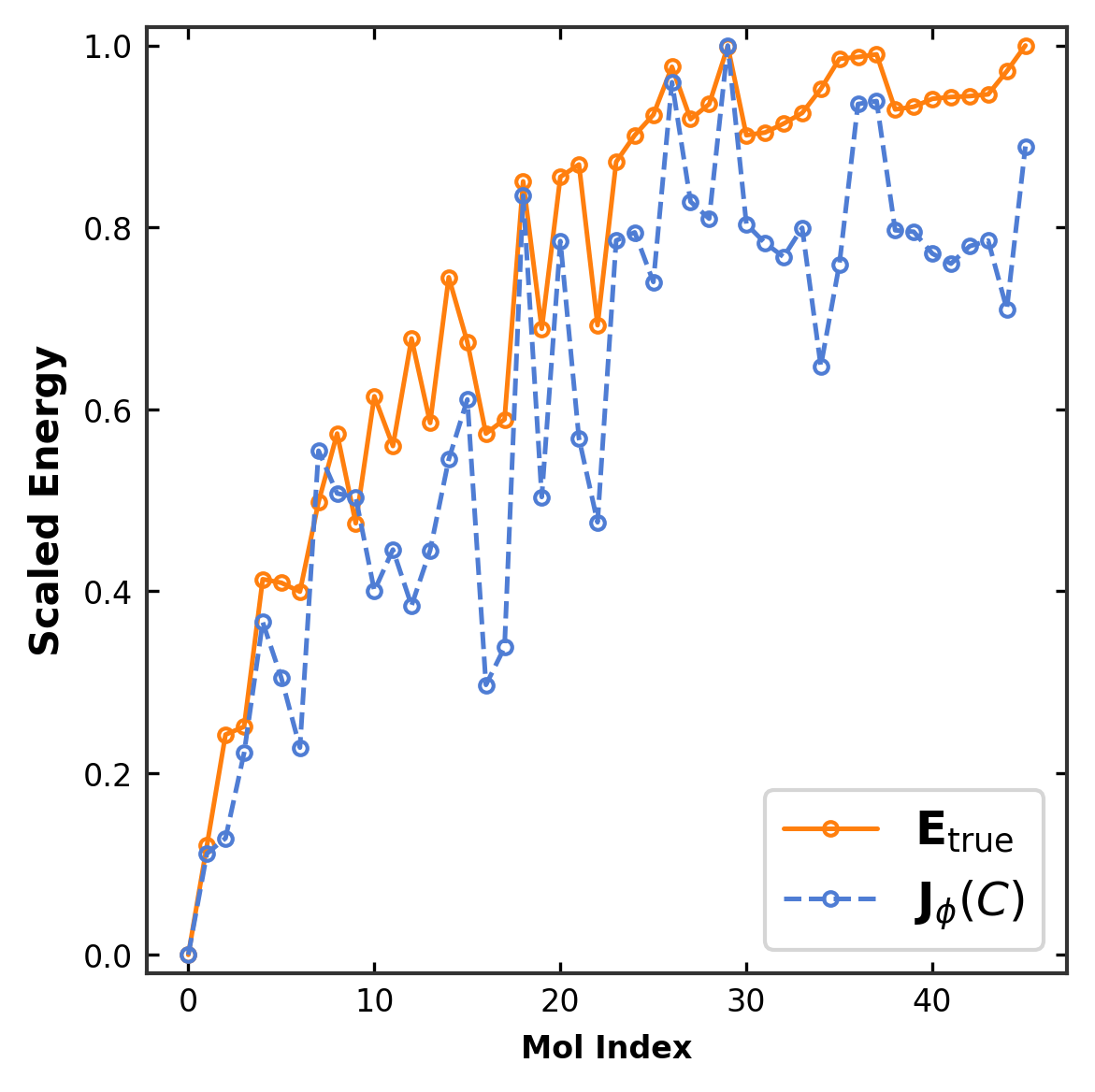} &
      \includegraphics[width=.215\textwidth]{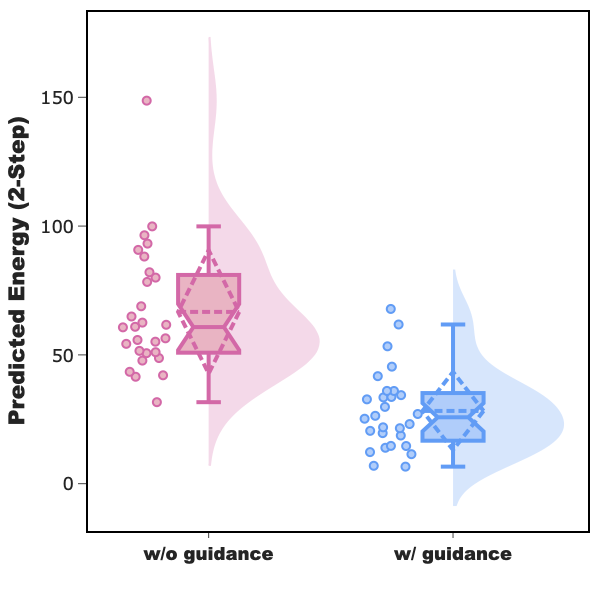} &
      \includegraphics[width=.215\textwidth]{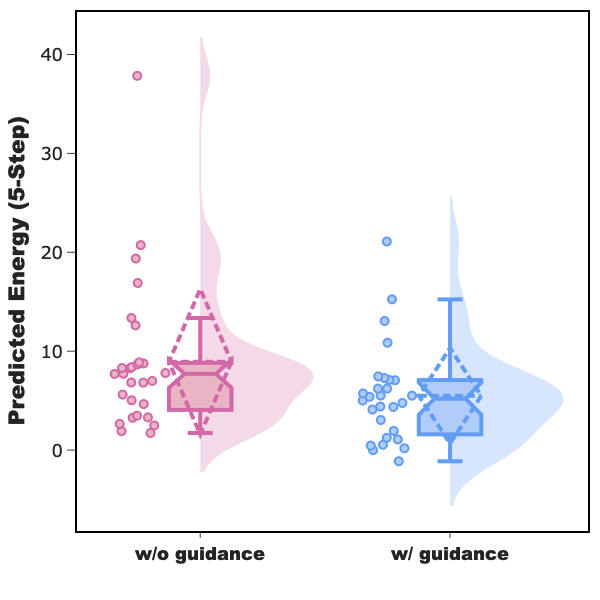}
   \end{tabular*}
   \caption{ (a) Six representative molecules from the GEOM-Drugs dataset. (b)
   Their energy landscapes (conformers within each molecule are ordered by their
   Boltzmann weights from the dataset): $\mathbf{E}_{\text{true}}$ denotes the ground-truth
   energy values provided by the GEOM dataset, and $\mathbf{J}_{\phi}(C)$ denotes the learned energy landscape given
   by our EBM $\mathbf{J}_{\phi}$. Both are normalized to the range [0, 1] to make
   their variation comparable. (c)–(d) Predicted energies from
   $\mathbf{J}_{\phi}$ for conformations generated using 2 ODE steps (c) and 5 ODE steps (d)
   for each molecule. The unguided
   baseline (ET-Flow; \textbf{w/o guidance}) is shown in red, and the guided
   model (\ours; \textbf{w/ guidance}) is shown in blue. }
   \label{fig: landscape_show}
\end{figure*}

   \clearpage
   \bibliographystyle{unsrt}
   \small
   \bibliography{manuscript/references}

@article{guimaraes2012use,
  title={Use of {3D} Properties to Characterize Beyond Rule-of-5 Property Space for Passive Permeation},
  author={Guimar{\~a}es, Cristiano R. W. and Mathiowetz, Alan M. and Shalaeva, Marina and Goetz, Gilles and Liras, Spiros},
  journal={Journal of Chemical Information and Modeling},
  volume={52},
  number={4},
  pages={882--890},
  year={2012},
  publisher={ACS Publications},
  doi={10.1021/ci300010y},
  url={https://doi.org/10.1021/ci300010y}
}

@article{schwab2010conformations,
  title={Conformations and {3D} pharmacophore searching},
  author={Schwab, Christof H.},
  journal={Drug Discovery Today: Technologies},
  volume={7},
  number={4},
  pages={e245--e253},
  year={2010},
  publisher={Elsevier},
  doi={10.1016/j.ddtec.2010.10.003},
  url={https://doi.org/10.1016/j.ddtec.2010.10.003}
}

@article{bhati2026integrative,
  title={Integrative Physics and Machine Learning-Based Optimal Binding Pose Generator for Protein-Ligand Complexes},
  author={Bhati, Agastya P. and Dong, Jinyu and Black, Sean and Wan, Shunzhou and Gutta, Vineeth and Zhang, Xibei and Bieniek, Mateusz and Green, Darren V. S. and Chandrasekaran, Sunita and Stahlberg, Eric and Coveney, Peter V.},
  journal={ChemRxiv},
  year={2026},
  doi={10.26434/chemrxiv.15009330/v1},
  url={https://doi.org/10.26434/chemrxiv.15009330/v1},
  note={Preprint, version 1. \mbox{DOI: \url{https://doi.org/10.26434/chemrxiv.15009330/v1}}}
}

@article{hawkins2017conformation,
  title={Conformation Generation: The State of the Art},
  author={Hawkins, Paul C. D.},
  journal={Journal of Chemical Information and Modeling},
  volume={57},
  number={8},
  pages={1747--1756},
  year={2017},
  publisher={ACS Publications},
  doi={10.1021/acs.jcim.7b00221},
  url={https://doi.org/10.1021/acs.jcim.7b00221}
}

@article{ballard2015exploiting,
  title={Exploiting the potential energy landscape to sample free energy},
  author={Ballard, Andrew J. and Martiniani, Stefano and Stevenson, Jacob D. and Somani, Sandeep and Wales, David J.},
  journal={WIREs Computational Molecular Science},
  volume={5},
  number={3},
  pages={273--289},
  year={2015},
  publisher={Wiley Online Library},
  doi={10.1002/wcms.1217},
  url={https://doi.org/10.1002/wcms.1217}
}

@article{de2016role,
  title={Role of Molecular Dynamics and Related Methods in Drug Discovery},
  author={De Vivo, Marco and Masetti, Matteo and Bottegoni, Giovanni and Cavalli, Andrea},
  journal={Journal of Medicinal Chemistry},
  volume={59},
  number={9},
  pages={4035--4061},
  year={2016},
  publisher={ACS Publications},
  doi={10.1021/acs.jmedchem.5b01684},
  url={https://doi.org/10.1021/acs.jmedchem.5b01684}
}

@article{pracht2020automated,
  title={Automated exploration of the low-energy chemical space with fast quantum chemical methods},
  author={Pracht, Philipp and Bohle, Fabian and Grimme, Stefan},
  journal={Physical Chemistry Chemical Physics},
  volume={22},
  number={14},
  pages={7169--7192},
  year={2020},
  publisher={Royal Society of Chemistry},
  doi={10.1039/c9cp06869d},
  url={https://doi.org/10.1039/c9cp06869d}
}

@article{parr1979local,
  title={Local density functional theory of atoms and molecules},
  author={Parr, Robert G. and Gadre, Shridhar R. and Bartolotti, Libero J.},
  journal={Proceedings of the National Academy of Sciences},
  volume={76},
  number={6},
  pages={2522--2526},
  year={1979},
  doi={10.1073/pnas.76.6.2522},
  url={https://doi.org/10.1073/pnas.76.6.2522}
}

@article{muller1994glossary,
  title={Glossary of terms used in physical organic chemistry ({IUPAC} Recommendations 1994)},
  author={Muller, P.},
  journal={Pure and Applied Chemistry},
  volume={66},
  number={5},
  pages={1077--1184},
  year={1994},
  doi={10.1351/pac199466051077},
  url={https://doi.org/10.1351/pac199466051077}
}

@book{zhou2021machine,
  title={Machine Learning},
  author={Zhou, Zhi-Hua},
  year={2021},
  publisher={Springer Singapore},
  doi={10.1007/978-981-15-1967-3},
  isbn={9789811519666},
  url={https://doi.org/10.1007/978-981-15-1967-3}
}

@book{alpaydin2021machine,
  title={Machine Learning},
  author={Alpayd{\i}n, Ethem},
  edition={Revised and updated},
  year={2021},
  publisher={MIT Press},
  doi={10.7551/mitpress/13811.001.0001},
  isbn={9780262542524},
  url={https://doi.org/10.7551/mitpress/13811.001.0001}
}

@article{janiesch2021machine,
  title={Machine learning and deep learning},
  author={Janiesch, Christian and Zschech, Patrick and Heinrich, Kai},
  journal={Electronic Markets},
  volume={31},
  number={3},
  pages={685--695},
  year={2021},
  publisher={Springer},
  doi={10.1007/s12525-021-00475-2},
  url={https://doi.org/10.1007/s12525-021-00475-2}
}

@article{thomas2018tensor,
  title={Tensor field networks: Rotation- and translation-equivariant neural networks for {3D} point clouds},
  author={Thomas, Nathaniel and Smidt, Tess and Kearnes, Steven and Yang, Lusann and Li, Li and Kohlhoff, Kai and Riley, Patrick},
  journal={arXiv preprint arXiv:1802.08219},
  year={2018},
  eprint={1802.08219},
  archivePrefix={arXiv},
  url={https://arxiv.org/abs/1802.08219}
}

@inproceedings{lipmanflow,
  title={Flow Matching for Generative Modeling},
  author={Lipman, Yaron and Chen, Ricky T. Q. and Ben-Hamu, Heli and Nickel, Maximilian and Le, Matthew},
  booktitle={International Conference on Learning Representations},
  year={2023},
  url={https://openreview.net/forum?id=PqvMRDCJT9t},
  note={\url{https://openreview.net/forum?id=PqvMRDCJT9t}}
}

@inproceedings{liu2022flow,
  title={Flow Straight and Fast: Learning to Generate and Transfer Data with Rectified Flow},
  author={Liu, Xingchao and Gong, Chengyue and Liu, Qiang},
  booktitle={International Conference on Learning Representations},
  year={2023},
  url={https://openreview.net/forum?id=XVjTT1nw5z},
  note={\url{https://openreview.net/forum?id=XVjTT1nw5z}}
}

@article{lipman2024flow,
  title={Flow Matching Guide and Code},
  author={Lipman, Yaron and Havasi, Marton and Holderrieth, Peter and Shaul, Neta and Le, Matt and Karrer, Brian and Chen, Ricky T. Q. and Lopez-Paz, David and Ben-Hamu, Heli and Gat, Itai},
  journal={arXiv preprint arXiv:2412.06264},
  year={2024},
  eprint={2412.06264},
  archivePrefix={arXiv},
  url={https://arxiv.org/abs/2412.06264}
}

@article{tong2024improving,
  title={Improving and generalizing flow-based generative models with minibatch optimal transport},
  author={Tong, Alexander and Fatras, Kilian and Malkin, Nikolay and Huguet, Guillaume and Zhang, Yanlei and Rector-Brooks, Jarrid and Wolf, Guy and Bengio, Yoshua},
  journal={Transactions on Machine Learning Research},
  pages={1--34},
  year={2024},
  url={https://openreview.net/forum?id=CD9Snc73AW}
}

@inproceedings{feng2025guidance,
  title={On the Guidance of Flow Matching},
  author={Feng, Ruiqi and Yu, Chenglei and Deng, Wenhao and Hu, Peiyan and Wu, Tailin},
  booktitle={Proceedings of the 42nd International Conference on Machine Learning},
  series={Proceedings of Machine Learning Research},
  volume={267},
  pages={16993--17029},
  year={2025},
  publisher={PMLR},
  url={https://proceedings.mlr.press/v267/feng25s.html}
}

@article{zheng2023guided,
  title={Guided Flows for Generative Modeling and Decision Making},
  author={Zheng, Qinqing and Le, Matt and Shaul, Neta and Lipman, Yaron and Grover, Aditya and Chen, Ricky T. Q.},
  journal={arXiv preprint arXiv:2311.13443},
  year={2023},
  eprint={2311.13443},
  archivePrefix={arXiv},
  url={https://arxiv.org/abs/2311.13443}
}

@inproceedings{kollovieh2024flow,
  title={Flow Matching with Gaussian Process Priors for Probabilistic Time Series Forecasting},
  author={Marcel Kollovieh and Marten Lienen and David L{\"u}dke and Leo Schwinn and Stephan G{\"u}nnemann},
  booktitle={International Conference on Learning Representations},
  year={2025},
  url={https://openreview.net/forum?id=uxVBbSlKQ4},
  note={\url{https://openreview.net/forum?id=uxVBbSlKQ4}}
}

@inproceedings{zhang2025energy,
  title={Energy-Weighted Flow Matching for Offline Reinforcement Learning},
  author={Zhang, Shiyuan and Zhang, Weitong and Gu, Quanquan},
  booktitle={International Conference on Learning Representations},
  pages={17943--17970},
  year={2025},
  url={https://proceedings.iclr.cc/paper_files/paper/2025/file/2d9c6cdb4cfe93869c090fea7375044b-Paper-Conference.pdf}
}

@inproceedings{ma2024sit,
  title={{SiT}: Exploring Flow and Diffusion-Based Generative Models with Scalable Interpolant Transformers},
  author={Ma, Nanye and Goldstein, Mark and Albergo, Michael S. and Boffi, Nicholas M. and Vanden-Eijnden, Eric and Xie, Saining},
  booktitle={Computer Vision -- ECCV 2024},
  series={Lecture Notes in Computer Science},
  pages={23--40},
  year={2024},
  publisher={Springer},
  doi={10.1007/978-3-031-72980-5_2},
  url={https://doi.org/10.1007/978-3-031-72980-5_2}
}

@inproceedings{dhariwal2021diffusion,
  title={Diffusion Models Beat {GANs} on Image Synthesis},
  author={Dhariwal, Prafulla and Nichol, Alexander},
  booktitle={Advances in Neural Information Processing Systems},
  volume={34},
  pages={8780--8794},
  year={2021},
  url={https://proceedings.neurips.cc/paper_files/paper/2021/file/49ad23d1ec9fa4bd8d77d02681df5cfa-Paper.pdf}
}

@inproceedings{song2023loss,
  title={Loss-Guided Diffusion Models for Plug-and-Play Controllable Generation},
  author={Song, Jiaming and Zhang, Qinsheng and Yin, Hongxu and Mardani, Morteza and Liu, Ming-Yu and Kautz, Jan and Chen, Yongxin and Vahdat, Arash},
  booktitle={Proceedings of the 40th International Conference on Machine Learning},
  series={Proceedings of Machine Learning Research},
  volume={202},
  pages={32483--32498},
  year={2023},
  publisher={PMLR},
  url={https://proceedings.mlr.press/v202/song23k.html}
}

@inproceedings{chung2022diffusion,
  title={Diffusion Posterior Sampling for General Noisy Inverse Problems},
  author={Chung, Hyungjin and Kim, Jeongsol and Mccann, Michael T. and Klasky, Marc L. and Ye, Jong Chul},
  booktitle={International Conference on Learning Representations},
  year={2023},
  url={https://openreview.net/pdf/dd7f2e1f5581d91eb4c1ff34fec78b93d3dfa599.pdf},
  note={\url{https://openreview.net/pdf/dd7f2e1f5581d91eb4c1ff34fec78b93d3dfa599.pdf}}
}

@inproceedings{lu2023contrastive,
  title={Contrastive Energy Prediction for Exact Energy-Guided Diffusion Sampling in Offline Reinforcement Learning},
  author={Lu, Cheng and Chen, Huayu and Chen, Jianfei and Su, Hang and Li, Chongxuan and Zhu, Jun},
  booktitle={Proceedings of the 40th International Conference on Machine Learning},
  series={Proceedings of Machine Learning Research},
  volume={202},
  pages={22825--22855},
  year={2023},
  publisher={PMLR},
  url={https://proceedings.mlr.press/v202/lu23d.html}
}

@inproceedings{balcerak2025energy,
  title={Energy Matching: Unifying Flow Matching and Energy-Based Models for Generative Modeling},
  author={Balcerak, Michal and Amiranashvili, Tamaz and Terpin, Antonio and Shit, Suprosanna and Bogensperger, Lea and Kaltenbach, Sebastian and Koumoutsakos, Petros and Menze, Bjoern},
  booktitle={Advances in Neural Information Processing Systems},
  volume={38},
  pages={8583--8609},
  year={2025},
  publisher={Curran Associates, Inc.},
  doi={10.52202/085713-0291},
  url={https://proceedings.neurips.cc/paper_files/paper/2025/file/0cbbdfb0a4098af8dc7a497a5e59aff7-Paper-Conference.pdf}
}

@inproceedings{stark2023harmonic,
  title={Harmonic Self-Conditioned Flow Matching for joint Multi-Ligand Docking and Binding Site Design},
  author={St{\"a}rk, Hannes and Jing, Bowen and Barzilay, Regina and Jaakkola, Tommi},
  booktitle={Proceedings of the 41st International Conference on Machine Learning},
  series={Proceedings of Machine Learning Research},
  volume={235},
  pages={46468--46494},
  year={2024},
  publisher={PMLR},
  url={https://proceedings.mlr.press/v235/stark24a.html}
}

@article{jing2023eigenfold,
  title={{EigenFold}: Generative Protein Structure Prediction with Diffusion Models},
  author={Jing, Bowen and Erives, Ezra and Pao-Huang, Peter and Corso, Gabriele and Berger, Bonnie and Jaakkola, Tommi},
  journal={arXiv preprint arXiv:2304.02198},
  year={2023},
  eprint={2304.02198},
  archivePrefix={arXiv},
  url={https://arxiv.org/abs/2304.02198}
}

@inproceedings{cuturi2013sinkhorn,
  title={Sinkhorn Distances: Lightspeed Computation of Optimal Transport},
  author={Cuturi, Marco},
  booktitle={Advances in Neural Information Processing Systems},
  volume={26},
  year={2013},
  url={https://proceedings.neurips.cc/paper_files/paper/2013/file/af21d0c97db2e27e13572cbf59eb343d-Paper.pdf},
  note={\url{https://proceedings.neurips.cc/paper_files/paper/2013/file/af21d0c97db2e27e13572cbf59eb343d-Paper.pdf}}
}

@inproceedings{welling2011bayesian,
  title={Bayesian learning via stochastic gradient Langevin dynamics},
  author={Welling, Max and Teh, Yee Whye},
  booktitle={Proceedings of the 28th international conference on machine learning (ICML-11)},
  pages={681--688},
  year={2011},
  url={https://icml.cc/2011/papers/398_icmlpaper.pdf}
}

@article{ba2016layer,
  title={Layer Normalization},
  author={Ba, Jimmy Lei and Kiros, Jamie Ryan and Hinton, Geoffrey E.},
  journal={arXiv preprint arXiv:1607.06450},
  year={2016},
  eprint={1607.06450},
  archivePrefix={arXiv},
  url={https://arxiv.org/abs/1607.06450}
}

@inproceedings{schutt2021equivariant,
  title={Equivariant message passing for the prediction of tensorial properties and molecular spectra},
  author={Sch{\"u}tt, Kristof and Unke, Oliver and Gastegger, Michael},
  booktitle={Proceedings of the 38th International Conference on Machine Learning},
  volume={139},
  series={Proceedings of Machine Learning Research},
  pages={9377--9388},
  year={2021},
  publisher={PMLR},
  url={https://proceedings.mlr.press/v139/schutt21a.html}
}

@article{unke2019physnet,
  title={{PhysNet}: A Neural Network for Predicting Energies, Forces, Dipole Moments, and Partial Charges},
  author={Unke, Oliver T. and Meuwly, Markus},
  journal={Journal of Chemical Theory and Computation},
  volume={15},
  number={6},
  pages={3678--3693},
  year={2019},
  publisher={ACS Publications},
  doi={10.1021/acs.jctc.9b00181},
  url={https://doi.org/10.1021/acs.jctc.9b00181}
}

@inproceedings{tholke2022torchmd,
  title={{TorchMD-NET}: Equivariant Transformers for Neural Network based Molecular Potentials},
  author={Th{\"o}lke, Philipp and De Fabritiis, Gianni},
  booktitle={International Conference on Learning Representations},
  year={2022},
  url={https://openreview.net/forum?id=zNHzqZ9wrRB},
  note={\url{https://openreview.net/forum?id=zNHzqZ9wrRB}}
}

@inproceedings{dehghani2023scaling,
  title={Scaling Vision Transformers to 22 Billion Parameters},
  author={Dehghani, Mostafa and Djolonga, Josip and Mustafa, Basil and Padlewski, Piotr and Heek, Jonathan and Gilmer, Justin and Steiner, Andreas Peter and Caron, Mathilde and Geirhos, Robert and Alabdulmohsin, Ibrahim and others},
  booktitle={Proceedings of the 40th International Conference on Machine Learning},
  series={Proceedings of Machine Learning Research},
  volume={202},
  pages={7480--7512},
  year={2023},
  publisher={PMLR},
  url={https://proceedings.mlr.press/v202/dehghani23a.html}
}

@inproceedings{esser2024scaling,
  title={Scaling Rectified Flow Transformers for High-Resolution Image Synthesis},
  author={Esser, Patrick and Kulal, Sumith and Blattmann, Andreas and Entezari, Rahim and M{\"u}ller, Jonas and Saini, Harry and Levi, Yam and Lorenz, Dominik and Sauer, Axel and Boesel, Frederic and others},
  booktitle={Proceedings of the 41st International Conference on Machine Learning},
  series={Proceedings of Machine Learning Research},
  volume={235},
  pages={12606--12633},
  year={2024},
  publisher={PMLR},
  url={https://proceedings.mlr.press/v235/esser24a.html}
}

@inproceedings{kingma2013auto,
  title={Auto-Encoding Variational Bayes},
  author={Kingma, Diederik P. and Welling, Max},
  booktitle={International Conference on Learning Representations},
  year={2014},
  url={https://arxiv.org/abs/1312.6114},
  note={\url{https://arxiv.org/abs/1312.6114}}
}

@inproceedings{rezende2015variational,
  title={Variational Inference with Normalizing Flows},
  author={Rezende, Danilo and Mohamed, Shakir},
  booktitle={Proceedings of the 32nd International Conference on Machine Learning},
  series={Proceedings of Machine Learning Research},
  volume={37},
  pages={1530--1538},
  year={2015},
  publisher={PMLR},
  url={https://proceedings.mlr.press/v37/rezende15.html}
}

@inproceedings{sohl2015deep,
  title={Deep Unsupervised Learning using Nonequilibrium Thermodynamics},
  author={Sohl-Dickstein, Jascha and Weiss, Eric and Maheswaranathan, Niru and Ganguli, Surya},
  booktitle={Proceedings of the 32nd International Conference on Machine Learning},
  series={Proceedings of Machine Learning Research},
  volume={37},
  pages={2256--2265},
  year={2015},
  publisher={PMLR},
  url={https://proceedings.mlr.press/v37/sohl-dickstein15.html}
}

@inproceedings{song2019generative,
  title={Generative Modeling by Estimating Gradients of the Data Distribution},
  author={Song, Yang and Ermon, Stefano},
  booktitle={Advances in Neural Information Processing Systems},
  volume={32},
  year={2019},
  url={https://proceedings.neurips.cc/paper_files/paper/2019/file/3001ef257407d5a371a96dcd947c7d93-Paper.pdf},
  note={\url{https://proceedings.neurips.cc/paper_files/paper/2019/file/3001ef257407d5a371a96dcd947c7d93-Paper.pdf}}
}

@inproceedings{ho2020denoising,
  title={Denoising Diffusion Probabilistic Models},
  author={Ho, Jonathan and Jain, Ajay and Abbeel, Pieter},
  booktitle={Advances in Neural Information Processing Systems},
  volume={33},
  pages={6840--6851},
  year={2020},
  url={https://proceedings.neurips.cc/paper_files/paper/2020/file/4c5bcfec8584af0d967f1ab10179ca4b-Paper.pdf}
}

@inproceedings{song2020score,
  title={Score-Based Generative Modeling through Stochastic Differential Equations},
  author={Song, Yang and Sohl-Dickstein, Jascha and Kingma, Diederik P. and Kumar, Abhishek and Ermon, Stefano and Poole, Ben},
  booktitle={International Conference on Learning Representations},
  year={2021},
  url={https://openreview.net/forum?id=PxTIG12RRHS},
  note={\url{https://openreview.net/forum?id=PxTIG12RRHS}}
}

@article{ho2022classifier,
  title={Classifier-Free Diffusion Guidance},
  author={Ho, Jonathan and Salimans, Tim},
  journal={arXiv preprint arXiv:2207.12598},
  year={2022},
  eprint={2207.12598},
  archivePrefix={arXiv},
  url={https://arxiv.org/abs/2207.12598}
}

@inproceedings{rombach2022high,
  title={High-Resolution Image Synthesis With Latent Diffusion Models},
  author={Rombach, Robin and Blattmann, Andreas and Lorenz, Dominik and Esser, Patrick and Ommer, Bj\"orn},
  booktitle={Proceedings of the IEEE/CVF Conference on Computer Vision and Pattern Recognition (CVPR)},
  pages={10684--10695},
  year={2022},
  doi={10.1109/CVPR52688.2022.01042},
  url={https://openaccess.thecvf.com/content/CVPR2022/html/Rombach_High-Resolution_Image_Synthesis_With_Latent_Diffusion_Models_CVPR_2022_paper.html}
}

@inproceedings{song2023consistency,
  title={Consistency Models},
  author={Song, Yang and Dhariwal, Prafulla and Chen, Mark and Sutskever, Ilya},
  booktitle={Proceedings of the 40th International Conference on Machine Learning},
  series={Proceedings of Machine Learning Research},
  volume={202},
  pages={32211--32252},
  year={2023},
  publisher={PMLR},
  url={https://proceedings.mlr.press/v202/song23a.html}
}

@article{zheng2024ensemble,
  title={Ensemble {Kalman} Diffusion Guidance: A Derivative-free Method for Inverse Problems},
  author={Hongkai Zheng and Wenda Chu and Austin Wang and Nikola Borislavov Kovachki and Ricardo Baptista and Yisong Yue},
  journal={Transactions on Machine Learning Research},
  year={2025},
  issn={2835-8856},
  url={https://openreview.net/forum?id=XPEEsKneKs},
  note={\url{https://openreview.net/forum?id=XPEEsKneKs}}
}

@article{axelrod2022geom,
  title={{GEOM}, energy-annotated molecular conformations for property prediction and molecular generation},
  author={Axelrod, Simon and Gomez-Bombarelli, Rafael},
  journal={Scientific Data},
  volume={9},
  number={1},
  pages={185},
  year={2022},
  publisher={Nature Publishing Group UK London},
  doi={10.1038/s41597-022-01288-4},
  url={https://doi.org/10.1038/s41597-022-01288-4}
}

@misc{DVN/JNGTDF_2021,
  title={{GEOM}},
  author={Axelrod, Simon and Gomez-Bombarelli, Rafael},
  year={2021},
  publisher={Harvard Dataverse},
  howpublished={Harvard Dataverse, Version V4, \url{https://doi.org/10.7910/DVN/JNGTDF}},
  version={V4},
  doi={10.7910/DVN/JNGTDF},
  url={https://doi.org/10.7910/DVN/JNGTDF}
}

@article{rappe1992uff,
  title={{UFF}, a full periodic table force field for molecular mechanics and molecular dynamics simulations},
  author={Rapp{\'e}, Anthony K. and Casewit, Carla J. and Colwell, K. S. and Goddard, III, William A. and Skiff, W. Mason},
  journal={Journal of the American Chemical Society},
  volume={114},
  number={25},
  pages={10024--10035},
  year={1992},
  publisher={ACS Publications},
  doi={10.1021/ja00051a040},
  url={https://doi.org/10.1021/ja00051a040}
}

@article{halgren1996merck,
  title={Merck molecular force field. V. Extension of {MMFF94} using experimental data, additional computational data, and empirical rules},
  author={Halgren, Thomas A.},
  journal={Journal of Computational Chemistry},
  volume={17},
  number={5-6},
  pages={616--641},
  year={1996},
  publisher={Wiley Online Library},
  doi={10.1002/(sici)1096-987x(199604)17:5/6<616::aid-jcc5>3.0.co;2-x},
  url={https://doi.org/10.1002/(sici)1096-987x(199604)17:5/6<616::aid-jcc5>3.0.co;2-x}
}

@article{liberti2014euclidean,
  title={Euclidean Distance Geometry and Applications},
  author={Liberti, Leo and Lavor, Carlile and Maculan, Nelson and Mucherino, Antonio},
  journal={SIAM Review},
  volume={56},
  number={1},
  pages={3--69},
  year={2014},
  publisher={SIAM},
  doi={10.1137/120875909},
  url={https://doi.org/10.1137/120875909}
}

@misc{rdkit,
  title={{RDKit}: Open-source cheminformatics},
  author={{RDKit contributors}},
  howpublished={\url{https://www.rdkit.org/}},
  url={https://www.rdkit.org/},
  note={Accessed: 2025-08-07}
}

@inproceedings{simm2019generative,
  title={A Generative Model for Molecular Distance Geometry},
  author={Simm, Gregor N. C. and Hern{\'a}ndez-Lobato, Jos{\'e} Miguel},
  booktitle={Proceedings of the 37th International Conference on Machine Learning},
  series={Proceedings of Machine Learning Research},
  volume={119},
  pages={8949--8958},
  year={2020},
  publisher={PMLR},
  url={https://proceedings.mlr.press/v119/simm20a.html}
}

@inproceedings{xu2021learning,
  title={Learning Neural Generative Dynamics for Molecular Conformation Generation},
  author={Minkai Xu and Shitong Luo and Yoshua Bengio and Jian Peng and Jian Tang},
  booktitle={International Conference on Learning Representations},
  year={2021},
  url={https://openreview.net/forum?id=pAbm1qfheGk},
  note={\url{https://openreview.net/forum?id=pAbm1qfheGk}}
}

@inproceedings{xu2021end,
  title={An End-to-End Framework for Molecular Conformation Generation via Bilevel Programming},
  author={Xu, Minkai and Wang, Wujie and Luo, Shitong and Shi, Chence and Bengio, Yoshua and Gomez-Bombarelli, Rafael and Tang, Jian},
  booktitle={Proceedings of the 38th International Conference on Machine Learning},
  series={Proceedings of Machine Learning Research},
  volume={139},
  pages={11537--11547},
  year={2021},
  publisher={PMLR},
  url={https://proceedings.mlr.press/v139/xu21f.html}
}

@inproceedings{shi2021learning,
  title={Learning Gradient Fields for Molecular Conformation Generation},
  author={Shi, Chence and Luo, Shitong and Xu, Minkai and Tang, Jian},
  booktitle={Proceedings of the 38th International Conference on Machine Learning},
  series={Proceedings of Machine Learning Research},
  volume={139},
  pages={9558--9568},
  year={2021},
  publisher={PMLR},
  url={https://proceedings.mlr.press/v139/shi21b.html}
}

@inproceedings{ganea2021geomol,
  title={{GeoMol}: Torsional Geometric Generation of Molecular {3D} Conformer Ensembles},
  author={Ganea, Octavian and Pattanaik, Lagnajit and Coley, Connor and Barzilay, Regina and Jensen, Klavs and Green, William and Jaakkola, Tommi},
  booktitle={Advances in Neural Information Processing Systems},
  volume={34},
  pages={13757--13769},
  year={2021},
  url={https://proceedings.neurips.cc/paper_files/paper/2021/file/725215ed82ab6306919b485b81ff9615-Paper.pdf}
}

@inproceedings{guan2021energy,
  title={Energy-inspired molecular conformation optimization},
  author={Guan, Jiaqi and Qian, Wesley Wei and Liu, Qiang and Ma, Wei-Ying and Ma, Jianzhu and Peng, Jian},
  booktitle={International Conference on Learning Representations},
  year={2022},
  url={https://openreview.net/forum?id=7QfLW-XZTl},
  note={\url{https://openreview.net/forum?id=7QfLW-XZTl}}
}

@inproceedings{xu2022geodiff,
  title={{GeoDiff}: A Geometric Diffusion Model for Molecular Conformation Generation},
  author={Minkai Xu and Lantao Yu and Yang Song and Chence Shi and Stefano Ermon and Jian Tang},
  booktitle={International Conference on Learning Representations},
  year={2022},
  url={https://openreview.net/forum?id=PzcvxEMzvQC},
  note={\url{https://openreview.net/forum?id=PzcvxEMzvQC}}
}

@inproceedings{jing2022torsional,
  title={Torsional Diffusion for Molecular Conformer Generation},
  author={Jing, Bowen and Corso, Gabriele and Chang, Jeffrey and Barzilay, Regina and Jaakkola, Tommi},
  booktitle={Advances in Neural Information Processing Systems},
  volume={35},
  pages={24240--24253},
  year={2022},
  url={https://proceedings.neurips.cc/paper_files/paper/2022/file/994545b2308bbbbc97e3e687ea9e464f-Paper-Conference.pdf}
}

@inproceedings{wang2023swallowing,
  title={Swallowing the Bitter Pill: Simplified Scalable Conformer Generation},
  author={Wang, Yuyang and Elhag, Ahmed A. A. and Jaitly, Navdeep and Susskind, Joshua M. and Bautista, Miguel \'{A}ngel},
  booktitle={Proceedings of the 41st International Conference on Machine Learning},
  series={Proceedings of Machine Learning Research},
  volume={235},
  pages={50400--50418},
  year={2024},
  publisher={PMLR},
  url={https://proceedings.mlr.press/v235/wang24q.html}
}

@article{fan2024ec,
  title={{EC-Conf}: A ultra-fast diffusion model for molecular conformation generation with equivariant consistency},
  author={Fan, Zhiguang and Yang, Yuedong and Xu, Mingyuan and Chen, Hongming},
  journal={Journal of Cheminformatics},
  volume={16},
  number={1},
  pages={107},
  year={2024},
  publisher={Springer},
  doi={10.1186/s13321-024-00893-2},
  url={https://doi.org/10.1186/s13321-024-00893-2}
}

@inproceedings{hassan2024flow,
  title={{ET-Flow}: Equivariant Flow-Matching for Molecular Conformer Generation},
  author={Hassan, Majdi and Shenoy, Nikhil and Lee, Jungyoon and St\"{a}rk, Hannes and Thaler, Stephan and Beaini, Dominique},
  booktitle={Advances in Neural Information Processing Systems},
  volume={37},
  pages={128798--128824},
  year={2024},
  url={https://proceedings.neurips.cc/paper_files/paper/2024/file/e8bd617e7dd0394ceadf37b4a7773179-Paper-Conference.pdf}
}

@inproceedings{cao2025efficient,
  title={Efficient Molecular Conformer Generation with {SO}(3)-Averaged Flow Matching and Reflow},
  author={Cao, Zhonglin and Geiger, Mario and Costa, Allan Dos Santos and Reidenbach, Danny and Kreis, Karsten and Geffner, Tomas and Pellegrini, Franco and Zhou, Guoqing and Kucukbenli, Emine},
  booktitle={Proceedings of the 42nd International Conference on Machine Learning},
  series={Proceedings of Machine Learning Research},
  volume={267},
  pages={6522--6542},
  year={2025},
  publisher={PMLR},
  url={https://proceedings.mlr.press/v267/cao25b.html}
}

@inproceedings{xu2025cofm,
  title={{CoFM}: Molecular Conformation Generation via Flow Matching in {SE(3)}-Invariant Latent Space},
  author={Xu, Guikun and Yu, Yankai and Jiang, Yongquan and Yang, Yan and Bian, Yatao},
  booktitle={ICML 2025 Generative AI and Biology (GenBio) Workshop},
  year={2025},
  url={https://openreview.net/forum?id=C0jrjy4F1D},
  note={\url{https://openreview.net/forum?id=C0jrjy4F1D}}
}

@inproceedings{vaswani2017attention,
  title={Attention is All you Need},
  author={Vaswani, Ashish and Shazeer, Noam and Parmar, Niki and Uszkoreit, Jakob and Jones, Llion and Gomez, Aidan N. and Kaiser, \L ukasz and Polosukhin, Illia},
  booktitle={Advances in Neural Information Processing Systems},
  volume={30},
  year={2017},
  url={https://proceedings.neurips.cc/paper_files/paper/2017/file/3f5ee243547dee91fbd053c1c4a845aa-Paper.pdf},
  note={\url{https://proceedings.neurips.cc/paper_files/paper/2017/file/3f5ee243547dee91fbd053c1c4a845aa-Paper.pdf}}
}

@misc{landrum2013rdkit,
  title={{RDKit}: A software suite for cheminformatics, computational chemistry, and predictive modeling},
  author={Landrum, Greg},
  howpublished={\url{https://www.rdkit.org/RDKit_Overview.pdf}},
  url={https://www.rdkit.org/RDKit_Overview.pdf},
  note={Software overview, undated}
}

@inproceedings{hu2019strategies,
  title={Strategies for Pre-training Graph Neural Networks},
  author={Hu, Weihua and Liu, Bowen and Gomes, Joseph and Zitnik, Marinka and Liang, Percy and Pande, Vijay and Leskovec, Jure},
  booktitle={International Conference on Learning Representations},
  year={2020},
  url={https://openreview.net/forum?id=HJlWWJSFDH},
  note={\url{https://openreview.net/forum?id=HJlWWJSFDH}}
}

@inproceedings{brody2021attentive,
  title={How Attentive are Graph Attention Networks?},
  author={Shaked Brody and Uri Alon and Eran Yahav},
  booktitle={International Conference on Learning Representations},
  year={2022},
  url={https://openreview.net/forum?id=F72ximsx7C1},
  note={\url{https://openreview.net/forum?id=F72ximsx7C1}}
}

@inproceedings{rampavsek2022recipe,
  title={Recipe for a General, Powerful, Scalable Graph Transformer},
  author={Ramp\'{a}\v{s}ek, Ladislav and Galkin, Michael and Dwivedi, Vijay Prakash and Luu, Anh Tuan and Wolf, Guy and Beaini, Dominique},
  booktitle={Advances in Neural Information Processing Systems},
  volume={35},
  pages={14501--14515},
  year={2022},
  url={https://proceedings.neurips.cc/paper_files/paper/2022/file/5d4834a159f1547b267a05a4e2b7cf5e-Paper-Conference.pdf}
}

@inproceedings{ying2021transformers,
  title={Do Transformers Really Perform Badly for Graph Representation?},
  author={Ying, Chengxuan and Cai, Tianle and Luo, Shengjie and Zheng, Shuxin and Ke, Guolin and He, Di and Shen, Yanming and Liu, Tie-Yan},
  booktitle={Advances in Neural Information Processing Systems},
  volume={34},
  pages={28877--28888},
  year={2021},
  url={https://proceedings.neurips.cc/paper_files/paper/2021/file/f1c1592588411002af340cbaedd6fc33-Paper.pdf}
}

@article{xu2021molecule3d,
  title={{Molecule3D}: A Benchmark for Predicting {3D} Geometries from Molecular Graphs},
  author={Xu, Zhao and Luo, Youzhi and Zhang, Xuan and Xu, Xinyi and Xie, Yaochen and Liu, Meng and Dickerson, Kaleb and Deng, Cheng and Nakata, Maho and Ji, Shuiwang},
  journal={arXiv preprint arXiv:2110.01717},
  year={2021},
  eprint={2110.01717},
  archivePrefix={arXiv},
  url={https://arxiv.org/abs/2110.01717}
}

@inproceedings{xu2023gtmgc,
  title={{GTMGC}: Using Graph Transformer to Predict Molecule's Ground-State Conformation},
  author={Xu, Guikun and Jiang, Yongquan and Lei, PengChuan and Yang, Yan and Chen, Jim},
  booktitle={International Conference on Learning Representations},
  year={2024},
  url={https://openreview.net/forum?id=F7QnIKlC1N},
  note={\url{https://openreview.net/forum?id=F7QnIKlC1N}}
}

@inproceedings{luo2024bridging,
  title={Bridging Geometric States via Geometric Diffusion Bridge},
  author={Luo, Shengjie and Xu, Yixian and He, Di and Zheng, Shuxin and Liu, Tie-Yan and Wang, Liwei},
  booktitle={Advances in Neural Information Processing Systems},
  volume={37},
  pages={109283--109322},
  year={2024},
  url={https://proceedings.neurips.cc/paper_files/paper/2024/file/c565514eb8cf8808d82a612c5f7a1920-Paper-Conference.pdf}
}

@inproceedings{wangwgformer,
  title={{WGF}ormer: An {SE}(3)-Transformer Driven by {W}asserstein Gradient Flows for Molecular Ground-State Conformation Prediction},
  author={Wang, Fanmeng and Cheng, Minjie and Xu, Hongteng},
  booktitle={Proceedings of the 42nd International Conference on Machine Learning},
  series={Proceedings of Machine Learning Research},
  volume={267},
  pages={62652--62671},
  year={2025},
  publisher={PMLR},
  url={https://proceedings.mlr.press/v267/wang25s.html}
}

@inproceedings{kim2025rebind,
  title={{REBIND}: Enhancing Ground-state Molecular Conformation Prediction via Force-Based Graph Rewiring},
  author={Kim, Taewon and Seo, Hyunjin and Ahn, Sungsoo and Yang, Eunho},
  booktitle={International Conference on Learning Representations},
  pages={8359--8376},
  year={2025},
  url={https://proceedings.iclr.cc/paper_files/paper/2025/file/1812042b83f20707a898ff6f8af7db84-Paper-Conference.pdf}
}

@inproceedings{satorras2021n,
  title={{E(n)} Equivariant Graph Neural Networks},
  author={Satorras, V\'{\i}ctor Garcia and Hoogeboom, Emiel and Welling, Max},
  booktitle={Proceedings of the 38th International Conference on Machine Learning},
  series={Proceedings of Machine Learning Research},
  volume={139},
  pages={9323--9332},
  year={2021},
  publisher={PMLR},
  url={https://proceedings.mlr.press/v139/satorras21a.html}
}

@article{pracht2024crest,
  title={{CREST}---A program for the exploration of low-energy molecular chemical space},
  author={Pracht, Philipp and Grimme, Stefan and Bannwarth, Christoph and Bohle, Fabian and Ehlert, Sebastian and Feldmann, Gereon and Gorges, Johannes and M{\"u}ller, Marcel and Neudecker, Tim and Plett, Christoph and others},
  journal={The Journal of Chemical Physics},
  volume={160},
  number={11},
  pages={114110},
  year={2024},
  publisher={AIP Publishing},
  doi={10.1063/5.0197592},
  url={https://doi.org/10.1063/5.0197592}
}

@article{bannwarth2019gfn2,
  title={{GFN2-xTB}---An Accurate and Broadly Parametrized Self-Consistent Tight-Binding Quantum Chemical Method with Multipole Electrostatics and Density-Dependent Dispersion Contributions},
  author={Bannwarth, Christoph and Ehlert, Sebastian and Grimme, Stefan},
  journal={Journal of Chemical Theory and Computation},
  volume={15},
  number={3},
  pages={1652--1671},
  year={2019},
  publisher={ACS Publications},
  doi={10.1021/acs.jctc.8b01176},
  url={https://doi.org/10.1021/acs.jctc.8b01176}
}

@misc{xu2026enflowprocessed,
  title={Energy-Guided Generative Modeling for Low-Energy Molecular Structure Discovery},
  author={Xu, Guikun},
  year={2026},
  publisher={Zenodo},
  howpublished={Zenodo, version v1, \url{https://doi.org/10.5281/zenodo.18044125}},
  version={v1},
  doi={10.5281/zenodo.18044125},
  url={https://doi.org/10.5281/zenodo.18044125},
  note={Processed GEOM data}
}

@misc{xu2026enflowcheckpoints,
  title={Energy-Guided Generative Modeling for Low-Energy Molecular Structure Discovery},
  author={Xu, Guikun},
  year={2026},
  publisher={Zenodo},
  howpublished={Zenodo, version v1, \url{https://doi.org/10.5281/zenodo.18045494}},
  version={v1},
  doi={10.5281/zenodo.18045494},
  url={https://doi.org/10.5281/zenodo.18045494},
  note={Model checkpoints}
}
\end{document}